\documentclass[a4paper,11pt]{article}

\usepackage[margin=1in, heightrounded]{geometry}

\usepackage{newpxtext}          
\usepackage{microtype}          
\microtypesetup{expansion=true, protrusion=true, final}

\usepackage{amsmath,amssymb,mathtools,bm,amsthm}
\usepackage{libertinus}

\usepackage{graphicx}
\graphicspath{{figures/}{fig/}{images/}{./figures/}}
\usepackage{subcaption}
\usepackage{float}
\usepackage[labelfont=bf,skip=8pt]{caption}

\usepackage{tikz}
\usetikzlibrary{shapes.geometric, arrows.meta, positioning, calc, shadows.blur}

\usepackage{booktabs}
\usepackage{tabularx}
\usepackage{multirow}
\usepackage{array}
\usepackage{longtable}
\usepackage{adjustbox}

\usepackage[ruled,vlined,linesnumbered,algoruled]{algorithm2e}
\SetAlCapFnt{\small\bfseries}
\SetAlCapNameFnt{\small\bfseries}
\SetAlCapHSkip{0pt}
\IncMargin{-\parindent}

\usepackage{xcolor}
\usepackage{listings}
\lstdefinestyle{moderncode}{
    basicstyle=\ttfamily\small,
    keywordstyle=\color{blue!80!black}\bfseries,
    commentstyle=\color{gray!60}\itshape,
    stringstyle=\color{red!70!black},
    numberstyle=\tiny\color{gray!60},
    numbers=left,
    numbersep=10pt,
    backgroundcolor=\color{gray!6},
    frame=single,
    framerule=0.4pt,
    rulecolor=\color{gray!40},
    breaklines=true,
    breakatwhitespace=true,
    tabsize=2,
    showstringspaces=false,
    captionpos=b,
    aboveskip=1.2em,
    belowskip=1em,
    language=Python,
    morekeywords={self,nn,torch,tensorflow,numpy,asarray,zeros,ones,
                  Input,Output,Return,Predict,Compute,Update,forward}
}
\definecolor{topbg}{HTML}{E8EDF2}   \definecolor{topec}{HTML}{6B7D8A}
\definecolor{relbg}{HTML}{DCEEFB}   \definecolor{relec}{HTML}{6BAED6}
\definecolor{relsub}{HTML}{EAF4FC}  \definecolor{relsec}{HTML}{A5CDE6}
\definecolor{datbg}{HTML}{FDE8D0}   \definecolor{datec}{HTML}{D4A373}
\definecolor{datsub}{HTML}{FEF3E6}  \definecolor{datsec}{HTML}{DFC09E}
\definecolor{cmpbg}{HTML}{D9F0D6}   \definecolor{cmpec}{HTML}{7DBB72}
\definecolor{cmpsub}{HTML}{E9F5E7}  \definecolor{cmpsec}{HTML}{A3D49A}
\definecolor{thmbg}{HTML}{E8DAF0}   \definecolor{thmec}{HTML}{A07DBB}
\definecolor{thmsub}{HTML}{F1E6F7}  \definecolor{thmsec}{HTML}{BFA3D4}
\definecolor{linecol}{HTML}{90A4AE}
\definecolor{txtcol}{HTML}{263238}

\usepackage[
    backend=biber,
    style=numeric-comp,
    sorting=none,
    giveninits=true,
    maxbibnames=99,
    maxcitenames=2,
    doi=true,
    url=true,
    eprint=false,
    backref=true
]{biblatex}
\usepackage[
    colorlinks=true,
    linkcolor=blue!70!black,
    citecolor=blue!70!black,
    urlcolor=blue!70!black,
    pdfauthor={Your Name},
    pdftitle={Your Title},
    pdfkeywords={PDE, foundation models, exascale, differentiable physics}
]{hyperref}
\usepackage{bookmark}           

\usepackage[autostyle=true,english=american]{csquotes}
\usepackage{siunitx}            
\usepackage{enumitem}
\usepackage{pdflscape}
\usepackage{balance}
\usepackage{lineno}

\usepackage{setspace}
\usepackage{authblk}

\definecolor{jwbpurple}{RGB}{128,0,128}

\usepackage{setspace}
\usepackage{textcomp}   
\usepackage{subdepth}   
\usepackage{newunicodechar}
\newunicodechar{₂}{$_2$}

\begin{document}

\title{\Large Partial Differential Equations in the Age of Machine Learning:
A Critical Synthesis of Classical, Machine Learning, and Hybrid Methods}

\author[1*]{Mohammad Nooraiepour}
\author[2]{Jakub Wiktor Both}
\author[3]{Teeratorn Kadeethum}
\author[4]{Saeid Sadeghnejad}

\affilsep=0.5em plus 0.2em minus 0.2em
\affil[1,*]{\raggedright \footnotesize Faculty of Mathematics and Natural Sciences, University of Oslo, 1047 Blindern, 0316 Oslo, Norway}
\affil[2]{\raggedright \footnotesize Department of Mathematics, University of Bergen, Bergen, Norway}
\affil[3]{\raggedright \footnotesize AI Lab, Siemens Energy, Orlando, FL 32826, USA}
\affil[4]{\raggedright \footnotesize Institute for Geosciences, Applied Geology, Friedrich-Schiller-University Jena, 07749 Jena, Germany}
\affil[*]{%
  \vspace{0.5ex}
  \footnotesize
  \textit{Corresponding author. Email}: \href{mailto:monoo@uio.no}{\texttt{monoo@uio.no}}%
}

\date{}

\maketitle

\begin{abstract}
\begin{spacing}{0.99}
\noindent Partial differential equations (PDEs) govern physical phenomena across the full range of scientific scales, yet their computational solution remains one of the defining challenges of modern science. This critical review examines two mature but epistemologically distinct paradigms for PDE solution, classical numerical methods and machine learning approaches, through a unified evaluative framework organized around six fundamental computational challenges: high dimensionality, nonlinearity, geometric complexity, discontinuities, multiscale
phenomena, and multiphysics coupling. Classical methods, including finite difference, finite element, finite volume, and spectral discretizations, are assessed for their structure-preserving properties, rigorous convergence theory, and scalable solver design; their persistent limitations in high-dimensional and geometrically complex settings are characterized precisely. Machine learning approaches, including physics-informed neural networks, neural operators, graph
architectures, transformers, generative models, and hybrid frameworks, are introduced under a taxonomy organized by the degree to which physical knowledge is incorporated and subjected to the same critical evaluation applied to classical methods. The central finding is that classical methods are deductive---errors are bounded by quantities derivable from PDE structure and discretization parameters---while machine learning methods are inductive---accuracy depends on statistical proximity to the training distribution. This epistemological distinction, rather than computational speed, is the primary criterion governing responsible method selection. We identify three genuine complementarities between the paradigms and develop principles for hybrid design, including a framework for the structure inheritance problem that addresses when classical guarantees propagate through hybrid couplings, and an error budget decomposition that separates discretization, neural approximation, and coupling contributions. We further assess emerging frontiers, including foundation models for scientific computing, differentiable programming for inverse design, quantum algorithms, and exascale co-design, evaluating each against the structural constraints that determine whether current barriers are fundamental or contingent on engineering progress. The conclusion is not that one paradigm supersedes the other, but that their principled integration, grounded in the complementary strengths each provides, offers the most credible path toward computationally tractable solutions to the grand challenge problems that neither can address alone.\\

\end{spacing}
\end{abstract}


\section{Introduction}
\label{sec:introduction}

Partial differential equations (PDEs) constitute the mathematical foundation for describing physical phenomena across scientific and engineering disciplines, from quantum wave functions to atmospheric circulation and subsurface flow. While PDEs provide an elegant theoretical framework, their computational solution remains one of the central challenges of scientific computing. This challenge has driven three centuries of algorithmic development in numerical analysis and, more recently, has motivated increasing interest in machine-learning-based approaches for PDE solution.

\begin{figure}[t]
    \centering
    \includegraphics[width=0.95\textwidth]{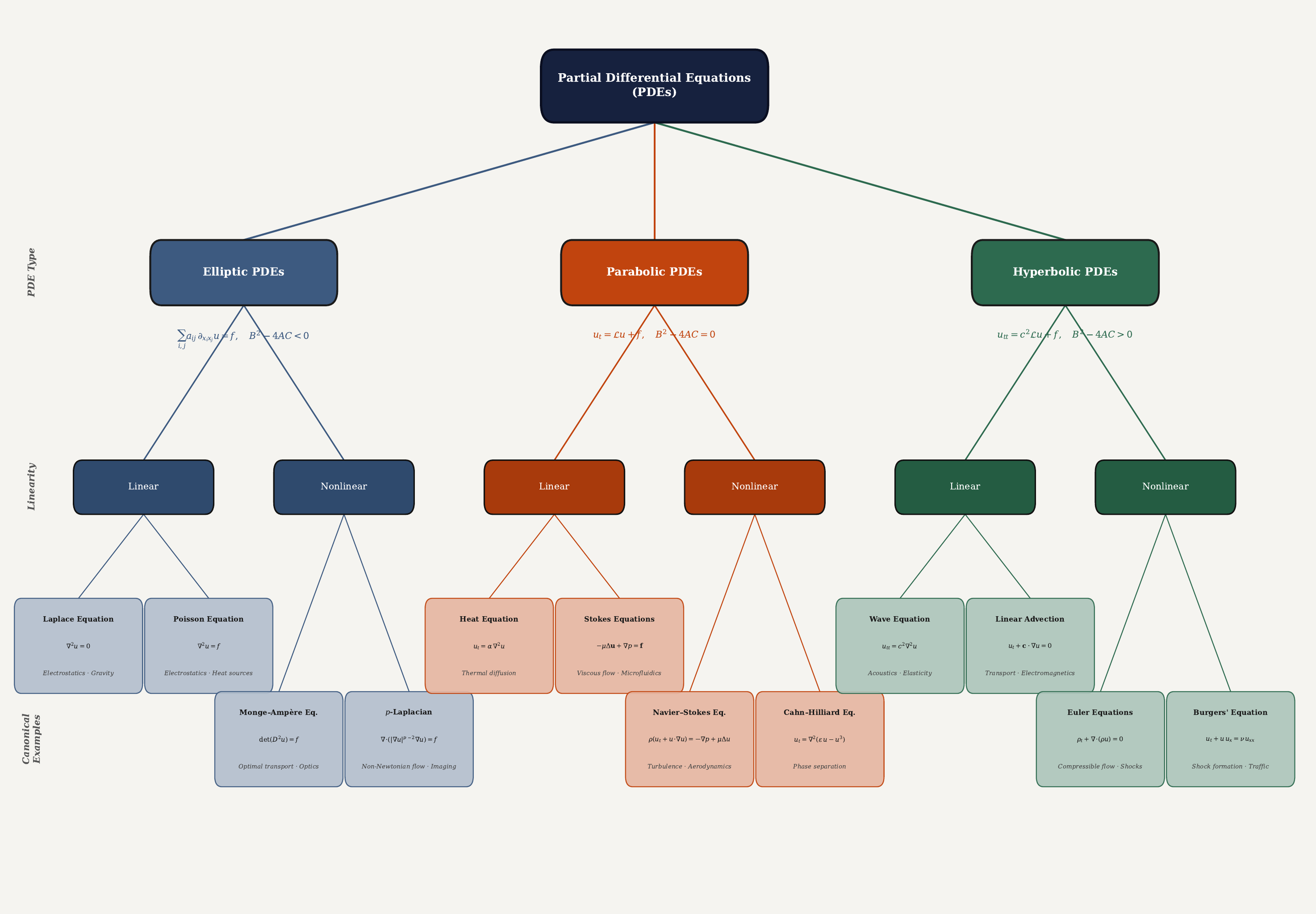}
    \caption{Hierarchical taxonomy of partial differential equations (PDEs), organized by type (elliptic, parabolic, hyperbolic), linearity, and canonical examples with their governing mathematical forms.}
    \label{fig:pde_taxonomy}
\end{figure}

The fundamental taxonomy of PDEs distinguishes elliptic, parabolic, and hyperbolic
equations based on their characteristic structure and physical behavior~\cite{strauss2007partial,olver2014introduction} (Fig.~\ref{fig:pde_taxonomy}). Elliptic equations govern steady-state equilibrium phenomena---electrostatics, structural mechanics, gravitational potential---where boundary conditions influence the solution globally throughout the entire domain~\cite{dupaigne2011stable, ponce2016elliptic}. Parabolic equations, including the heat equation, the heat equation and the viscous Navier–Stokes equations, and the Cahn-Hilliard equation for phase separation, govern dissipative
processes evolving irreversibly in time, with smoothing properties that regularize
even discontinuous initial data~\cite{nochetto2016pde,perthame2015parabolic}.
Hyperbolic equations describe wave propagation with finite information speed,
preserving sharp gradients and discontinuities along characteristic curves---a
property with direct consequences for numerical scheme design that distinguishes
them fundamentally from elliptic and parabolic problems~\cite{godlewski2013numerical,
lefloch2002hyperbolic,leveque2002finite}. Many problems of practical importance exhibit mixed character that cannot be
assigned cleanly to any single class: the compressible Navier-Stokes equations
transition between parabolic and hyperbolic regimes with flow velocity; multiphase
flow in porous media couples parabolic diffusion with hyperbolic convection;
fluid-structure interaction and reaction-diffusion systems couple equations of
distinct types across moving interfaces. This classification is not merely descriptive. It determines which mathematical structures---variational principles, conservation forms, characteristic theory, energy-dissipation inequalities---govern the problem, and therefore which discretization strategies, solver designs, and structure-preserving properties are appropriate. A structured overview of this classification, together with canonical equations and their physical domains, is presented in Figure~\ref{fig:pde_taxonomy}. The six computational challenges examined in Section~\ref{sec:challenges} arise
precisely at the boundaries of this classification: in the nonlinearity that complicates each type, in the geometric complexity that all types must negotiate, and above all, in the multiphysics coupling that combines types whose mathematical characters conflict.

The distinction between linear and nonlinear PDEs profoundly shapes both
analytical tractability and computational strategy~\cite{schneider2017nonlinear,
logan2008introduction}. Linear equations satisfy the superposition principle, enabling analytical solution techniques and computationally efficient sparse linear solvers whose convergence properties are well understood~\cite{saad2003iterative}.
Nonlinear PDEs exhibit qualitatively richer behavior: solutions may develop
finite-time singularities from smooth initial data, nonuniqueness may arise even under seemingly well-posed formulations, and small parameter changes can trigger bifurcations between qualitatively different solution regimes---phenomena central to turbulence,
pattern formation, and shock dynamics. Critically, nonlinearity does not merely increase computational cost; it fundamentally alters the robustness requirements of numerical methods, demanding solver designs that remain stable and convergent across parameter regimes in which the character of the solution itself changes.
This distinction between nominal accuracy and practical robustness recurs
throughout this review as a primary criterion for evaluating both classical and
machine learning approaches.

These mathematical structures — and the robustness requirements nonlinearity imposes on them — constrain admissible computational methods and profoundly influence solution accuracy. Classical numerical method development has been driven primarily by three application domains: structural mechanics (finite element methods emphasizing variational formulations), computational fluid dynamics (finite volume and spectral methods for conservation laws), and wave propagation (finite difference schemes for time-dependent hyperbolic systems). Modern methods increasingly emphasize structure-preserving properties enforced through careful algorithmic design \cite{hairer2006geometric}. For elliptic problems coupling multiple fields, this includes inf-sup stability conditions ensuring compatible discretization of mixed formulations in solid mechanics and incompressible flow \cite{boffi2013mixed,bathe2001inf,bochev2004inf}. For parabolic equations, energy-dissipative schemes maintain thermodynamic consistency. For hyperbolic systems, entropy-stable and energy-preserving schemes prevent spurious solution behavior while maintaining conservation. These structure-preserving properties include maximum principles, conservation laws, energy stability, symplectic structure, and inf-sup compatibility.
They represent decades of refinement in numerical analysis and constitute essential benchmarks for evaluating emerging machine-learning-based approaches.

The computational demands of contemporary applications expose critical limitations of classical numerical methods. High-dimensional problems, from quantum many-body systems \cite{cazorla2017simulation} to financial risk models \cite{golbabai2019numerical}, face exponential scaling that renders grid-based approaches intractable beyond modest dimensions \cite{rude2018research,alber2019integrating,downey2018think}. Multiphysics phenomena \cite{faroughi2022physics,cai2021physics,zhao2024comprehensive} couple processes across vastly separated spatial and temporal scales, from molecular to planetary. Complex patient-specific geometries in biomedical applications and intricate industrial configurations require mesh generation efforts that can exceed simulation costs \cite{mazumder2015numerical,karniadakis2021physics,markidis2021old}. These computational barriers have intensified interest in machine learning methods \cite{huang2025partial,lawal2022physics,cai2021physics} that promise to transcend classical scaling limitations by leveraging implicit function representations and learned approximations.

The landscape of machine learning methods for PDE solution has transformed dramatically in the past decade. Physics-informed neural networks~\cite{raissi2019physics,lawal2022physics,cai2021physics}
embed differential constraints directly into loss functions, enabling solution
approximation with sparse training data. Neural operators~\cite{lu2021learning,li2020fourier,kovachki2023neural} learn mappings
between function spaces rather than pointwise approximations, sharing conceptual
structure with snapshot-based reduced-order models while extending them to nonlinear, data-driven settings. Foundation models pretrained on diverse PDE families demonstrate preliminary zero-shot generalization~\cite{herde2024poseidon,sun2025towards}, suggesting potential paths toward transferable solution operators. These developments raise two sets of questions that are related but distinct, and that most existing surveys address inadequately. The first concerns accuracy and certification: when do neural methods outperform classical approaches, what rigorous error bounds exist for learned solutions, and how can predictions be verified for safety-critical applications where solution errors carry physical consequences? The second concerns physical structure: can neural methods preserve---by design rather than by coincidence---the maximum principles, conservation laws, energy-dissipation properties, symplectic structure, and inf-sup compatibility that classical methods achieve through careful mathematical construction? These are not peripheral concerns. The structure-preserving properties established in the preceding paragraph took decades
to develop in classical methods and constitute the primary mechanism by which those
methods earn trust in scientific and engineering practice. Any evaluation of machine learning approaches to PDE solution that does not systematically address both questions is incomplete.

This review critically examines both classical numerical approaches and emerging machine learning techniques for PDE solution. Rather than merely surveying methods, we focus on fundamental trade-offs: classical methods \cite{christofides2002nonlinear,klainerman2010pde,offner2023approximation} provide provable accuracy guarantees, rigorous error bounds, and structure-preserving properties through mathematical design, but face severe scaling challenges with dimensionality and geometric complexity \cite{mazumder2015numerical,karniadakis2021physics,markidis2021old}. Neural approaches \cite{huang2025partial,lawal2022physics,cai2021physics,toscano2025pinns,gonon2024overview,kovachki2023neural} offer inference speed, dimensional flexibility, and mesh-free operation, but currently lack systematic error control, struggle with extrapolation beyond training regimes, and require extensive validation for structure preservation. The central contribution of this synthesis is the identification of genuine complementarities between the two paradigms — regions where each addresses difficulties that are fundamental for the other — and the development of principles for hybrid design grounded in the structure inheritance problem and the error budget framework that governs how classical guarantees propagate through coupled systems.

The review is organized around a logical progression from problem characterization
through method evaluation to synthesis and outlook. Section~\ref{sec:challenges} establishes the six fundamental computational challenges that motivate new approaches---high dimensionality, nonlinearity, geometric complexity,
discontinuities, multiscale phenomena, and multiphysics coupling---providing the
evaluative vocabulary used throughout. Sections~\ref{sec:classicmethods} and~\ref{sec:evaluationclassical} together constitute a rigorous treatment of classical numerical methods: the former develops their mathematical foundations and structure-preserving properties; the latter critically evaluates where decades of refinement have achieved practical solutions and where fundamental barriers persist.
Sections~\ref{sec:mlmethods} and~\ref{subsec:ml_challenges} introduce the principal families of machine learning approaches for PDE solution under a unified taxonomy, organized by the degree to which physical knowledge is incorporated, and subject them to the same critical assessment applied to classical methods. Section~\ref{sec:synthesis} does not summarize what preceded it, but advances the
analysis: it examines what fundamentally distinguishes the two paradigms at the level
of mathematical character and error structure, identifies their genuine complementarities, and develops principles for hybrid design, including a framework
for the structure inheritance problem and the error budget decomposition that governs how classical guarantees propagate through coupled classical-learned systems.
Section~\ref{sec:emerging} assesses four developments that will shape the near-term
capability landscape---large-scale foundation models for scientific computing, quantum algorithms, differentiable programming for inverse design, and exascale algorithmic co-design---evaluating each against the structural constraints that determine whether current barriers are fundamental or contingent on engineering progress.

This review is written for researchers and practitioners who require evidence-based
guidance rather than a catalogue of methods: the goal is not encyclopedic coverage but critical synthesis. Specifically, the paper aims to identify what demonstrably works in practice and why, what remains uncertain despite optimistic claims, where the genuine complementarities between classical and machine learning approaches lie, and where the most consequential open problems are---those whose resolution would alter the capability landscape rather than refine it at the margins. We assume familiarity with the foundations of PDE theory and numerical analysis, and take as a starting point the observation that neither paradigm is self-sufficient: classical methods provide the mathematical structure and certification that scientific
applications demand, while machine learning provides the adaptability and amortization that high-dimensional and many-query problems require. Understanding both---and the principled discipline of combining them---is the central
competence this review is designed to deliver.

\section{Fundamental Challenges in Computational PDE Solution}
\label{sec:challenges}

The computational solution of partial differential equations (PDEs) confronts six fundamental challenges that establish the practical limits of both classical numerical methods and emerging machine learning approaches. These challenges arise from the mathematical structure of PDEs, the complexity of the physical phenomena they model, and the computational resources required to approximate them. Understanding these interconnected difficulties is essential for evaluating methodological trade-offs, identifying research frontiers, and selecting appropriate solution strategies. 

Well-posed PDE problems satisfy three fundamental criteria in the sense of Hadamard \cite{klainerman2010pde,ashyralyev2012well}: \emph{existence} (a solution exists), \emph{uniqueness} (the solution is determined uniquely by the problem data), and \emph{stability} (small perturbations in initial or boundary conditions produce correspondingly small changes in the solution). These properties ensure that PDE models yield meaningful predictions and that computational approximations converge to true solutions. The notion of solution itself must often be generalized beyond classical differentiability: weak solutions formulated through variational principles \cite{dupaigne2011stable} or distributional solutions using test functions allow treatment of problems where classical solutions fail to exist. This mathematical framework underlies all computational methods, as algorithms must respect the structure that ensures well-posedness, and solution concepts must match the problem's regularity.

Beyond computational cost, the fundamental challenges raise critical questions about solution verification and quality assessment \cite{fernandez2022regularity}. Regularity theory, establishing solution smoothness under specified conditions, provides the mathematical foundation for error estimates in classical numerical methods. Elliptic regularity results guarantee that solutions exhibit greater smoothness than boundary data might suggest, enabling high-order approximation \cite{wang2006schauder}. Parabolic equations exhibit instantaneous smoothing that regularizes even discontinuous initial data \cite{sell2013dynamics}, while hyperbolic equations propagate singularities along characteristics without smoothing. 

\begin{figure}[t]
    \centering
    \includegraphics[width=0.95\textwidth]{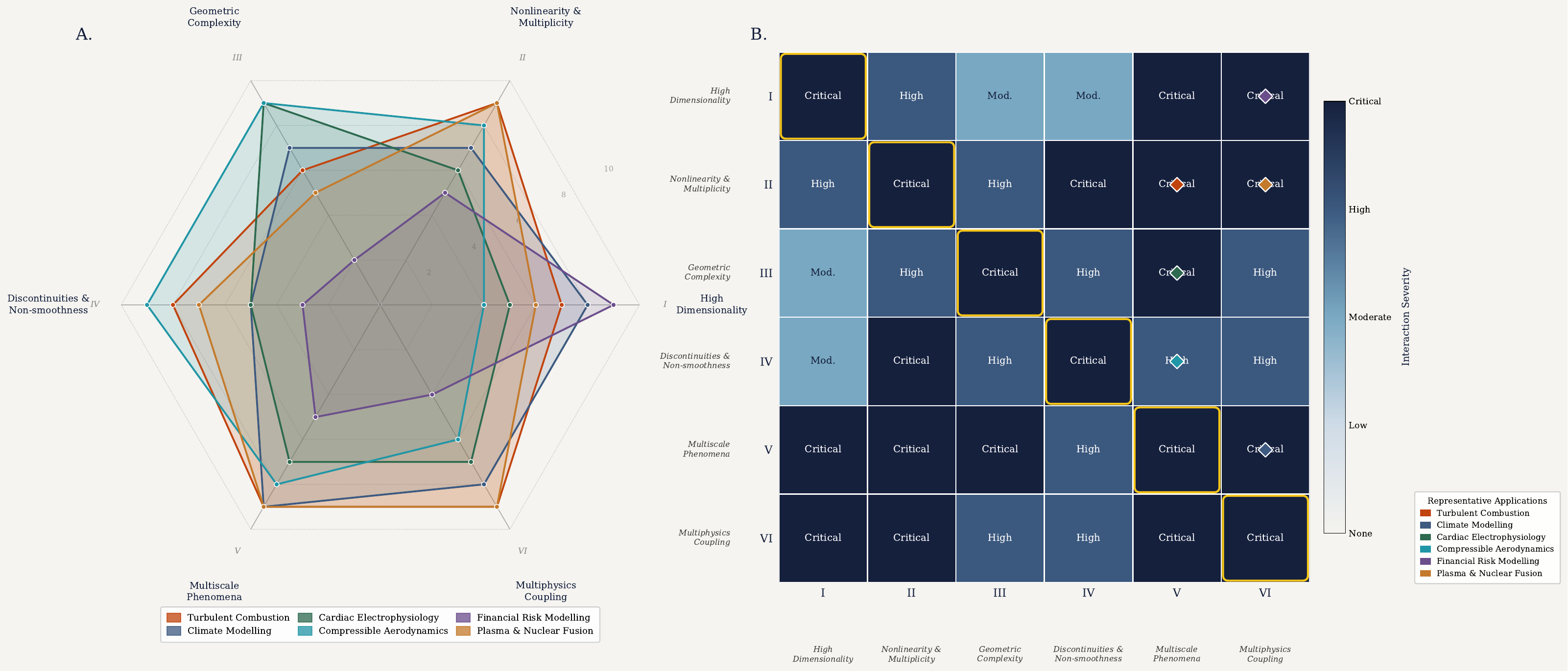}
    \caption{Characterisation of the six fundamental computational challenges in PDE solution.
    \textbf{A.}~Radar chart showing the challenge severity profile (scored 0--10) for six representative application domains across the axes: (I)~high dimensionality, (II)~nonlinearity and multiplicity,
    (III)~geometric complexity, (IV)~discontinuities and non-smoothness,
    (V)~multiscale phenomena, and (VI)~multiphysics coupling.
    \textbf{B.}~Symmetric interaction-severity matrix quantifying how each pair of challenges amplifies computational difficulty. The diagonal entries denote inherent challenge severity, and the markers
    indicate the dominant co-occurring challenge pair for each application.}
    \label{fig:pde_challenges}
\end{figure}

Figure~\ref{fig:pde_challenges} provides a structured overview of these six challenges and their pairwise interactions. Panel~A reveals that no single application is confined to a single challenge axis — every domain considered spans at least four challenges simultaneously, with turbulent combustion and plasma physics exhibiting near-critical scores across all six challenges. Panel~B shows that multiscale phenomena and multiphysics coupling are the most pervasively interactive challenges, amplifying difficulty across every other axis, a pattern that recurs throughout the methodological discussions in Sections~\ref{sec:classicmethods}--\ref{subsec:ml_challenges}.

The challenges we describe below represent practical computational barriers that arise even when well-posedness is mathematically established. Each challenge manifests distinct difficulties in discretization, approximation, and solution verification. While we present them individually for clarity, real-world applications frequently involve multiple challenges simultaneously, as discussed in Section~\ref{subsec:concurrent_challenges}.

\subsection{The Curse of Dimensionality}
\label{subsec:dimensionality}

High-dimensional PDEs arise naturally across diverse application domains \cite{han2018solving,schwab2011sparse}: quantum many-body systems where dimensionality scales with particle count \cite{continentino2017quantum}, financial models where dimensions correspond to asset or risk factor counts \cite{golbabai2019numerical}, and parameter optimization where design variables create high-dimensional spaces. The curse of dimensionality, termed by Bellman~\cite{bellman1966dynamic}, manifests as exponential scaling: for a PDE in $d$ spatial dimensions discretized with $N$ points per dimension, traditional grid-based methods require $N^d$ degrees of freedom, yielding memory requirements of $O(N^d)$ and computational cost exceeding $O(N^d)$ per time step for explicit schemes and $O(N^{d+1})$ or higher for implicit schemes requiring linear system solution.

Consider the multi-asset Black-Scholes equation for option pricing with $d$ underlying assets \cite{golbabai2019numerical,frey2011nonlinear}:
\begin{equation}
\label{eq:black_scholes}
\frac{\partial V}{\partial t} + \frac{1}{2}\sum_{i,j=1}^d \rho_{ij}\sigma_i\sigma_j S_i S_j \frac{\partial^2 V}{\partial S_i \partial S_j} + \sum_{i=1}^d r S_i \frac{\partial V}{\partial S_i} - rV = 0,
\end{equation}
where $V(S_1,\ldots,S_d,t)$ represents the option value, $S_i$ denotes asset prices, $\sigma_i$ are volatilities, $\rho_{ij}$ is the correlation matrix, and $r$ is the risk-free interest rate. For a portfolio with $d = 20$ assets and modest discretization using $N = 100$ points per dimension, storing the solution requires approximately $10^{40}$ floating-point numbers—orders of magnitude beyond available memory. Similarly, the many-body Schrödinger equation \cite{berezin2012schrodinger} operates in $3N_p$-dimensional configuration space for $N_p$ particles, making direct solution computationally intractable beyond a few particles.


\subsection{Nonlinearity and Multiplicity of Solutions}
\label{subsec:nonlinearity}

Nonlinear PDEs are commonly key to accurate modeling of complex phenomena, thus offering high fidelity and principled generalization across regimes, respecting and preserving invariances, structure and conservation laws. However, Nonlinear PDEs exhibit qualitatively different phenomena from their linear counterparts \cite{schneider2017nonlinear,christofides2002nonlinear,logan2008introduction}: nonuniqueness of solutions may occur even for well-posed problems, finite-time singularities can develop from smooth initial data, and small parameter changes trigger bifurcations between qualitatively different solution regimes. These behaviors challenge both numerical solution methods and the ability to verify computational results.

The Navier-Stokes equations for incompressible viscous flow exemplify these challenges \cite{lemarie2023navier,tsai2018lectures,cai2021physics}:
\begin{align}
\label{eq:navier_stokes_momentum}
\frac{\partial \mathbf{u}}{\partial t} + (\mathbf{u} \cdot \nabla)\mathbf{u} &= -\frac{1}{\rho}\nabla p + \nu \nabla^2 \mathbf{u}, \\
\label{eq:navier_stokes_continuity}
\nabla \cdot \mathbf{u} &= 0,
\end{align}
where $\mathbf{u}(\mathbf{x},t)$ is the velocity field, $p(\mathbf{x},t)$ is pressure, $\rho$ is density, and $\nu$ is kinematic viscosity. The quadratic convective nonlinearity $(\mathbf{u} \cdot \nabla)\mathbf{u}$ generates energy transfer across spatial scales, ultimately producing turbulent flow at high Reynolds numbers $\mathrm{Re} = UL/\nu$ (where $U$ and $L$ are characteristic velocity and length scales). Direct numerical simulation requires resolving all scales from domain size $L$ down to the Kolmogorov dissipation scale $\eta \sim \mathrm{Re}^{-3/4}L$, yielding computational requirements scaling as $\mathrm{Re}^{9/4}$ in three dimensions \cite{hunt1991kolmogorov,klewicki2010reynolds}, rendering high-Reynolds-number turbulent flows computationally prohibitive even for moderate Reynolds numbers.

Nonlinear wave equations present different complexities. The focusing nonlinear Schrödinger equation \cite{berezin2012schrodinger}
\begin{equation}
\label{eq:nls}
i\frac{\partial \psi}{\partial t} + \nabla^2 \psi + |\psi|^2\psi = 0
\end{equation}
can develop finite-time singularities through wave collapse, requiring adaptive numerical methods that accurately track solution focusing while preserving conservation laws (mass, momentum, energy) and symplectic structure \cite{sulem2007nonlinear}. Soliton-bearing systems demand schemes that maintain delicate balances between nonlinearity and dispersion over extended temporal evolution. The potential for nonuniqueness of solutions in supercritical regimes necessitates careful initialization and validation against physical constraints.

\subsection{Geometric Complexity and Domain Irregularity}
\label{subsec:geometry}

Real-world applications involve intricate three-dimensional geometries with multiple spatial scales, sharp geometric features, and topological complexity \cite{lim2001accuracy,babuska2012modeling,hairer2006geometric,formaggia2012solving}. Aerospace applications require resolving thin boundary layers of thickness $\delta \sim \mathrm{Re}^{-1/2}L$ around complex aircraft configurations while extending computational domains to far-field boundaries, creating aspect ratios exceeding $10^6$ \cite{martins2022aerodynamic,sobester2014aircraft}. Biomedical simulations involve patient-specific anatomies with thin vessel walls, trabeculated cardiac structures, and fibrous tissue architectures spanning multiple spatial scales \cite{sikkandar2019computational,kainz2018advances,wittek2016finite}. The high aspect ratio apparent in the structure of many biological, industrial, and natural materials has also resulted in a rise of mixed-dimensional PDEs, reducing thin structures to lower dimensional geometries and coupling PDEs across dimensions through compatible interface conditions. Examples of 1d-3d coupled systems include vessel structures in biological tissue~\cite{possenti2021mesoscale}, soil-root systems~\cite{koch2018new} and reinforced solid structures~\cite{steinbrecher2020mortar}, while 2d-3d coupled systems include biomedical membranes~\cite{bociu2021multilayered}, fractured and faulted geophysical media~\cite{berre2019flow} and coated materials~\cite{gurtin1975continuum}. 

Cardiac electrophysiology illustrates these geometric challenges. Electrical activation propagates through anisotropic myocardium according to the monodomain equation:
\begin{equation}
\label{eq:monodomain}
\frac{\partial V}{\partial t} = \nabla \cdot (\mathbf{D}(\mathbf{x}) \nabla V) + I_{\mathrm{ion}}(V, \mathbf{w}) + I_{\mathrm{stim}}(\mathbf{x},t),
\end{equation}
where $V(\mathbf{x},t)$ is transmembrane potential, $\mathbf{D}(\mathbf{x})$ is the spatially-varying diffusion tensor encoding fibrous architecture, $I_{\mathrm{ion}}(V,\mathbf{w})$ represents ionic currents with gating variables $\mathbf{w}$, and $I_{\mathrm{stim}}(\mathbf{x},t)$ is external stimulation. The diffusion tensor exhibits eigenvalue variations exceeding one order of magnitude due to anisotropic fiber orientation. The geometric domain includes thin atrial walls (1--3 mm thickness), complex trabecular structures, and coronary vasculature, features that significantly influence activation patterns and span scales from submillimeter to centimeter scales.

Complex geometries and embedded thin structures create highly irregular domains and extreme aspect ratios that induce stiffness, anisotropy, and stringent multi‑scale resolution requirements, independent of the discretization approach. This degrades conditioning and complicates spatial approximation, making advanced strategies for resolution, stabilization, and scale separation essential and turning geometric complexity into a dominant driver of computational cost and algorithmic difficulty.


\subsection{Discontinuities and Nonsmooth Solution Features}
\label{subsec:discontinuities}

Many physically important PDEs develop discontinuous solutions or sharp gradients even from smooth initial data \cite{jeffrey2020modeling,keyes2013multiphysics,abbasi2025challenges}. Hyperbolic conservation laws give rise to shock waves where characteristics intersect, leading to discontinuities in conserved quantities such as density, velocity, and pressure. Entropy conditions enable selection of physically relevant and unique solutions, compatible with irreversible processes. 
The compressible Euler equations exemplify shock wave formation \cite{serre1999systems,christodoulou2007euler}:
\begin{align}
\label{eq:euler_mass}
\frac{\partial \rho}{\partial t} + \nabla \cdot (\rho \mathbf{u}) &= 0, \\
\label{eq:euler_momentum}
\frac{\partial (\rho \mathbf{u})}{\partial t} + \nabla \cdot (\rho \mathbf{u} \otimes \mathbf{u} + p\mathbf{I}) &= 0, \\
\label{eq:euler_energy}
\frac{\partial E}{\partial t} + \nabla \cdot ((E + p)\mathbf{u}) &= 0,
\end{align}
where $\rho$ is density, $\mathbf{u}$ is velocity, $p$ is pressure, $E$ is total energy, and $\mathbf{I}$ is the identity tensor. These equations conserve mass, momentum, and energy, but admit discontinuous weak solutions that satisfy entropy conditions.

Problem parameters and coefficients may be piecewise constant. Interface problems involve PDEs with discontinuous coefficients across material boundaries \cite{li2006immersed,de2008challenges}. In electromagnetic scattering, Maxwell's equations \cite{huray2009maxwell} must be solved across interfaces between materials with vastly different permittivity and permeability. While electromagnetic fields satisfy controlled jump conditions at interfaces, their spatial derivatives typically exhibit discontinuities, requiring specialized numerical treatments to maintain accuracy and suppress spurious reflections. Free boundary problems add complexity by coupling field evolution to the unknown locations of interfaces \cite{gupta2017classical}, as in phase transitions, solidification, and fluid-structure interaction.

\subsection{Multiscale Temporal and Spatial Phenomena}
\label{subsec:multiscale}

Multiscale PDEs involve physical processes across vastly separated spatial or temporal scales, where fine-scale features exert significant influence on macroscopic behavior \cite{weinan2011principles,peng2021multiscale} -- originating through material contrasts or geometrical features. Direct resolution of all scales requires computational resources scaling with the scale separation ratio, rendering brute-force approaches infeasible for realistic separation ratios. 

Spatially heterogeneous media with rapidly oscillating coefficients exemplify spatial multiscale challenges \cite{vanden2007heterogeneous,huang2015multiscale}:
\begin{equation}
\label{eq:multiscale_elliptic}
\frac{\partial u^\epsilon}{\partial t} = \nabla \cdot \left(a\left(\mathbf{x},\frac{\mathbf{x}}{\epsilon}\right) \nabla u^\epsilon\right),
\end{equation}
where $\epsilon \ll 1$ represents the microscale-to-macroscale ratio and $a(\mathbf{x},\mathbf{y})$ is a periodic or random coefficient capturing microstructure. 

Figure~\ref{fig:multiscale} illustrates the challenge: as the scale-separation ratio $\epsilon$ decreases, the coefficient $a(\mathbf{x}, \mathbf{x}/\epsilon)$ oscillates rapidly (panel~A), the solution $u^\epsilon$ develops fine microstructure visible only at sub-$\epsilon$ scales compared to the smooth homogenized limit $u^*$ (panel~B), and direct finite element discretization exhibits an $\mathcal{O}(\epsilon)$ error floor until $h \ll \epsilon$ --- requiring $\mathcal{O}(\epsilon^{-d})$ degrees of freedom in $d$ dimensions and rendering brute-force simulation intractable (panel~C).

Direct discretization requires mesh spacing $h \ll \epsilon$, yielding $O(\epsilon^{-d})$ degrees of freedom in $d$ dimensions. For composite materials with microscale $\epsilon \sim 10^{-6}$ in three dimensions, this produces computationally intractable problem sizes. Homogenization theory provides rigorous mathematical frameworks for scale separation \cite{pavliotis2008multiscale}, but the practical implementation for geometrically complex microstructures with multiple-scale interactions remains challenging. Temporal multiscale phenomena occur in stiff systems that couple fast and slow dynamics \cite{kuehn2015multiple}. 


\begin{figure}[t]
    \centering
    \includegraphics[width=0.95\textwidth]{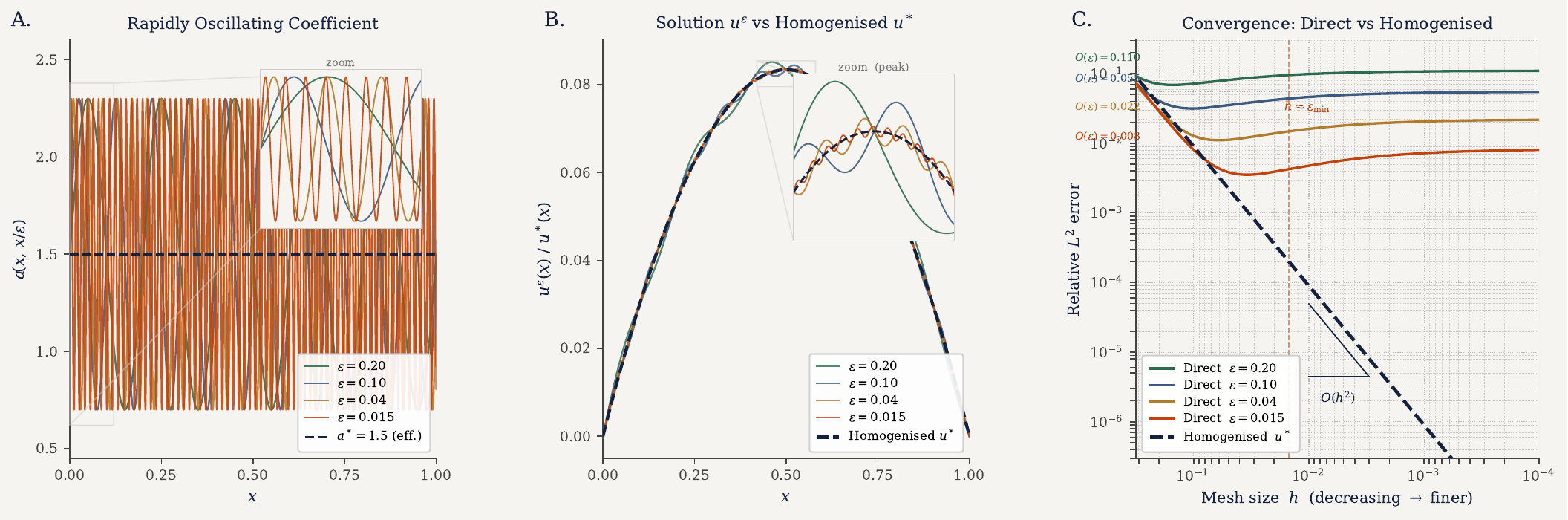}
    \caption{Illustration of the multiscale challenge for PDEs in spatially heterogeneous media with rapidly oscillating coefficient $a(x, x/\epsilon)$. Shown across four scale-separation ratios $\epsilon \in \{0.20, 0.10, 0.04, 0.015\}$: 
    \textbf{(A)}~Coefficient $a(x, x/\epsilon) = 1.5 + 0.8\sin(2\pi x/\epsilon)$ for decreasing $\epsilon$. The dashed line indicates the homogenized value $a^* = 1.5$. Inset: magnification near $x=0$ showing increasingly rapid oscillations that direct discretization must resolve.
    \textbf{(B)}~Solutions $u^\epsilon(x)$ (solid) compared to the smooth homogenized limit $u^*(x)$ (dashed). Inset: zoom at the peak ($x \approx 0.5$) revealing that oscillatory corrections decay with $\epsilon$ but remain visible on the microscale, even where the macroscopic profiles are nearly indistinguishable.
    \textbf{(C)}~Relative $L^2$ error versus mesh size $h$ (decreasing left to right) for direct finite element discretization and the homogenized solver. Direct simulation shows an $\mathcal{O}(\epsilon)$ error plateau for $h \gg \epsilon$, with second-order $\mathcal{O}(h^2)$ convergence only when $h \ll \epsilon$ — a resolution demand that becomes computationally intractable for small $\epsilon$. The homogenized approach yields uniform $\mathcal{O}(h^2)$ convergence independent of $\epsilon$ at greatly reduced cost.}
    \label{fig:multiscale}
    \label{fig:multiscale}
\end{figure}

\subsection{Multiphysics Coupling and Cross-Field Compatibility}
\label{subsec:multiphysics}

Multiphysics problems couple multiple physical processes governed by different PDEs, often with distinct mathematical character (elliptic, parabolic, hyperbolic) and disparate temporal or spatial scales \cite{keyes2013multiphysics}. The coupling introduces additional challenges beyond those inherent in individual field equations: Function spaces must align with the PDE's variational, differential and potential Hilbert‑complex (de-Rham) structure, which underlies the cross-field compatibility and hence the well‑posedness of the coupled problem~\cite{boffi2013mixed,arnold2018finite}.

The prototypical example is fluid-structure interaction, coupling incompressible Navier-Stokes equations \eqref{eq:navier_stokes_momentum}--\eqref{eq:navier_stokes_continuity} with elastodynamics:
\begin{equation}
\label{eq:elastodynamics}
\rho_s \frac{\partial^2 \mathbf{d}}{\partial t^2} = \nabla \cdot \boldsymbol{\sigma}(\mathbf{d}) + \mathbf{f}_s,
\end{equation}
where $\mathbf{d}(\mathbf{x},t)$ is structural displacement, $\rho_s$ is structural density, $\boldsymbol{\sigma}$ is the stress tensor (typically depending nonlinearly on displacement through constitutive relations), and $\mathbf{f}_s$ represents body forces. At the fluid-structure interface $\Gamma(t)$, kinematic constraints (velocity continuity) and dynamic conditions (traction balance) must be satisfied:
\begin{align}
\label{eq:fsi_kinematic}
\mathbf{u}|_{\Gamma} &= \frac{\partial \mathbf{d}}{\partial t}\Big|_{\Gamma}, \\
\label{eq:fsi_dynamic}
\boldsymbol{\sigma}_f \cdot \mathbf{n}|_{\Gamma} &= \boldsymbol{\sigma}_s \cdot \mathbf{n}|_{\Gamma},
\end{align}
where $\boldsymbol{\sigma}_f$ is the fluid stress, $\boldsymbol{\sigma}_s$ is the structural stress, and $\mathbf{n}$ is the interface normal.

A fundamental challenge in mixed problems arises from the need for \emph{inf-sup stability} (also called Ladyzhenskaya-Babuška-Brezzi or LBB stability) \cite{boffi2013mixed, bathe2001inf,bochev2004inf}, resulting in compatibility conditions function spaces and restricting which solution methodology can be combined to solve the coupled problem. For the Stokes problem (steady, zero-Reynolds-number limit of Navier-Stokes):
\begin{align}
\label{eq:stokes_momentum}
-\nabla \cdot \boldsymbol{\sigma}(\mathbf{u},p) &= \mathbf{f}, \\
\label{eq:stokes_continuity}
\nabla \cdot \mathbf{u} &= 0,
\end{align}
where $\boldsymbol{\sigma}(\mathbf{u},p) = -p\mathbf{I} + \mu(\nabla\mathbf{u} + \nabla\mathbf{u}^T)$ with viscosity $\mu$, the velocity and pressure spaces must satisfy the inf-sup condition:
\begin{equation}
\label{eq:infsup}
\inf_{q \in Q} \sup_{\mathbf{v} \in V} \frac{(\nabla \cdot \mathbf{v}, q)}{|\mathbf{v}|_V |q|_Q} \geq \beta > 0,
\end{equation}
where $V$ and $Q$ are appropriate function spaces for velocity and pressure, and $\beta$ is a positive constant, ideally independent of any material constants. Violating this condition leads to non-unique solutions and in practical cases spurious pressure modes. Similar stability conditions arise in elasticity (Stokes equations with symmetric gradients), electromagnetics (mixed formulations of Maxwell's equations), poroelasticity (coupled elasticity and porous media flow) and other mixed-field coupled systems.


Coupled problems are inherently challenging. The coupling introduces stricter stability and compatibility constraints than the subproblems themselves and does not reduce to assembling effective single-physics methods~\cite{keyes2013multiphysics}.
Moveover, the total dimensionality in multiphysics systems is determined not solely by physical spatial dimension but by the product of spatial dimension and the number of coupled fields. A three-dimensional thermo-mechanical-electromagnetic problem effectively operates in a much higher-dimensional space than suggested by geometry alone, reintroducing aspects of the curse of dimensionality through field coupling rather than spatial dimension.

\subsection{Concurrent Challenges and Compound Difficulty}
\label{subsec:concurrent_challenges}

As Figure~\ref{fig:pde_challenges} illustrates, the challenges enumerated in Sections~\ref{subsec:dimensionality}--\ref{subsec:multiphysics} rarely occur in isolation; realistic applications frequently involve multiple challenges simultaneously, with interactions that amplify computational complexity beyond simple superposition. Understanding these interactions is crucial for developing effective solution strategies and assessing method applicability.

Turbulent combustion exemplifies concurrent challenge presence: high Reynolds numbers create turbulent flow (nonlinearity, Section~\ref{subsec:nonlinearity}) with multiscale structure from integral scales to Kolmogorov scales (multiscale phenomena, Section~\ref{subsec:multiscale}); chemical reactions introduce additional stiffness across temporal scales and high-dimensional composition spaces (dimensionality, Section~\ref{subsec:dimensionality}); flame fronts form thin interfaces with steep gradients approaching discontinuities (discontinuities, Section~\ref{subsec:discontinuities}); realistic configurations involve complex combustor geometries (geometric complexity, Section~\ref{subsec:geometry}); and the coupling between fluid dynamics, heat transfer, and chemical kinetics constitutes a multiphysics system (Section~\ref{subsec:multiphysics}).

Climate modeling similarly involves concurrent challenges: atmospheric and oceanic flows exhibit turbulent multiscale dynamics (multiscale phenomena, Section~\ref{subsec:multiscale}); cloud microphysics couples fluid mechanics with phase transitions and radiative transfer (coupled problems, Section~\ref{subsec:multiphysics}); realistic geography introduces complex terrain and coastline geometry (geometric complexity, Section~\ref{subsec:geometry}); and the global domain combined with numerous chemical species and physical processes creates effective high dimensionality (dimensionality, Section~\ref{subsec:dimensionality}). The temporal range from seconds (convection) to centuries (climate change) introduces additional multiscale complexity (multiscale phenomena, Section~\ref{subsec:multiscale}).

These interactions create emergent difficulties not captured by analyzing individual challenges separately. Geometric complexity can amplify multiscale effects by creating local regions of extreme scale separation. Nonlinearity can generate high-dimensional chaotic dynamics even in low-dimensional spatial domains. Discontinuities interact with geometric features to produce complex wave patterns. Multiphysics coupling can introduce new solution regimes that are absent in individual subsystems.

Methodological implications of concurrent challenges include: (1) no single numerical method excels across all challenge combinations, and hybrid approaches combining multiple techniques become necessary; (2) solution verification becomes significantly more difficult when multiple challenges interact, as individual verification techniques may be inadequate; (3) machine learning approaches might offer potential advantages in implicitly discovering low-dimensional structure in high-dimensional multiphysics systems, but face difficulties with discontinuities and lack rigorous error bounds; (4) classical methods provide mathematical guarantees but struggle with geometric complexity and high dimensionality. These aspects will be discussed in detail in the following sections (Sections \ref{sec:classicmethods}-\ref{subsec:ml_challenges}).

\section{Classical Numerical Methods for PDE Solution}
\label{sec:classicmethods}

The computational solution of PDEs has evolved through decades of mathematical innovation, producing a sophisticated ecosystem of numerical methods with distinct capabilities and fundamental trade-offs. The six challenges identified in Section~\ref{sec:challenges}, namely, high dimensionality, nonlinearity, geometric complexity, discontinuities, multiscale phenomena, and multiphysics coupling, have driven the development of specialized solution strategies, each exploiting distinct mathematical structures and making different trade-offs among accuracy, efficiency, and applicability. This section examines classical approaches that form the foundation of modern computational science, establishing the baseline of capabilities against which any new methodology must be evaluated. Table~\ref{tab:classical-pde-methods} provides a comprehensive quantitative comparison of the methods discussed in this section, including computational complexity classes, adaptivity capabilities, theoretical guarantees, and distinguishing characteristics that guide method selection for specific problem types. Figure~\ref{fig:discretisation_paradigms} provides a side-by-side comparison of the four paradigms on a common non-trivial domain, illustrating the fundamental differences in how each method partitions the computational space, enforces the governing equations, and handles geometric boundaries — differences that directly determine each method's strengths and limitations against the challenges of Section~\ref{sec:challenges}.

We organize this presentation around four major discretization paradigms: finite difference methods (Section~\ref{subsec:fdm}) approximate derivatives on structured grids, achieving simplicity and computational efficiency at the cost of geometric flexibility; finite element methods (Section~\ref{subsec:fem}) employ variational formulations on unstructured meshes, providing geometric adaptability and rigorous error analysis; finite volume methods (Section~\ref{subsec:fvm}) discretize conservation laws directly, ensuring exact conservation properties critical for fluid dynamics; and spectral methods (Section~\ref{subsec:spectral}) expand solutions in global basis functions, delivering exponential convergence for smooth problems. Cross-cutting computational strategies that enhance multiple base methods, adaptive mesh refinement, multigrid solvers, and domain decomposition, are examined in Section~\ref{subsec:crosscutting}. Finally, Section~\ref{subsec:meshless} briefly discusses meshless and boundary element approaches that eliminate classical mesh connectivity requirements.

%

\begin{figure}[t]
    \centering
    \includegraphics[width=\textwidth]{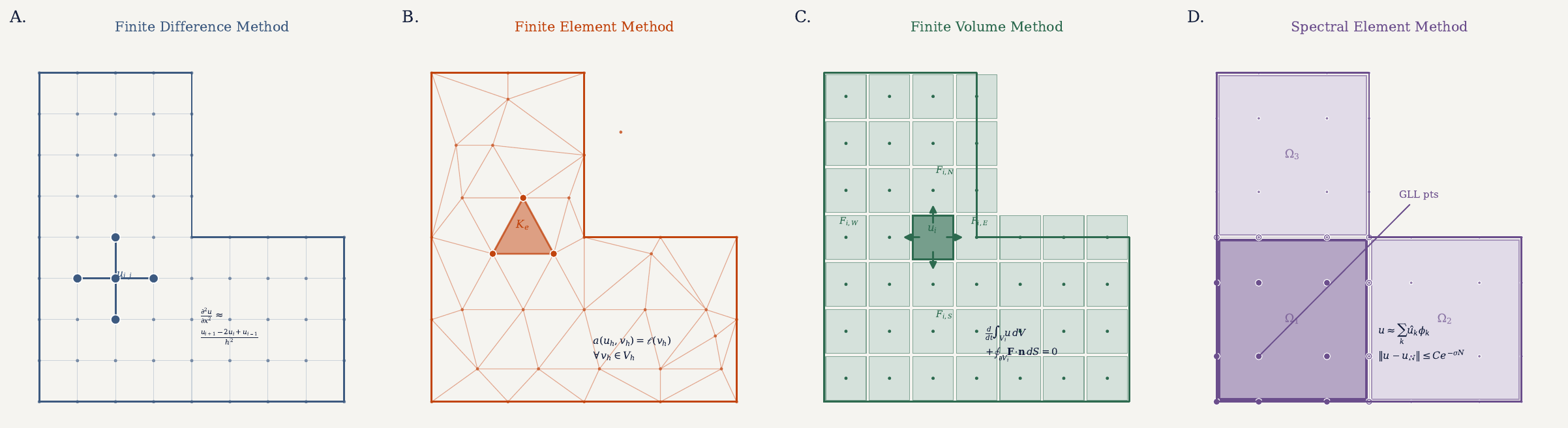}
    \caption{Illustration of four principal discretization paradigms on a common L-shaped domain, highlighting differences in geometric flexibility and approximation properties.
    \textbf{(A)} Finite difference method (FDM): structured Cartesian grid with the canonical five-point stencil emphasized. It shows the central-difference approximation $\partial^2 u / \partial x^2 \approx (u_{i+1} - 2u_i + u_{i-1})/h^2$ at the stencil center.
    \textbf{(B)} Finite element method (FEM): unstructured triangulation with one representative element $K_e$ highlighted. The global system is governed by the weak form $a(u_h, v_h) = \ell(v_h)$ for all test functions $v_h \in V_h$.
    \textbf{(C)} Finite volume method (FVM): structured control-volume partition enforcing local conservation. The integral form $\frac{d}{dt} \int_{V_i} u \, dV + \oint_{\partial V_i} \mathbf{F} \cdot \mathbf{n} \, dS = 0$ is applied cell-wise; flux arrows indicate the anti-symmetric condition $F_{ij} = -F_{ji}$.
    \textbf{(D)} Spectral element method (SEM): three macro-elements $\Omega_1$--$\Omega_3$ with Gauss--Lobatto--Legendre (GLL) quadrature points displayed. For smooth solutions, exponential convergence $\|u - u_N\| \leq C e^{-\sigma N}$ holds within each element.}
    \label{fig:discretisation_paradigms}
    \label{fig:discretisation_paradigms}
\end{figure}

\subsection{Finite Difference Methods (FDMs)}
\label{subsec:fdm}

\subsubsection{Mathematical Foundation}

Finite difference methods discretize PDEs by approximating derivatives through difference quotients on structured grids \cite{thomas2013numerical,mazumder2015numerical}, transforming continuous differential operators into algebraic systems. For a smooth function $u(x,t)$, standard approximations take the form:
\begin{align}
\label{eq:fdm_first}
\frac{\partial u}{\partial x}\bigg|_{i,n} &\approx \frac{u_{i+1}^n - u_{i-1}^n}{2\Delta x} + O((\Delta x)^2) \quad \text{(central difference)}, \\
\label{eq:fdm_second}
\frac{\partial^2 u}{\partial x^2}\bigg|_{i,n} &\approx \frac{u_{i+1}^n - 2u_i^n + u_{i-1}^n}{(\Delta x)^2} + O((\Delta x)^2),
\end{align}
where $u_i^n$ denotes the discrete solution at spatial index $i$ and time level $n$. This conceptual directness, combined with efficient memory access patterns on structured grids and straightforward vectorization on modern architectures, sustains FDM's continued relevance despite limitations in geometric flexibility.

\subsubsection{Advanced Schemes and Contemporary Developments~\label{sec:traditional-methods:advanced-schemes}}

Compact finite difference schemes achieve spectral-like resolution through implicit derivative relations \cite{lele1992compact}, using narrower stencils than explicit high-order schemes while maintaining superior accuracy per grid point. Recent work by Ramos et al.~\cite{ramos2024compact} developed tri-diagonal schemes with enhanced resolution characteristics for Burgers' equation, while Salian and Appadu~\cite{salian2024novel} introduced central compact formulations achieving tenth-order accuracy for third derivatives in dispersive wave equations. The systematic framework of summation-by-parts (SBP) operators with weak boundary conditions enables provably stable high-order schemes through discrete analogues of integration by parts \cite{svard2014review}, addressing long-standing stability concerns in boundary treatment.

For problems involving discontinuities (Section~\ref{subsec:discontinuities}), weighted essentially non-oscillatory (WENO) schemes balance high-order accuracy with robust shock-capturing through adaptive stencil weighting \cite{shu2020essentially}. The WENO-NZ variant \cite{han2023improved} recovers optimal fifth-order convergence at critical points, resolving a persistent challenge for earlier formulations. On unstructured meshes, adaptive-order WENO-AO schemes \cite{balsara2016efficient} achieve smooth fifth-to-third-order transitions near discontinuities. The ALW-MR-WENO formulation \cite{lin2025increasingly} delivers significant CPU time savings through optimized two-stencil reconstructions, demonstrating that algorithmic refinement continues to yield substantial performance improvements.

\subsubsection{Computational Characteristics and Domain of Effectiveness}

Finite difference methods excel in computational efficiency for problems on regular domains. Structured grid topology enables cache-optimal memory access patterns and straightforward SIMD (Single Instruction, Multiple Data) vectorization on modern processors. These properties make FDM particularly well-suited for geometric multigrid acceleration \cite{hackbusch1978multi}, achieving optimal $O(N)$ solution complexity for elliptic problems. Compact schemes approach spectral accuracy while preserving algorithmic simplicity, making them essential for direct numerical simulation of turbulence \cite{lele1992compact} and large-eddy simulation, where resolution per degree of freedom directly determines computational feasibility. WENO schemes maintain high-order accuracy in smooth flow regions while suppressing spurious oscillations near shocks \cite{shu2020essentially}, making them indispensable for compressible flows with complex shock structures.

The fundamental limitation of FDM is geometric inflexibility. Complex geometries require coordinate transformations that introduce metric terms, significantly complicating implementation and potentially degrading accuracy near curved boundaries. While immersed boundary methods \cite{fadlun2000combined} partially address this by embedding irregular boundaries in Cartesian grids, they cannot match the geometric adaptability of unstructured mesh methods. Higher-order schemes require wide computational stencils—tenth-order accuracy typically demands 11-point stencils—which complicate boundary treatment and parallel domain decomposition. For multiscale problems (Section~\ref{subsec:multiscale}), uniform grid refinement to resolve the finest scales becomes prohibitively expensive, motivating the development of adaptive mesh refinement (Section~\ref{subsec:crosscutting}).

Representative software implementations include finite difference modules in PETSc \cite{balay2019petsc} for structured grid problems and specialized codes like Dedalus \cite{burns2020dedalus} combining finite differences with spectral methods for atmospheric and astrophysical applications.

\subsection{Finite Element Methods (FEMs)}
\label{subsec:fem}

\subsubsection{Mathematical Foundation and Variational Framework}

The finite element method reformulates PDEs through variational principles \cite{fenner2013finite,zienkiewicz2005finite}, seeking approximate solutions $u_h \in V_h \subset V$ satisfying the weak form:
\begin{equation}
\label{eq:fem_weak}
a(u_h,v_h) = \ell(v_h) \quad \forall v_h \in V_h,
\end{equation}
where the bilinear form $a(\cdot,\cdot)$ and linear functional $\ell(\cdot)$ arise from integration by parts of the strong PDE formulation, and $V_h = \text{span}\{\phi_1, \ldots, \phi_N\}$ is constructed from piecewise polynomial basis functions. For nonlinear PDEs, the bilinear form generalizes to a nonlinear form $a(u_h; u_h, v_h)$. This variational foundation provides systematic a priori error analysis through Céa's lemma:
\begin{equation}
\label{eq:cea}
\|u - u_h\|_V \leq C \inf_{v_h \in V_h} \|u - v_h\|_V,
\end{equation}
establishing that FEM approximation quality is bounded by the best approximation in the discrete space, with a constant $C$ depending on the stability properties of the continuous problem.

\subsubsection{Adaptive Strategies and Convergence Theory}

A posteriori error estimation enables both guaranteed global error control and efficient adaptive refinement \cite{becker2001optimal,repin2008posteriori}. By balancing errors from discretization, iterative solvers, numerical quadrature, and regularization, adaptive stopping criteria avoid unnecessary computation \cite{fevotte2024adaptive}. The finite element framework offers two independent refinement mechanisms that address different aspects of solution complexity.

The $h$-version refines mesh geometry while maintaining fixed polynomial degree, achieving algebraic convergence $\|u - u_h\|_{L^2} = O(h^{p+1})$ for smooth solutions of regularity $u \in H^{p+1}$ \cite{ern2021finite}. This strategy effectively resolves localized features but faces diminishing returns for smooth solutions. The $p$-version increases polynomial degree on fixed mesh topology, delivering exponential convergence $\|u - u_p\|_E \leq C e^{-bp^{\alpha}}$ when solution regularity permits \cite{morin2002convergence}, with $\alpha$ depending on problem-specific smoothness. The $hp$-version combines both strategies optimally: geometric refinement addresses singularities while polynomial enrichment exploits smoothness, achieving $\|u - u_{hp}\|_E \leq C e^{-bN^{1/3}}$ even for problems containing singularities \cite{bangerth2003adaptive}, where $N$ denotes total degrees of freedom.

Goal-oriented adaptive finite element methods (GOAFEM) enable high-accuracy approximation of localized target functionals, such as point stresses or integrated fluxes—without requiring uniform global refinement \cite{oden2001goal}. Recent work by Becker et al.~\cite{becker2023goal} established optimal computational complexity for GOAFEM, proving that adaptive algorithms with contractive iterative solvers achieve optimal convergence rates in total computational cost, settling a fundamental theoretical question.

\subsubsection{Extensions and Specialized Formulations}

The variational framework enables systematic extensions addressing specific application requirements. Mixed finite element methods introduce multiple field variables, which are essential for problems that require the simultaneous approximation of primal and dual quantities. For incompressible Stokes flow:
\begin{align}
\label{eq:stokes_mixed_1}
a(\mathbf{u}_h, \mathbf{v}_h) + b(\mathbf{v}_h, p_h) &= \ell(\mathbf{v}_h) \quad \forall \mathbf{v}_h \in \mathbf{V}_h, \\
\label{eq:stokes_mixed_2}
b(\mathbf{u}_h, q_h) &= 0 \quad \forall q_h \in Q_h,
\end{align}
where velocity space $\mathbf{V}_h$ and pressure space $Q_h$ must satisfy the inf-sup stability condition (Section~\ref{subsec:multiphysics}) to ensure unique, stable pressure approximation \cite{boffi2013mixed}. Classical stable element pairs include Taylor-Hood ($P_2 \times P_1$: quadratic velocity, linear pressure), MINI element ($P_1^+ \ times P_1$: low-order FEM with enriched element-wise cubic bubbles), and Raviart-Thomas ($RT_0 \times P_0$: lowest-order mixed formulation). Stabilized finite element methods—including streamline upwind/Petrov-Galerkin (SUPG) \cite{tezduyar1991stabilized} and variational multiscale methods \cite{bazilevs2007variational}—circumvent inf-sup restrictions through carefully designed residual-based stabilization terms. When writing PDEs in mixed form by coupling first order equations similar saddle-point structures appear. Problem-specific developments have resulted in locking-free finite element pairs for elasticity~\cite{arnold2007mixed,yi2022locking}, stable discretization of electromagnetism rooted in discrete differential forms~\cite{hiptmair2002finite}.

Discontinuous Galerkin (DG) methods employ element-wise discontinuous approximations coupled through numerical fluxes at interfaces:
\begin{equation}
\label{eq:dg}
\sum_K \left[ \int_K \mathcal{L}u_h v_h \, dx + \int_{\partial K} \hat{f}(u_h^+, u_h^-) \cdot \mathbf{n} v_h \, ds \right] = \int_\Omega f v_h \, dx,
\end{equation}
where $\mathcal{L}$ denotes the differential operator, $K$ ranges over mesh elements, $\hat{f}$ is the numerical flux depending on traces $u_h^{\pm}$ from adjacent elements, and $\mathbf{n}$ is the outward normal. This hybrid approach combines finite element geometric flexibility with finite volume conservation properties. Recent stability analysis \cite{gazca2025stability} established uniform $L^\infty(L^2)$ bounds for DG applied to non-Newtonian flows with power-law rheology. Asynchronous DG variants \cite{nowick2015asynchronous} introduce asynchrony-tolerant flux formulations that eliminate synchronization bottlenecks on massively parallel systems. The highly localized inter-element communication and arithmetic intensity of higher-order DG methods make them particularly well-suited for exploiting contemporary exascale architectures \cite{bastian2020exa}.

Isogeometric analysis (IGA) employs NURBS (Non-Uniform Rational B-Splines) basis functions directly from CAD geometry representations \cite{hughes2005isogeometric}, eliminating geometric approximation errors and providing higher-order continuity than standard finite elements. For applications where geometry processing dominates the workflow, particularly in shape optimization \cite{wall2008isogeometric}, IGA enables direct progression from design to simulation without intermediate meshing.

Space-time finite element methods extend the variational framework from spatial to temporal domains, introducing shape functions and integration over combined space-time elements \cite{tezduyar1992new}. This approach provides simultaneous control of spatial and temporal discretization errors and naturally handles moving domain problems \cite{masud1997space}. Problem-specific enrichment of discrete spaces $V_h$ enables targeted resolution of localized features: the eXtended Finite Element Method (XFEM) \cite{fries2010extended,flemisch2016review} incorporates discontinuous or singular functions for crack propagation and interface problems, while the Enhanced Galerkin method blends standard FEM with DG enrichment for local conservation \cite{lee2016locally}.

The Virtual Element Method (VEM) represents a paradigm shift by enabling general polygonal and polyhedral elements while preserving rigorous approximation theory \cite{beirao2013basic}. VEM employs virtual shape functions that are not explicitly constructed but whose action on polynomial spaces is fully characterized through projection operators $\Pi_k^{\nabla}$. Element-wise formulations take the form:
\begin{equation}
\label{eq:vem}
\sum_{K}
\left[
\int_K \nabla \Pi_k^{\nabla} u_h \cdot \nabla \Pi_k^{\nabla} v_h \, dx
\;+\;
S^K\!\big((I-\Pi_k^{\nabla})u_h,(I-\Pi_k^{\nabla})v_h\big)
\right]
=
\sum_{K} \int_K f \, \Pi_k^{0} v_h \, dx,
\end{equation}
where $S^K$ denotes a problem-dependent stabilization that ensures coercivity. This flexibility proves particularly valuable for problems on complex polyhedral meshes and for adaptive refinement without conformity constraints.

\subsubsection{Computational Characteristics and Domain of Effectiveness}

Finite element methods excel at geometric complexity (Section~\ref{subsec:geometry}) through unstructured mesh capability, naturally representing domains with intricate boundaries, internal interfaces, and curved features. The variational framework ensures systematic error estimation and optimal approximation properties—even for curved domains when employing suitable parametrizations \cite{bernardi1989optimal}. Local basis support yields sparse stiffness matrices amenable to efficient iterative solvers \cite{saad2003iterative}. Higher-order $p$ and $hp$ variants achieve exponential convergence for smooth solutions, dramatically reducing problem size compared to low-order methods. Mixed formulations provide direct access to derived quantities (fluxes, stresses) with provable accuracy, while DG methods offer exceptional parallel scalability and simplified $hp$-adaptivity without conformity constraints.

The mature software ecosystem—including FEniCS \cite{alnaes2015fenics}, deal.II \cite{bangerth2007deal}, DUNE \cite{bastian2021dune}, MFEM \cite{anderson2021mfem}, and PETSc \cite{balay2019petsc}—provides production-quality implementations with adaptive refinement, parallel scaling to hundreds of thousands of cores, and multiphysics coupling capabilities. This ecosystem represents decades of software engineering investment and extensive validation across diverse applications.

Fundamental limitations include mesh generation complexity for three-dimensional domains, often requiring manual intervention and geometric simplification for highly intricate configurations. The assembly process introduces computational overhead compared to matrix-free approaches, particularly impacting explicit time-stepping schemes. Higher-order methods incur increased cost per degree of freedom and face conditioning challenges limiting practical polynomial degrees to approximately $p \leq 10$ for most problems. For discontinuous solutions (Section~\ref{subsec:discontinuities}), standard continuous finite elements produce spurious oscillations requiring specialized stabilization or DG formulations. The relative computational costs, accuracy characteristics, and implementation requirements of these FEM variants, compared with other discretization paradigms, are summarized in Table~\ref{tab:classical-pde-methods}.

\subsection{Finite Volume Methods (FVMs)}
\label{subsec:fvm}

\subsubsection{Mathematical Foundation}

Finite volume methods discretize the integral form of conservation laws over control volumes $V_i$ \cite{leveque2002finite,mazumder2015numerical}:
\begin{equation}
\label{eq:fvm_integral}
\frac{d}{dt} \int_{V_i} u \, dV + \int_{\partial V_i} \mathbf{F}(u) \cdot \mathbf{n} \, dS = \int_{V_i} S(u) \, dV,
\end{equation}
where $u$ represents conserved quantities, $\mathbf{F}(u)$ is the flux function, $\mathbf{n}$ denotes outward normal, and $S(u)$ represents source terms. Applying the divergence theorem and approximating integrals yields the semi-discrete system:
\begin{equation}
\label{eq:fvm_discrete}
V_i \frac{du_i}{dt} + \sum_{j \in \mathcal{N}(i)} A_{ij} F_{ij} = V_i S_i,
\end{equation}
where $u_i$ represents cell-averaged values, $\mathcal{N}(i)$ denotes neighboring cells sharing interfaces with cell $i$, $A_{ij}$ is interface area, and $F_{ij} = -F_{ji}$ denotes numerical flux ensuring exact discrete conservation. This integral formulation guarantees local conservation over arbitrary control volumes, regardless of mesh quality or solution regularity—a property critical for fluid dynamics and reactive transport where spurious mass sources or sinks corrupt long-time evolution.

\subsubsection{Discretization Strategies and Recent Advances}

Cell-centered schemes naturally associate primary unknowns with cell centroids, with control volumes coinciding with mesh cells. Diffusive fluxes arising from gradient terms require careful interface treatment. The two-point flux approximation (TPFA) \cite{eymard2000finite} computes interface fluxes using values from adjacent cells only, achieving second-order accuracy on orthogonal meshes. For anisotropic diffusion tensors misaligned with mesh topology, multi-point flux approximations (MPFA) \cite{aavatsmark2002introduction} employ values from cell neighborhoods to maintain consistency, though at increased computational cost and complexity. Linear MPFA schemes can exhibit non-monotone behavior \cite{nordbotten2007monotonicity}; nonlinear variants \cite{terekhov2017cell,schneider2017monotone} restore monotonicity through flux limiters at the cost of introducing nonlinearity.

Advective fluxes employ upwind-biased formulations that naturally stabilize hyperbolic transport. For problems coupling multiple flux types—diffusion, advection, and reaction—careful flux discretization proves essential for robustness and efficiency of the overall nonlinear solution process \cite{hamon2018implicit}. High-order reconstruction within cells combined with flux limiters enables WENO-type schemes in finite volume frameworks, maintaining sharp shock capture while achieving high-order accuracy in smooth regions.

Vertex-centered formulations construct dual control volumes around mesh nodes, with flux computations using stencils analogous to those in finite element methods \cite{eymard2000finite}. This approach ensures compatibility with finite element codes, facilitating coupling of elliptic and hyperbolic equations \cite{eymard2000finite}, interface-coupled fluid-structure interaction \cite{slone2002dynamic}, and domain-coupled poroelasticity \cite{prevost2014two}. Adaptivity concepts developed for FEM transfer naturally to vertex-centered FVM frameworks \cite{vohralik2008residual,erath2016adaptive}.

While originally developed for scalar conservation laws in fluid mechanics, FVM has been successfully extended to vector-valued systems, including linear elasticity \cite{cardiff2021thirty,nordbotten2025two}, demonstrating the versatility of the conservation-law discretization paradigm. Comprehensive stability and convergence theory is well-established \cite{eymard2000finite}, with convergence rates derivable in specific cases through equivalence between FVM and mixed finite element methods with numerical quadrature \cite{arbogast1997mixed}.

\subsubsection{Computational Characteristics and Domain of Effectiveness}

Finite volume methods maintain exact local conservation, fluxes leaving one cell precisely balance fluxes entering neighbors, ensuring global conservation to machine precision independent of discretization quality. This property proves essential for long-time integration, where accumulation of conservation errors corrupts physical solutions. The integral formulation naturally handles discontinuous solutions and material interfaces (Section~\ref{subsec:discontinuities}), making FVM the method of choice for shock-dominated compressible flows. Geometric flexibility through unstructured polyhedral meshes enables complex domain treatment while preserving conservation. Compact computational stencils—typically involving only immediate neighbors—facilitate parallelization and adaptive refinement. Upwind flux formulations provide inherent stabilization for hyperbolic problems without introducing artificial dissipation parameters requiring tuning.

Achieving high-order accuracy while maintaining monotonicity near extrema remains challenging, particularly for multidimensional problems where genuinely multidimensional reconstruction becomes substantially more complex than dimension-by-dimension approaches \cite{titarev2004finite,colella2011high}. Cell-centered schemes may exhibit reduced accuracy on highly skewed meshes where cell-averaged values diverge significantly from cell-center point values. Vertex-centered approaches introduce complexity in flux computation and gradient reconstruction, often requiring larger stencils than cell-centered counterparts. Both formulations can suffer checkerboard pressure-velocity decoupling on certain mesh configurations without appropriate co-located or staggered variable arrangements.

Representative implementations include OpenFOAM \cite{jasak2007openfoam} for general CFD applications, CLAWPACK \cite{mandli2016clawpack} for hyperbolic conservation laws with adaptive refinement, and specialized codes for reservoir simulation and subsurface flow.

\subsection{Spectral Methods}
\label{subsec:spectral}

\subsubsection{Mathematical Foundation}

Spectral methods expand solutions in global, orthogonal, infinitely differentiable basis functions \cite{guo1998spectral,boyd2001chebyshev}:
\begin{equation}
\label{eq:spectral_expansion}
u(x,t) \approx \sum_{k=0}^{N} \hat{u}_k(t) \phi_k(x),
\end{equation}
where basis functions $\phi_k$ are Fourier modes for periodic problems or orthogonal polynomials (Chebyshev, Legendre) for non-periodic domains. Substituting this expansion into the PDE and enforcing residual orthogonality (Galerkin approach) or satisfaction at collocation points yields systems of ordinary differential equations for expansion coefficients $\hat{u}_k(t)$. The terminology "Fourier spectral method" or "Chebyshev spectral method" specifies the basis function choice.

For smooth, analytic solutions, spectral methods deliver exponential (spectral) convergence: $\|u - u_N\|_\infty \leq C e^{-\sigma N}$ for appropriate constants $C$ and $\sigma$ depending on solution analyticity. Fast transform algorithms—Fast Fourier Transform (FFT) for Fourier bases, Fast Chebyshev Transform for polynomial bases—achieve $O(N \log N)$ complexity for derivative evaluation \cite{elliott1983fast,brigham1988fast}, making spectral methods computationally efficient per degree of freedom for smooth problems. Collocation formulations reduce spatial discretization to systems of ordinary differential equations, enabling straightforward time integration.

\subsubsection{Spectral Element Methods: Combining Spectral Accuracy with Domain Decomposition}

The Spectral Element Method (SEM) combines spectral method accuracy with finite element geometric flexibility through domain decomposition \cite{patera1984spectral,van1996spectral}. Within each element, high-order Lagrange polynomials defined on Gauss-Lobatto-Legendre (GLL) quadrature points achieve spectral convergence while maintaining only $C^0$ inter-element continuity. The tensor-product structure of higher-order bases enables matrix-free implementations, crucial for large-scale applications, thereby avoiding the explicit storage of assembled matrices.

SEM proves particularly well-suited for high-performance computing architectures. NekRS \cite{fischer2022nekrs}, a GPU-accelerated spectral element solver, enables exascale simulations of incompressible Navier-Stokes equations \cite{merzari2024energy}. Capabilities include $k$-th order accurate time-split formulations for temporal accuracy, nonconforming mesh connectivity via overset grids for geometric flexibility, and native GPU Lagrangian particle tracking for multiphase flows. Applications span atmospheric boundary layer turbulence to nuclear reactor thermal hydraulics, demonstrating spectral methods' continued relevance at contemporary computing frontiers. Recent extensions address nonlinear and nonlocal PDEs on complex two-dimensional domains by efficiently evaluating convolution integrals \cite{roden2025multishape}.

Spectral methods find primary application in wave propagation on shallow water surfaces \cite{han2021numerical} and general relativistic spacetime evolution \cite{chung2024spectral}, where minimal numerical dispersion and dissipation prove essential for long-time accuracy.

\subsubsection{Computational Characteristics and Domain of Effectiveness}

Spectral methods deliver exponential convergence for smooth problems (Section~\ref{subsec:nonlinearity}), achieving accuracy levels unattainable by algebraically convergent methods. Minimal numerical dispersion and dissipation make them ideal for long-time wave propagation where accumulated phase errors corrupt solutions. Fast transform algorithms achieve computational efficiency competitive with that of low-order methods despite their high formal accuracy. For periodic geometries, pure spectral methods achieve optimal performance with minimal implementation complexity \cite{spalart1991spectral}. Spectral element decomposition extends these advantages to complex domains while preserving high-order accuracy.

Fundamental limitations restrict applicability. The Gibbs phenomenon severely degrades accuracy for problems with discontinuities or sharp gradients (Section~\ref{subsec:discontinuities}), producing global oscillations that corrupt the entire solution domain \cite{gottlieb1997gibbs}. Time-step restrictions for explicit integration become severe due to eigenvalue clustering from high-order spatial discretization \cite{hesthaven2007spectral}, with stability constraints degrading as CFL number increases. For non-periodic problems, global basis functions yield dense differentiation matrices and ill-conditioned linear systems, compromising efficiency compared to sparse local methods. Boundary condition implementation becomes complex, particularly for mixed or time-dependent constraints. The smoothness requirement makes classical spectral methods unsuitable for shocks, material interfaces, or non-smooth solutions without sophisticated filtering techniques \cite{hesthaven2007spectral}. Spectral element decomposition mitigates some limitations but introduces inter-element continuity constraints.

\subsection{Adaptive, Multilevel, and Multiscale Strategies}
\label{subsec:crosscutting}

The numerical methods described in Sections~\ref{subsec:fdm}--\ref{subsec:spectral} and compared systematically in Table~\ref{tab:classical-pde-methods} provide foundational discretization strategies, but the practical solution of challenging PDEs requires additional computational technologies that enhance performance across multiple base methods. These cross-cutting techniques, such as adaptive mesh refinement, multigrid solvers, and domain decomposition, address computational complexity, enable parallel scalability, and efficiently resolve multiscale features (Section~\ref{subsec:multiscale}). Unlike the method-specific techniques discussed previously, these technologies apply broadly across finite difference, finite element, and finite volume discretizations.

\subsubsection{Adaptive Mesh Refinement}

Adaptive mesh refinement (AMR) dynamically concentrates computational resolution where solution features demand higher accuracy, guided by error estimators derived from local gradients, residual indicators, or feature detection \cite{sarris2006adaptive,bangerth2003adaptive}. The refinement criterion typically takes the form:
\begin{equation}
\label{eq:amr_criterion}
\eta_K > \theta_{\text{refine}} \max_{K'} \eta_{K'} \implies \text{refine element } K,
\end{equation}
where $\eta_K$ represents the local error indicator on element $K$ and $\theta_{\text{refine}} \in (0,1)$ is a user-specified refinement threshold. For FEM, rigorous a posteriori error estimates provide $\eta_K$ with guaranteed bounds on actual errors \cite{verfurth1996review}; for finite difference and finite volume methods, heuristic indicators based on solution gradients or undivided differences typically guide refinement.

Block-structured AMR organizes refinement within rectangular grid patches at different resolution levels, enabling vectorization and cache-efficient computation within patches \cite{berger1989local,dubey2014survey}. Conservative interpolation at refinement boundaries maintains discrete conservation:
\begin{equation}
\label{eq:amr_conservation}
F_{\text{coarse}} = \frac{1}{r^{d-1}} \sum_{\text{fine faces}} F_{\text{fine}},
\end{equation}
where $r$ denotes the refinement ratio and $d$ is spatial dimension. This ensures that coarse-grid fluxes precisely match the sum of fine-grid fluxes at resolution boundaries, preventing spurious sources or sinks. AMR effectively addresses problems involving smooth regions and localized features (boundary layers, shocks, reaction fronts), concentrating degrees of freedom where needed while avoiding the uniform refinement costs.

\subsubsection{Multigrid and Algebraic Multilevel Methods}

Multigrid methods achieve optimal $O(N)$ computational complexity for solving discrete elliptic problems by exploiting complementary error-smoothing properties across spatial scales \cite{brandt1979multigrid,hackbusch1978multi}. Classical iterative methods (Jacobi, Gauss-Seidel) efficiently damp high-frequency errors but slowly reduce low-frequency components. Multigrid addresses this by transferring low-frequency errors to coarser grids where they appear as high-frequency and are efficiently damped.

The standard V-cycle algorithm for solving $\mathbf{A}_h \mathbf{u}_h = \mathbf{f}_h$ proceeds:
\begin{enumerate}
\item \textbf{Pre-smoothing}: Apply $\nu_1$ iterations of smoother $S$: $\mathbf{u}_h^{(1)} = S^{\nu_1}(\mathbf{u}_h^{(0)}, \mathbf{f}_h)$
\item \textbf{Restriction}: Transfer residual to coarse grid: $\mathbf{r}_{2h} = \mathbf{I}_h^{2h}(\mathbf{f}_h - \mathbf{A}_h \mathbf{u}_h^{(1)})$
\item \textbf{Coarse solve}: Solve (or recursively apply multigrid to) $\mathbf{A}_{2h} \mathbf{e}_{2h} = \mathbf{r}_{2h}$
\item \textbf{Prolongation}: Interpolate correction to fine grid: $\mathbf{u}_h^{(2)} = \mathbf{u}_h^{(1)} + \mathbf{I}_{2h}^h \mathbf{e}_{2h}$
\item \textbf{Post-smoothing}: Apply $\nu_2$ smoothing iterations: $\mathbf{u}_h^{(3)} = S^{\nu_2}(\mathbf{u}_h^{(2)}, \mathbf{f}_h)$
\end{enumerate}

Geometric multigrid requires explicit mesh hierarchies and structured grid connectivity. Algebraic multigrid (AMG) generalizes these principles to unstructured problems through strength-of-connection analysis, automatically constructing coarse-level problems and inter-grid transfer operators from fine-level matrix entries alone \cite{stuben2001review}. This enables multigrid efficiency without geometric information or structured grids. AMG achieves scalability as a preconditioner for iterative Krylov methods \cite{yang2002boomeramg,baker2010improving} in large-scale elliptic problems, provided the continuous problem's nullspace (rigid body modes, constant pressure) is correctly specified.

For coupled multiphysics systems (Section~\ref{subsec:multiphysics}), achieving similar scalability necessitates problem-specific strategies including block-structured preconditioners that respect physical field coupling \cite{keyes2013multiphysics,bui2020scalable,both2017robust,zabegaev2026block}. The multilevel framework extends beyond algebraic systems: subspace correction methods \cite{xu1992iterative} generalize these principles to iterative solvers operating on arbitrary functional subspace decompositions, providing unified theory encompassing domain decomposition and block preconditioning.

\subsubsection{Domain Decomposition Methods}

Domain decomposition methods partition computational domains into subdomains, enabling parallel solution on distributed memory architectures \cite{quarteroni1999domain,toselli2004domain}. These methods serve dual purposes: mathematical preconditioning for iterative solvers and practical parallelization strategies for high-performance computing.

Overlapping (Schwarz) methods solve local problems on overlapping subdomains with information exchange through overlapping regions. The additive Schwarz preconditioner combines local solutions:
\begin{equation}
\label{eq:schwarz}
\mathbf{M}_{\text{AS}}^{-1} = \sum_{i=1}^{N_{\text{sub}}} \mathbf{R}_i^T \mathbf{A}_i^{-1} \mathbf{R}_i,
\end{equation}
where $\mathbf{R}_i$ restricts global vectors to subdomain $i$ and $\mathbf{A}_i$ is the local operator. Convergence rates depend on subdomain overlap and on the inclusion of a coarse-grid correction for global information propagation.

Non-overlapping methods employ interface conditions coupling subdomains at boundaries. The Finite Element Tearing and Interconnecting (FETI) method \cite{farhat1991method} enforces continuity across subdomain interfaces via Lagrange multipliers, reducing the global problem to the interface degrees of freedom. Balancing domain decomposition (BDD) achieves optimal convergence rates by carefully handling constraints across subdomains \cite{mandel1993balancing}.

Domain decomposition enables strong scaling to hundreds of thousands of processor cores for production simulations, with sophisticated load balancing and dynamic repartitioning addressing adaptive refinement and multiphysics coupling \cite{boman2013scalable}. The interplay between mathematical convergence properties and parallel computing architecture significantly influences practical solver performance.

\subsubsection{Model Order Reduction and Multiscale Methods within the Subspace Correction Paradigm\label{sec:model-order-reduction}}
Many solver strategies beyond geometric multigrid and domain decomposition fit naturally into the subspace correction framework~\cite{xu1992iterative}. In this view, the trial space $V$ is decomposed into complementary subspaces, and the solution is obtained by applying corrections to these subspaces. Abstractly, one writes
\begin{equation}
V = V_0 \oplus \sum_{i=1}^m V_i, \qquad 
M^{-1} = P_0 A_0^{-1} P_0^{\top} \;+\; \sum_{i=1}^m R_i A_i^{-1} R_i^{\top}. \label{eq:subspace-correction}
\end{equation}
where $V_0$ is a coarse (global) space, $V_i$ are local or complementary spaces, and $P_0$ and $R_i$ are the associated prolongation/restriction operators; $A_0$ and $A_i$ are the operator restrictions. Classical multigrid and domain decomposition select $V_0$, $V_i$ from grid hierarchies or (overlapping/non‑overlapping) subdomains, but the same algebra carries over when subspaces are built spectrally rather than geometrically.

The Reduced Basis method constructs a global low‑dimensional subspace that approximates the solution manifold of a parametrized PDE~\cite{quarteroni2015reduced}. Expensive full‑order “snapshots” are computed offline and compressed (via greedy selection or proper orthogonal decomposition~\cite{benner2017model}) into a basis that delivers rigorous a priori error control for various problem settings; in the subsequent online phase signicantly smaller but dense linear systems need to be solved, enabling rapid queries for control, optimization, UQ, or embedded applications. In subspace‑correction terms, the reduced basis methods realizes a degenerate yet powerful two‑level method whose coarse space already captures the hard global couplings without further local corrections.

Multiscale FEM and FV methods aim at encoding highly heterogeneous or high‑contrast coefficients in local basis functions of a coarse function space~\cite{efendiev2009multiscale,jenny2003multi}. Each basis function solves cell or local spectral problems, so that the coarse space represents fine‑scale physics. This mirrors modern domain decomposition coarse spaces, where local generalized eigenvectors are selected to ensure energy‑stable coarse corrections under contrast~\cite{spillane2014abstract}. In fact, there is a close connection between multiscale discretization methods and domain decomposition solvers~\cite{nordbotten2008relationship}.

\subsection{Meshless and Boundary Element Methods}
\label{subsec:meshless}

Meshless methods represent an alternative discretization paradigm, eliminating traditional mesh connectivity by constructing approximations from scattered node distributions \cite{duarte1995review,li2013meshless}:
\begin{equation}
\label{eq:meshless}
u^h(\mathbf{x}) = \sum_{i=1}^{n(\mathbf{x})} \phi_i(\mathbf{x}) u_i,
\end{equation}
where shape functions $\phi_i(\mathbf{x})$ have local support and $n(\mathbf{x})$ indicates the number of nodes influencing point $\mathbf{x}$. This paradigm offers advantages for problems involving large deformations, crack propagation, and evolving geometries, where classical mesh-based approaches face expensive remeshing.

Radial Basis Function (RBF) methods construct global or local approximations using radially symmetric functions—multiquadric, Gaussian, thin-plate spline—with shape parameters controlling accuracy and matrix conditioning \cite{fornberg2015solving}. Pure RBF methods employ simple distributions of collocation points without a  computational mesh, though dense linear algebra can limit scalability. Local RBF variants with compactly supported basis functions achieve sparse approximations suitable for large-scale problems. Smooth Particle Hydrodynamics (SPH) discretizes PDEs using moving Lagrangian particles carried by material deformation, eliminating convective terms and naturally handling free surfaces in fluid dynamics \cite{lind2020review}.

The Method of Fundamental Solutions employs analytical fundamental solutions as basis functions $\phi_i(\mathbf{x}) = G(\mathbf{x}, \mathbf{x}_i)$, where $G$ is the Green's function, achieving exponential convergence while reducing problem dimensionality to boundaries \cite{alves2009choice}. Boundary Element Methods (BEM) reformulate PDEs as boundary integral equations, reducing volume problems to surface discretizations \cite{katsikadelis2016boundary}. This dimensionality reduction proves particularly effective for exterior domain problems (radiation, scattering) and infinite domain applications. Fast Multipole acceleration \cite{liu2006fast} achieves $O(N)$ complexity through hierarchical approximation of far-field interactions, enabling large-scale boundary element simulations previously intractable due to dense matrix computational costs.

Meshless methods excel at geometric flexibility, naturally handling complex domains, moving boundaries, and topological changes without remeshing. Node insertion or removal for adaptivity becomes straightforward compared to mesh-based approaches, which require connectivity updates. However, practical challenges include computational overhead from shape function evaluation requiring local system solutions, specialized techniques for essential boundary condition enforcement when approximations lack interpolation property, and a lack of systematic guidelines for optimal parameter selection (support sizes, shape parameters). Theoretical frameworks for stability and convergence remain less mature than mesh-based methods, and irregular matrix sparsity patterns complicate efficient parallel implementation.

The methods presented in this section represent decades of mathematical and computational development, each optimized for particular problem characteristics among the six fundamental challenges identified in Section~\ref{sec:challenges}. Table~\ref{tab:classical-pde-methods} synthesizes their computational complexity classes, theoretical guarantees, and distinguishing features, providing a systematic framework for method selection based on problem structure, available computational resources, and accuracy requirements. Section~\ref{sec:evaluationclassical} examines the practical performance of these methods through contemporary benchmarks and large-scale applications.

\section{Capabilities and Limitations of Classical Methods}
\label{sec:evaluationclassical}

Over seven decades of methodological development, a sophisticated ecosystem of classical numerical methods has emerged, as detailed in Section~\ref{sec:classicmethods}. These methods rest on rigorous mathematical foundations. Yet fundamental computational barriers persist despite algorithmic refinement and exponentially growing computing power. This section critically evaluates classical methods against the six challenges identified in Section~\ref{sec:challenges}, examining where decades of refinement have achieved practical solutions, where fundamental mathematical barriers constrain progress, and where enduring strengths establish requirements for any future methodology. Table~\ref{tab:classical-pde-methods} provides quantitative metrics supporting this analysis.

\subsection{Dimensionality: Exponential Scaling as a Fundamental Barrier}
\label{subsec:eval_dimensionality}

The curse of dimensionality (Section~\ref{subsec:dimensionality}) remains the most severe fundamental limitation of classical grid-based methods. For problems discretized with $N$ points per dimension in $d$ spatial dimensions, memory requirements scale as $O(N^d)$ and computational cost as $O(N^{d+\alpha})$ where $\alpha \geq 0$ depends on solver complexity. This exponential scaling in $d$ creates a practical ceiling: three-dimensional problems with $N=1000$ require $10^9$ degrees of freedom and terabytes of memory; extending to $d=10$ dimensions with merely $N=10$ points per dimension demands $10^{10}$ unknowns—already challenging. Moderate resolution with $N=50$ in $d=10$ requires $10^{17}$ unknowns, exceeding any conceivable computational resource.

\begin{figure}[h!]
    \centering
    \includegraphics[width=0.95\textwidth]{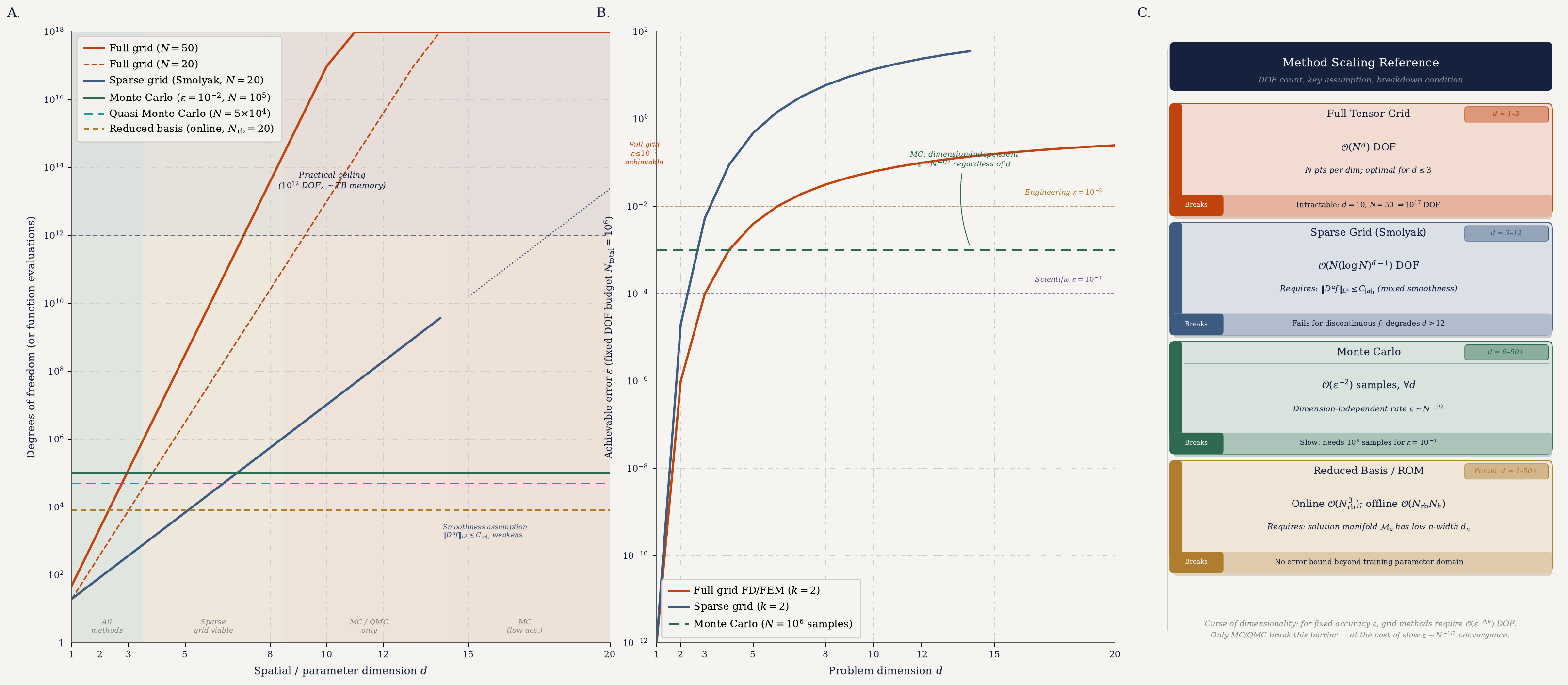}
    \caption{Quantitative illustration of the curse of dimensionality
    for classical grid-based and sampling methods applied to
    $d$-dimensional PDEs and parametric problems.
    \textbf{A.}~Degrees of freedom (or function evaluations) required
    for a fixed resolution $N$ as a function of dimension $d$,
    plotted on a logarithmic scale.
    Full tensor-grid discretisation with $N$ points per dimension
    scales as $N^d$ (crimson curves; $N = 20$ dashed, $N = 50$ solid),
    crossing the practical ceiling of $10^{12}$ degrees of freedom
    (shaded region, corresponding to terabyte-scale memory) at
    $d = 6$ for $N = 50$ and $d = 8$ for $N = 20$.
    The Smolyak sparse grid~\cite{schwab2011sparse} scales as
    $\mathcal{O}(N(\log_2 N)^{d-1})$ (blue), providing
    order-of-magnitude reductions over the full grid but requiring
    bounded mixed derivatives $\|D^\alpha f\|_{L^2} \leq C_{|\alpha|_1}$;
    the dashed continuation past $d = 14$ marks the regime where this
    smoothness condition generically fails.
    Monte Carlo (MC, emerald) and quasi-Monte Carlo (QMC, teal) methods
    are dimension-independent by construction, with sample counts
    determined solely by the target accuracy $\varepsilon$ and independent
    of $d$; the reduced basis method (gold, dashed) achieves low online
    cost $\mathcal{O}(N_{\mathrm{rb}}^3)$ when the solution manifold
    $\mathcal{M}_\mu$ has low Kolmogorov $n$-width, but its offline
    snapshot cost grows with the dimension of the parameter space.
    Background shading identifies four tractability regimes: all methods
    viable ($d \leq 3$); sparse grid viable ($d = 3$--$8$); only MC/QMC
    viable ($d = 8$--$14$); and MC viable at reduced accuracy only
    ($d > 14$).
    \textbf{B.}~Achievable error $\varepsilon$ for a fixed total
    degree-of-freedom budget of $N_{\mathrm{total}} = 10^6$, showing
    how accuracy degrades with dimension for each method.
    Full-grid FD/FEM with $k = 2$ and sparse-grid $k = 2$
   deliver engineering accuracy ($\varepsilon \leq 10^{-2}$)
    only up to $d \approx 4$ and $d \approx 6$, respectively, within
    this budget.  Monte Carlo  achieves a dimension-independent accuracy
    of $\varepsilon \sim N^{-1/2} = 10^{-3}$ regardless of $d$.
    Horizontal reference lines mark engineering ($\varepsilon = 10^{-2}$) and scientific ($\varepsilon = 10^{-4}$) accuracy thresholds.
    \textbf{C.}~The formal DOF scaling expression for the four method families, the key mathematical assumption required for that scaling to hold, and the condition under which the method breaks down.
    The reduced basis method occupies a qualitatively different
    position: its tractability depends not on the spatial dimension
    $d$ but on the intrinsic dimension of the solution manifold
    $\mathcal{M}_\mu$, quantified by the Kolmogorov $n$-width
    $d_n(\mathcal{M}_\mu) \to 0$; when this condition fails
    (e.g.\ for transport-dominated or bifurcating problems), the
    method loses its efficiency advantage.
    Taken together, the three panels establish that breaking the
    curse of dimensionality within the classical paradigm requires
    either strong a priori regularity assumptions (sparse grid) or
    a fundamental change in approximation paradigm (MC/QMC, reduced
    basis), each of which incurs its own structural cost.}
    \label{fig:curse_of_dimensionality}
\end{figure}

The curse of dimensionality is not merely a computational inconvenience but
a provably optimal lower bound for worst-case approximation of general
smooth functions on $d$-dimensional domains: any method that does not
exploit special problem structure requires at least
$\mathcal{O}(\varepsilon^{-d/k})$ degrees of freedom to achieve accuracy
$\varepsilon$ in a Sobolev space $H^k$~\cite{novak1988deterministic}.
For classical grid-based methods, this manifests directly: a problem
discretised with $N$ points per spatial dimension requires $\mathcal{O}(N^d)$
degrees of freedom, so that memory and compute scale exponentially with $d$.
A three-dimensional problem with $N = 1{,}000$ points per dimension requires
$10^9$ degrees of freedom and terabytes of working memory; extending to
$d = 10$ dimensions at the same resolution demands $10^{30}$ degrees of
freedom, exceeding the atom count of the observable universe.
Even modest resolution, $N = 50$ in $d = 10$, yields $10^{17}$ unknowns,
far beyond any conceivable computational resource.
Figure~\ref{fig:curse_of_dimensionality} makes this barrier concrete across
four method families — full tensor grid, sparse grid, Monte Carlo / QMC,
and reduced basis — and exposes both the regime of tractability and the
structural cost of each approach that escapes the exponential trap.

Three features of Fig.~\ref{fig:curse_of_dimensionality} deserve emphasis
before the individual methods are discussed.
First, the practical ceiling of $10^{12}$ degrees of freedom — corresponding
roughly to terabyte-scale memory on current hardware — is crossed by the
$N = 50$ full tensor grid at $d = 6$ and by the $N = 20$ grid at $d = 8$
(panel~A). These are not exotic high-dimensional problems: $d = 6$ arises naturally
in phase-space kinetic equations (3 spatial + 3 velocity dimensions),
multi-asset financial models, and tensor-product parametric PDEs. Second, panel~B reveals a sharper constraint than panel~A: even within a fixed compute budget of $N_{\mathrm{total}} = 10^6$ degrees of freedom,
full-grid FD/FEM can deliver engineering accuracy ($\varepsilon \leq 10^{-2}$)
only up to $d \approx 4$.
Monte Carlo's dimension-independent plateau at $\varepsilon \approx 10^{-3}$
appears as a floor: it is simultaneously the method's greatest strength
(the same accuracy regardless of $d$) and its fundamental limitation (it
cannot be refined below this floor without a quadratic increase in samples).
Third, the reduced basis method (panel~C) escapes the dimensional barrier
through a qualitatively different mechanism: it exploits the fact that for
many parametric PDE families the solution manifold $\mathcal{M}_\mu$
has low Kolmogorov $n$-width, meaning that solutions lie near a
low-dimensional affine subspace even when the parameter domain is
high-dimensional.
This makes it the method with the highest potential payoff when the
assumption holds — and the method with the most catastrophic failure
mode when it does not, since the offline snapshot cost provides no
error bound beyond the training parameter domain.

Recent advances in sparse grid methods \cite{schwab2011sparse} mitigate this for problems with sufficient smoothness and low-dimensional structure, reducing effective dimensionality from $d$ to approximately $\log^k d$ for appropriate problem classes. However, these methods require strong regularity assumptions that fail for the discontinuous solutions and sharp gradients characteristic of many applications (Section~\ref{subsec:discontinuities}). 

Monte Carlo methods circumvent grid-based discretization through random sampling, maintaining $O(N_{\text{samples}}^{-1/2})$ convergence rate independent of dimension~\cite{kuo2005lifting,weinan2021algorithms}. While dimension-independent, the slow convergence requires large number of samples for modest accuracy, which is practical for linear problems with simple geometries but prohibitively expensive for nonlinear PDEs requiring expensive PDE solves per sample. For high-dimensional problems that couple significant nonlinearity with geometric complexity (e.g., financial risk models with $d=50$ assets and complex payoff structures, quantum many-body systems), no existing classical method provides computationally tractable solutions beyond proof-of-concept demonstrations.

For high-dimensional parameter spaces, e.g., in optimization and uncertainty quantification, the reduced basis method (Section~\ref{sec:model-order-reduction}) is able to significantly reduce complexity for each PDE solve. When based on $N_{\text{rb}}\ll N$ snapshots, online efficiency reduces to $O(N_{\text{rb}}^3)$, but the generation of snapshots during the offline training phase still requires expensive computation of full-order solutions at carefully selected parameter values. More critically, extrapolation beyond the training parameter regime lacks rigorous error bounds, limiting applicability when parameter spaces are incompletely explored.

This fundamental barrier, proven optimal for worst-case complexity in many settings, suggests that breaking the curse of dimensionality may require fundamentally different approximation paradigms exploiting problem-specific structure rather than uniform discretization. The scaling landscape of Fig.~\ref{fig:curse_of_dimensionality} yields a formulation of what any future methodology must achieve: either exploit provable low-dimensional structure in the solution manifold — as the reduced basis method does, conditionally — or provide dimension-independent accuracy — as Monte Carlo does, slowly — and ideally both simultaneously, with certified error control that neither currently provides beyond its respective training or sampling regime.

\subsection{Nonlinearity: High Fidelity at the Cost of Robustness}
\label{subsec:eval_nonlinearity}

Classical methods, often based on principles like the Galerkin method, provide high-fidelity numerical models which is especially highlighted for nonlinear PDEs and problems built on intrinsic first principles. The explicit control over the design has enabled the development of discretization methods that honor conservation principles, preserve invariants, and encode governing symmetries by construction.

Yet Nonlinear PDEs (Section~\ref{subsec:nonlinearity}) expose critical differences between theoretical convergence guarantees, practical solvability and solver robustness. Newton-Raphson methods employed throughout finite element and spectral approaches (Table~\ref{tab:classical-pde-methods}) offer quadratic convergence near solutions but fail with poor initialization, requiring sophisticated globalization strategies, such as line search, trust-region, continuation methods~\cite{deuflhard2011newton}, that substantially increase implementation complexity. Robust solve performance for complex problems typically requires model‑aware solver design together with problem‑specific stability and convergence analysis.

For coupled multiphysics systems (Section~\ref{subsec:multiphysics}), nonlinearity compounds across fields. Interface-coupled fluid-structure interaction couples nonlinear Navier-Stokes equations with geometrically nonlinear elasticity, requiring simultaneous solution of equations \eqref{eq:navier_stokes_momentum}--\eqref{eq:navier_stokes_continuity} and \eqref{eq:elastodynamics} subject to interface conditions \eqref{eq:fsi_kinematic}--\eqref{eq:fsi_dynamic}. A similar situation arises in domain‑coupled settings such as poroelasticity. Monolithic solution preserves strong coupling but yields large nonlinear systems with saddle-point structure requiring specialized preconditioners respecting field coupling \cite{mardal2011preconditioning,bui2020scalable,both2017robust}. Partitioned approaches decouple fields but introduce coupling iterations whose convergence depends sensitively on physical parameter ratios and time-step sizes, with well-documented problem-specific stability challenges, e.g., for lightweight structures in dense fluids \cite{forster2007artificial} and poroelasticity~\cite{storvik2019optimization}.

Robust nonlinear solvers for coupled systems remain an active research frontier, where the theoretical foundations of classical methods prove essential. The requirement for provable convergence under physically reasonable assumptions, systematic globalization strategies, and adaptive coupling algorithms represents an enduring strength that any alternative methodology must match. Recent advances in physics-compatible preconditioners \cite{zabegaev2026block,lohmann2019physics} demonstrate continued progress, yet many industrially relevant configurations—high Reynolds number flows coupled with thin flexible structures, multiphase reactive flows with sharp interfaces remain at or beyond the boundary of tractable solution.

The practical performance of nonlinear solvers, in terms of robustness rather than nominal discretization accuracy, often determines whether realistic coupled problems can actually be solved, still posing one of the apparent  research questions for nonlinear PDEs.While high‑fidelity, first‑principles nonlinear models impose stringent robustness requirements, the availability of systematic nonlinear solution strategies together with supporting analysis and guarantees is a core strength of classical methods that future hybrid approaches should preserve and build upon.

\subsection{Geometric Complexity: Meshing as a Persistent Bottleneck}
\label{subsec:eval_geometry}

Classical mesh-based methods' treatment of geometric complexity (Section~\ref{subsec:geometry}) reveals a persistent dichotomy between theoretical capability and practical workflow. Finite element methods (Section~\ref{subsec:fem}) provide rigorous theoretical treatment of complex domains through unstructured mesh capability, achieving optimal approximation properties even for curved boundaries when appropriate element mappings are employed \cite{bernardi1989optimal}. Polyhedral discontinuous Galerkin, virtual element, and isogeometric methods (see Table~\ref{tab:classical-pde-methods}) extend this flexibility further, with PolyDG and VEM handling general polygonal/polyhedral elements and IGA eliminating geometric approximation entirely through direct CAD representation.

Yet despite its theoretical appeal, mesh generation for complex 3D geometries still dominates, often by far, the overall analysis time in real-world engineering workflows. While progress is still made throughout multi-decade research~\cite{thompson1998handbook}, open problems still remain~\cite{pietroni2022hex}. 
The need for the human-in-the-loop slows down workflows. Patient-specific biomedical geometries require extensive manual intervention to repair imaging artifacts, remove small features, and ensure mesh quality \cite{wittek2016finite}. Industrial CAD models contain geometric details (bolt holes, fillets, chamfers) spanning multiple length scales that must be simplified to enable tractable mesh generation and solution.

Automated mesh generation with rigorous quality guarantees remains an open challenge for general three-dimensional domains~\cite{pietroni2022hex,elshakhs2024comprehensive}. Delaunay refinement algorithms offer provable quality guarantees in two dimensions, yet their extension to three dimensions is possible only under significant restrictions. Concretely, while two-dimensional Delaunay refinement guarantees bounds on both the minimum angle and the radius-edge ratio of every output triangle~\cite{shewchuk2002delaunay}, the analogous three-dimensional algorithms guarantee only a bound
on the radius-edge ratio of tetrahedra. This bound does not preclude the formation of slivers, nearly degenerate tetrahedra with an arbitrarily small minimum dihedral angle that satisfy the
radius-edge criterion yet produce severely ill-conditioned element
stiffness matrices. Post-processing techniques such as sliver exudation can remove slivers in practice~\cite{cheng2000sliver}, but they require the input domain boundary to be free of small angles and provide no worst-case guarantee on the minimum dihedral angle of the final mesh. Termination of the refinement loop is likewise guaranteed only when the input piecewise-linear complex contains no sharp features, i.e., no dihedral angles below a
problem-dependent threshold~\cite{shewchuk1998tetrahedral}.

Advancing-front and octree-based methods have achieved considerable practical success~\cite{schoberl1997netgen,burstedde2011p4est}, but they typically require careful, problem-specific parameter tuning and provide no rigorous a priori bounds on mesh quality. Poor mesh quality due to, e.g., elements with high aspect ratios, small dihedral angles, or severe distortion, directly impairs the accuracy of finite element approximations~\cite{ern2021finite}, increasing the condition number of resulting stiffness matrices directly impeding the efficient numerical solution through iterative solvers, irrespective of the sophistication of the underlying discretization or solution method.

This geometric preprocessing bottleneck represents a fundamental workflow inefficiency where human expertise remains essential despite algorithmic and theoretical advances. The contrast between finite elements' geometric flexibility and the labor-intensive mesh generation required to exploit it motivates continued investigation of mesh-free alternatives and automatic geometry-to-solution workflows, particularly in applications in need for non-fixed geometries such as patient-specific biomedical applications, shape optimization and rapid design iteration in engineering. In many workflows, reframing the problem earlier in the pipeline, for instance, adopting mixed‑dimensional or embedded‑interface formulations in the presence of thin structures~\cite{berre2019flow}, can relax meshing constraints, even if it relocates complexity to model selection, interface conditions, and structure‑preserving discretizations.

\subsection{Discontinuities: Accuracy--Robustness Trade-offs}
\label{subsec:eval_discontinuities}

Discontinuities and sharp fronts fundamentally challenge high-order reconstructions and global approximations (Section~\ref{subsec:discontinuities}). Carefully designed designed nonlinear limiters and WENO schemes (Section~\ref{sec:traditional-methods:advanced-schemes}) enable reliable shock capturing and high-accuracy treatment of discontinuous solutions and sharp gradients resulting in their widespread use in convection-dominated flows~\cite{shu2020essentially}. 

Despite these advances, simultaneously achieving sharp shock capturing in discontinuous regions and high-order accuracy in smooth regions poses challenges. High-order schemes require wide stencils, which complicates parallel decomposition and boundary treatment. Moreover, achieving genuinely multidimensional reconstruction, essential for maintaining accuracy on unstructured meshes and for problems with oblique discontinuities, substantially increases complexity compared to dimension-by-dimension approaches \cite{titarev2004finite}. To conclude, limiter design involves a delicate balance: excessive limiting degrades accuracy in smooth regions, whereas insufficient limiting produces spurious oscillations. Optimal limiting strategies remain problem-dependent and require expertise for robust application.

For multiphysics problems that couple smooth and discontinuous fields, such as combustion with turbulent mixing and chemical reactions, or multiphase flows with complex interface topologies, the challenge intensifies. Different fields may require incompatible discretization strategies: spectral methods for smooth velocity fields versus shock-capturing schemes for density discontinuities finite volume/finite difference schemes for densities. Following the discussion of coupled problems (Section~\ref{subsec:eval_nonlinearity}), such combined approaches may introduce interface matching complexity and potential conservation violations if not designed carefully.

Spectral methods are particularly vulnerable in the presence of discontinuities (Section~\ref{subsec:spectral}). 
The Gibbs phenomenon~\cite{gottlieb1997gibbs} generates persistent global oscillations near discontinuities that pollute the entire solution, even in regions of local smoothness. While filtering and artificial diffusion may mitigate oscillations, these remedies erodes the exponential accuracy that constitutes the attraction of spectral discertization methods. This sharp contrast of exponential convergence for smooth problems versus outright failure in the presence of discontinuities underscores how strongly the regularity of the underlying problem governs method suitability across classical discretization strategies.

\subsection{Multiscale Phenomena: The Scale Separation Requirement}
\label{subsec:eval_multiscale}

Multiscale PDEs (Section~\ref{subsec:multiscale}) can be solved in two principled approaches using classical methods. Either any spatial multiscale nature is resolved by the mesh, or it is directly incorporated in the design of the method. The former approach poses a clear  computational challenge. For heterogeneous media with rapidly oscillating coefficients as described by equation~\eqref{eq:multiscale_elliptic}, direct discretization requires resolving the finest scale $\epsilon$, yielding $O(\epsilon^{-d})$ unknowns. Consequently, contemporary materials with microscale features ($\epsilon \sim 10^{-6}$) in three dimensions produce intractable problem sizes.

Multiscale finite element and multiscale finite volume methods (Section~\ref{sec:model-order-reduction}) address this through coarse-scale equations with microscopically informed basis functions, reducing complexity to $O(N_c \cdot N_f)$ where $N_c \ll \epsilon^{-d}$ represents coarse degrees of freedom and $N_f$ characterizes local fine-scale computation, required for constructing problem- and material-specific basis functions. This achieves order-of-magnitude efficiency improvements for problems with periodic or statistically homogeneous microstructure. Localized orthogonal decomposition methods exploit the exponential decay of corrector functions to achieve quasi-local approximation with optimal error estimates \cite{maalqvist2014localization}.

However, many practical problems exhibit multiple interacting scales without clear separation making multiscale methods non-applicable, requiring approaches like numerical homogenization~\cite{Altmann_Henning_Peterseim_2021},. When microstructure geometry varies spatially or evolves temporally, for instance, crack propagation in heterogeneous materials or microstructure coarsening in phase-field models, offline basis precomputation becomes invalid, forcing expensive online adaptation. The heterogeneous multiscale method's $O(N_M \cdot N_m^d)$ complexity reveals explicit curse-of-dimensionality dependence on microscale spatial dimension $d$, with costs exploding for three-dimensional microscale models.

Temporal multiscale stiffness in reaction-diffusion systems couples fast chemical reactions with characteristic time scales $\sim \epsilon$ and slow diffusion with characteristic time scales $\sim 1$ requiring time steps $\Delta t \lesssim \epsilon$ for explicit integration~\cite{wanner1996solving}. Implicit methods relax stability restrictions but require solving large nonlinear systems at each step, often with conditioning that degrades as stiffness increases. Implicit-explicit schemes partially address this but introduce splitting errors and require careful stability analysis for nonlinear coupling~\cite{ascher1997implicit}.

Bridging arbitrary scale gaps without assuming scale separation or statistical homogeneity remains a fundamental open challenge. Classical methods offer valuable conceptual frameworks, notably numerical homogenization theory and the derivation of effective equations, yet computationally tractable implementations for truly general problems continue to prove elusive.

\subsection{Multiphysics Coupling: Compatibility and Structure Preservation}
\label{subsec:eval_multiphysics}

Coupled multiphysics systems (Section~\ref{subsec:multiphysics}) lie at the frontier where the rigor of classical methods is both essential and constraining. While naïve compositions of methods for the constituent single-physics subproblems often fall short, progress has been driven by careful analysis and principled algorithmic design~\cite{keyes2013multiphysics}. A precise understanding of the coupling structure across problem classes has proved crucial for developing stable discretizations and scalable solvers. 

A central difficulty is the need to respect compatibility among function spaces as dictated by the governing PDEs. The inf–sup stability condition~\eqref{eq:infsup} for mixed formulations exemplifies fundamental mathematical constraints \cite{bathe2001inf,brezzi2008mixed}: discretization spaces for coupled fields cannot be made independently without risking spurious modes, loss of convergence, or nonuniqueness of solutions~\cite{elman2014finite}. Preserving the essential mathematical structure is a central strength of classical methods, and this perspective has led to tailored inf–sup–stable schemes across applications~\cite{boffi2013mixed}. A key milestone in modern applied mathematics has been the development of finite element exterior calculus~\cite{arnold2006finite}, understanding that discrete stability follows from continuous stability when finite‑element spaces form a commuting, exact sequence. Together, these developments have delivered, among others, stable, spurious‑mode–free velocity–pressure coupling for Stokes and Navier–Stokes flows, locking‑free stress–displacement approximations in elasticity including nearly incompressible materials, and compatible electromagnetic field representations; typically with implications on other multiphysics systems like magnetohydrodnamics or poroelasticity \cite{sani1981causeA,sani1981causeB,oyarzua2016locking,krysl2008locking,zhu2006multigrid,jiao2002general}.

The challenges of combining discretization schemes for coupled problems with saddle-point structure propagate to the solver level: the resulting algebraic systems are indefinite and strongly coupled, making robust preconditioning nontrivial~\cite{elman2014finite,mardal2011preconditioning}. Scalable large scale monolithic solution of coupled systems requires often block-structured solvers (Section~\ref{subsec:crosscutting}). Physics-based block preconditioners leverage approximate Schur complements and problem‑aligned block factorizations to obtain mesh‑ and parameter‑robust convergence where purely algebraic techniques may degrade~\cite{bui2020scalable,zabegaev2026block}. Gradient flow structures in PDEs also inform solver design through natural metric scalings and symmetric operator blocks, resulting in correct energy-dissipation behavior~\cite{both2019gradient}.

Yet scalability challenges persist. Strong coupling between disparate temporal scales—fast structural vibrations coupled with slow thermal diffusion—requires either implicit solution of the entire coupled system (expensive, complex preconditioning) or partitioned approaches with coupling iterations (convergence challenges, accuracy-stability trade-offs). Optimal coupling strategies remain problem-specific and require expertise and careful analysis. The requirement for structure-preserving discretizations extends beyond inf-sup stability to conservation laws, energy dissipation, and entropy stability. Structure-preserving methods (Table~\ref{tab:classical-pde-methods}), energy quadratization, symplectic integrators, and summation-by-parts schemes achieve provable conservation properties that are essential for long-time integration. These represent critical capabilities, and any alternative methodology must either preserve them or provide equivalent guarantees for physical consistency.

\subsection{Overall Assessment: Strengths, Guarantees, and Persistent Barriers}
\label{subsec:eval_final}

Taken together, the preceding analysis portrays classical methods as both indispensable and bounded. Their strength lies in a structure‑first paradigm: discretizations and solvers are built to respect the governing PDEs' intrinsic properties including compatibility (e.g., inf–sup), geometric structure (de~Rham complexes), and variational/thermodynamic principles (energy–dissipation).  When that structure is preserved end-to-end, from function-space design through preconditioning and time integration, one obtains robust, parameter-aware algorithms that scale to realistic multiphysics settings (Sections~\ref{subsec:eval_nonlinearity} and~\ref{subsec:eval_multiphysics}), supported by rigorous analysis that enables high-fidelity designs and verification of implementations. Scalable solvers tailored to the problems' structure enable efficient practical implementation.

At the same time, persistent barriers remain structural rather than merely implementational: exponential growth in degrees of freedom with dimension (Section~\ref{subsec:eval_dimensionality}); multiscale phenomena without clear scale separation (Section~\ref{subsec:eval_multiscale}); sharp fronts and discontinuities that force accuracy–robustness trade-offs (Section~\ref{subsec:eval_discontinuities}); and workflow friction from geometry handling (Section~\ref{subsec:eval_geometry}). For coupled and nonlinear systems (Sections~\ref{subsec:eval_nonlinearity} and~\ref{subsec:eval_multiphysics}), solver performance hinges on problem-aware analysis (e.g., parameter-robust preconditioning and coupling-stability criteria), underscoring that practicality is often only achieved when problem-specific theory, discretization choices, and linear-algebra design are developed simultaneously. Yet for many contemporary regimes, the supporting theory remains incomplete.

Figure~\ref{fig:method_challenge_map} synthesizes and summarizes the capability assessment covered above into a unified method--challenge matrix. The emerging pattern is clear: geometric complexity (column~III) and nonlinearity (column~II) are generally tractable with unstructured meshes and Newton-type solvers, respectively; in contrast, high dimensionality (column~I) and multiscale phenomena (column~V) remain persistent, fundamental barriers across all classical methods considered. Discontinuities (column~IV) sharply separate conservation-preserving schemes (FVM, DG-FEM), which handle them naturally, from spectral methods, which suffer catastrophic Gibbs oscillations. Multiphysics coupling (column~VI) is feasible but imposes stringent compatibility requirements, notably inf-sup stability in mixed problems, that severely limit admissible discretization pairings. These enduring limitations, especially in high dimensions and multiscale resolution, provide the core computational rationale for the machine learning techniques explored in Section~\ref{sec:mlmethods}.

\begin{figure}[h]
    \centering
    \includegraphics[width=0.95\textwidth]{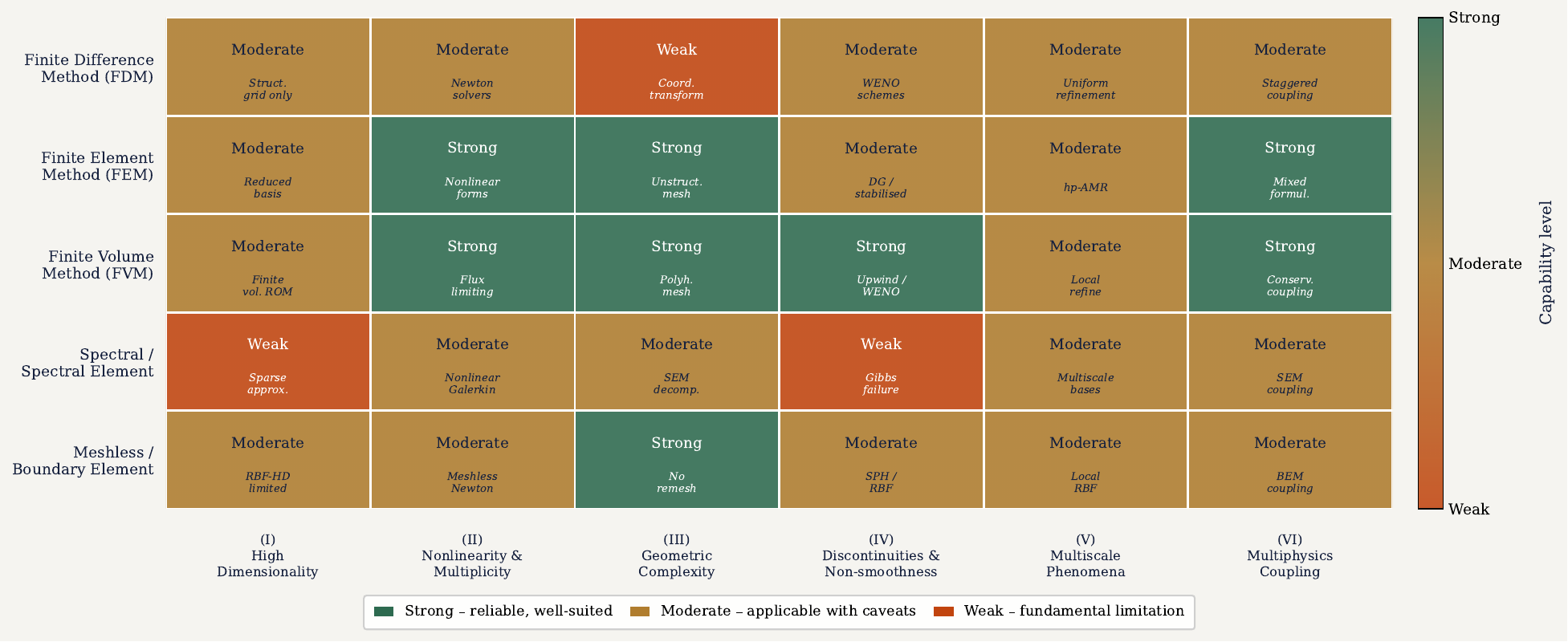}
    \caption{Capability assessment matrix for five classical discretization families across the six core computational challenges outlined in Section~\ref{sec:challenges}. Each cell is color-coded based on the method's inherent mathematical suitability, accompanied by a concise annotation of the key enabling mechanism or limiting factor. Rows: finite difference methods (FDM), finite element methods (FEM), finite volume methods (FVM), spectral/spectral element methods, and meshless/boundary element methods. Columns: (I)~high dimensionality, (II)~nonlinearity and solution multiplicity, (III)~geometric complexity, (IV)~discontinuities and non-smoothness, (V)~multiscale phenomena, (VI)~multiphysics coupling. No classical method excels uniformly across all challenges. The most intractable weaknesses---high dimensionality (I) and multiscale phenomena (V)---persist across the board and provide the primary impetus for exploring machine learning approaches in Section~\ref{sec:mlmethods}.}
    \label{fig:method_challenge_map}
    \label{fig:method_challenge_map}
\end{figure}

Between theory and practice, a human and the computing architecture remain in the loop and must be accounted for. Even in tightly specified community benchmarks that resemble real‑world complexity~\cite{spe11}, meaningful discrepancies persist across groups due to human choices in methods and implementations. Often, “the method’’ also cannot be separated from its implementation; practical performance depends on solver architecture and efficient implementation with minimal memory footprint and scalable use of modern HPC (e.g., domain decomposition).

These observations yield requirements for any alternative or hybrid methodology that aims to complement classical methods: preserve and extend the core classical strengths (compatibility, conservation, energy/entropy stability, verifiability) rather than neglecting or re-inventing them. Most importantly, the well‑posedness and stability are inherited rather than tuned. 

Concretely, alternative or hybrid approaches must preserve structure end-to-end (compatibility, conservation, energy/entropy inequalities) so that well-posedness is inherited; provide verifiability via a priori/a posteriori error control and certified quantities of interest; and deliver problem-aware robustness for nonlinear and coupled problem. Ideally, they should reduce the human-in-the-loop through assisted method selection or streamlined problem-to-solution workflows, and scale by design via multilevel, matrix-free, and communication-aware implementations.

\section{Machine Learning Methods for PDE Solution}
\label{sec:mlmethods}

ML approaches for modeling and solving PDEs in computational science problems represent a
fundamental paradigm shift from classical numerical methods discussed in
Section~\ref{sec:classicmethods}.
Rather than approximating differential operators on domain grids (like in FDM, FEM, and FVM),
ML models learn PDE solutions in high-dimensional spaces by optimizing network weights to predict
solution fields over space and time.
Through this input-to-output mapping, governing equations as well as boundary conditions along
with observational data (either from simulation or experiments) are usually enforced explicitly
or implicitly, depending on the ML modeling framework.
The broader trajectory of ML-for-PDE research has been shaped by, and builds directly upon, classical reduced-order modeling (ROM) techniques such as Proper Orthogonal Decomposition (POD), Dynamic Mode Decomposition (DMD), and Reduced Basis (RB) methods~\cite{quarteroni2015reduced,hesthaven2016certified}.
These classical surrogates established the foundational principle of learning low-dimensional representations of high-dimensional PDE solution manifolds, as many modern neural architectures can be understood as nonlinear extensions of this paradigm. Specifically, the subspace correction framework of Section~\ref{sec:model-order-reduction}, which unifies multigrid, domain decomposition, and reduced basis methods under a single algebraic identity, provides the most natural
mathematical lens through which to understand neural operators: FNO's spectral layers realize a learned global coarse space $V_0$, while DeepONet's trunk network constructs data-adaptive basis functions that generalize the POD modes of classical reduced basis methods to nonlinear, input-dependent settings.

The ML-based approaches specified here can address specific computational barriers identified in
Section~\ref{sec:evaluationclassical}.
In particular, the curse of dimensionality
(Section~\ref{subsec:dimensionality}) motivates the use of ML-based surrogate models for
forward prediction instead of directly solving the governing equations; geometric complexity
(Section~\ref{subsec:geometry}) motivates mesh-free formulations; and multi-query scenarios
(multi-physics or dimensions) justify computationally expensive ML training that can be used
for rapid prediction during solving inverse problems, parametric sensitivity analyses, and
uncertainty quantification.

Section~\ref{subsec:eval_final} articulated five requirements for any alternative or hybrid methodology: \emph{preserve
physical structure end-to-end} (compatibility, conservation, energy/entropy stability); \emph{provide verifiability} through a priori or a posteriori error control; \emph{deliver
problem-aware robustness} for nonlinear and coupled systems; \emph{reduce the human-in-the-loop burden} through streamlined problem-to-solution workflows; and \emph{scale
by design} via multilevel, matrix-free, or communication-aware implementations.
These five requirements serve as an evaluative backbone throughout this section: each ML
method family is introduced in Sections~\ref{sec:surrogate_models}--\ref{sec:hybrid_ml} and
assessed collectively against them in Section~\ref{subsec:ml_challenges}.
The concurrent challenge scenarios of Section~\ref{subsec:concurrent_challenges} exemplify the regimes that most motivate ML-PDE research, precisely because they combine the challenges that most severely strain the classical methods
of Section~\ref{sec:classicmethods}.

\subsection{A Taxonomy of ML Approaches}
\label{subsec:ml_family_overview}

\begin{figure}[t]
    \centering
    \includegraphics[width=1\linewidth]{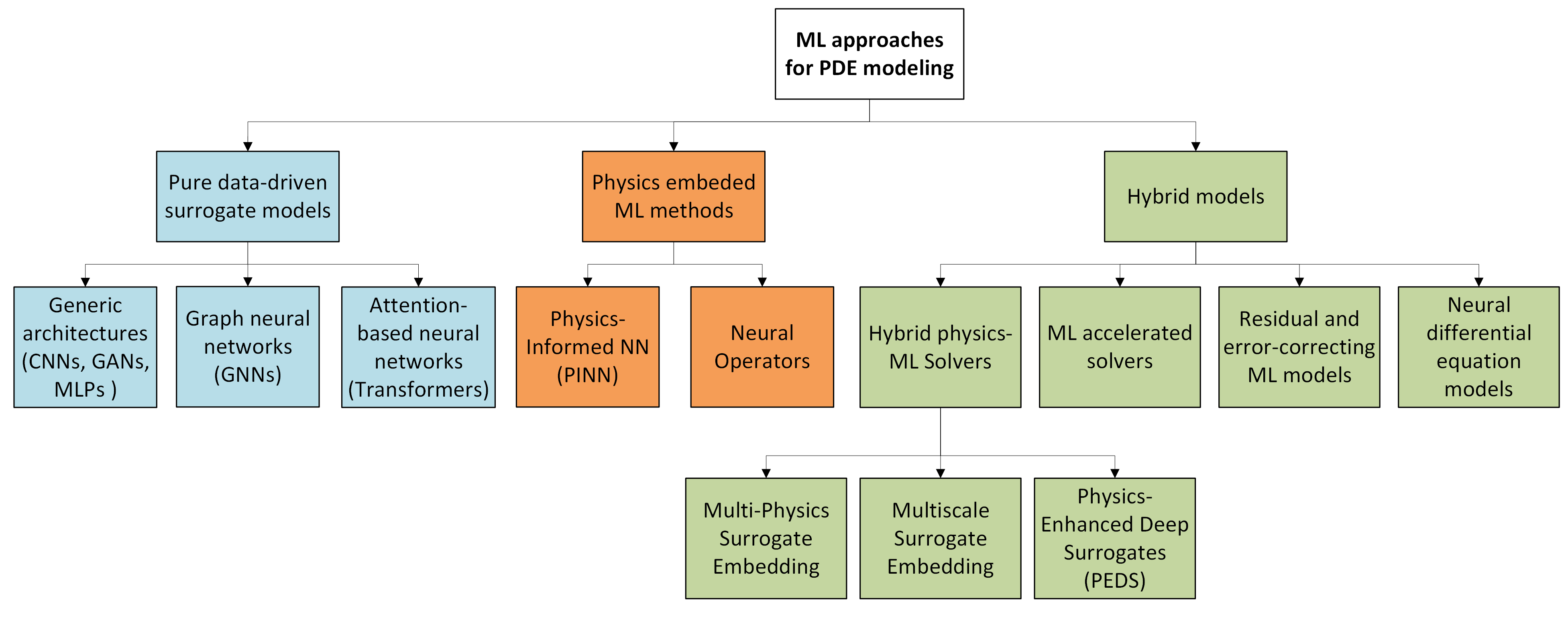}
    \caption{Classification of ML models for PDE modeling. Methods are organized by the degree to which physical knowledge is embedded into the learning process, ranging from purely data-driven surrogates to hybrid approaches that tightly couple learned components with classical solvers.}
    \label{fig:ML_classification}
\end{figure}

ML-based approaches to model and solve PDEs can be classified from different perspectives.
Here, we classify them according to how physical knowledge of the problem is incorporated into
the learning process. Figure~\ref{fig:ML_classification} summarizes the main ML approaches for PDE modeling.
The traditional ML approach implements \textit{pure data-driven surrogate models}
(Section~\ref{surrogate_models}), in which problem inputs (i.e., geometry, initial and
boundary conditions,  physical parameters, and source functions) are mapped directly to outputs (i.e., solution fields over space and/or time) in a purely black-box manner, without explicitly enforcing governing
equations. This class includes a wide variety of deep ML models like generic neural architectures
(e.g., convolutional neural networks (CNNs), generative models, and multi-layer perceptrons (MLPs)), graph neural networks (GNNs), and attention-based transformer models. In the generic category (Section \ref{subsubsec:generic_deep}), any architecture of deep neural networks, such as CNNs, generative models, or deep MLPs, can be implemented. The main drawback is that all deep learning models require huge datasets for training, i.e., they are data-hungry. GNNs (Section \ref{subsubsec:gnns}) and transformers (Section \ref{subsubsec:transformers}) provide architectural flexibility for irregular domains and long-range interactions, while generative models enable probabilistic solution characterization, which is essential when uncertainty arises from sparse observations or stochastic forcing.
The second class of ML models corresponds to \textit{physics-embedded ML methods} (Section~\ref{sec:physics-embedded_ml}). Governing physical laws derived directly from PDEs are explicitly integrated into the learning process. In physics-informed neural networks (PINNs), physics-derived constraints are directly incorporated into the loss function of the ML model, enabling solution approximation with minimal classical solver calls as input data (Section \ref{sec:pinns}).
Neural operators (Section \ref{sec:neural_operators}) are another subcategory of physics-embedded ML methods. They approximate solution operators between input-output function spaces. In other words, instead of learning grid-based data mappings, they learn mappings from input functions to output functions. Neural operators can be considered as nonlinear generalizations of classical ROM techniques: where POD-Galerkin methods learn a linear subspace of the solution manifold. Neural operators such as Fourier neural operator (FNO) and deep operator network (DeepONet) learn nonlinear mappings between infinite-dimensional function spaces, substantially extending expressiveness at the cost of interpretability and training data requirements.
Finally, \textit{hybrid models} (Section~\ref{sec:hybrid_ml}) couple ML models with classical numerical PDE solvers, preserving the theoretical guarantees identified as critical strengths in Section~\ref{sec:evaluationclassical}, while accelerating computational bottlenecks. In other words, this hybrid approach integrates ML models into the PDE solving process and does not completely replace the classical PDE-solving approaches. Within the hybrid category, ML models can be used in several ways. In hybrid physics-ML solvers (Section \ref{subsubsec:Hybrid_physics_ML_Solvers}), ML models replace selected parts of the physics or specific scales of the problem, such as constitutive relations, parametric terms, or other components within the classical numerical formulation. This class includes strategies such as multi-physics surrogate embedding, multiscale surrogate embedding, and physics-enhanced deep surrogates (PEDS). In ML-accelerated solvers (Section \ref{subsubsec:ML_accelerated_solvers}), learned models are introduced to speed up expensive solvers. Residual and error-correcting ML models (Section \ref{eq:residual_correction}) learn correction terms that compensate for modeling deficiencies or unresolved physics. Finally, neural differential equation models (Section \ref{subsubsec:Neural_Differential_Models}) parameterize parts of the system dynamics in a learnable, differentiable form, allowing evolution equations to be augmented.
The classification in Figure\ref{ML_classification}  is based on the functional role of the learned module  relative to the governing PDE. Methods are distinguished according to (i) whether the classical PDE solver is  retained during modeling, (ii) whether physical laws are enforced as constraints in ML models, and (iii) whether ML solver modifies the operator structure. This role-based taxonomy avoids ambiguities arising from architectural similarities across categories. It's worth noting that this classification is role-based rather than architecture-based. For example, ML architectures such as CNNs, GNNs, or Transformers may appear in multiple categories depending on how they are integrated into the modeling pipeline. This perspective avoids ambiguities arising from architectural similarities across categories. 
Table~\ref{tab:ml_method_families} categorizes the main families of ML methods for PDE modeling, highlighting computational complexity and data requirements. No single approach dominates universally and method selection requires matching problem characteristics (e.g., PDE type, geometry, data availability, accuracy requirements, etc.) to methodological strengths.
The subsequent subsections detail the specific architectures within each family, examining mathematical formulations, variants and recent advances, capabilities, and fundamental limitations that constrain applicability.

\begin{table}[t]
\centering
\caption{Machine learning method families for PDE solving: core principles, computational
complexity, and data scaling. Complexity reflects the typical computational cost per forward evaluation; actual performance depends on problem-specific factors including resolution, dimensionality, and architecture choices. }
\label{tab:ml_method_families}
\begin{tabular}{p{4.8cm}p{4.2cm}p{3.0cm}p{2.5cm}}
\toprule
\textbf{Method Family} & \textbf{Core Principle} & \textbf{Complexity} &
\textbf{Data Needs} \\
\midrule
\multicolumn{4}{l}{\textit{Pure Data-Driven Surrogates}} \\[2pt]
Generic Deep Surrogates
(CNNs, MLPs, U-Nets, RNNs)
  & Black-box regression from inputs to solution fields
  & $O(N)$--$O(N^2)$
  & High \\[4pt]
Graph Neural Networks
  & Message passing on meshes and point clouds
  & $O(|V| + |E|)$
  & Moderate \\[4pt]
Transformers
  & Global attention over spatial tokens
  & $O(N^2)$; $O(N \log N)$ with efficient attention
  & High \\[4pt]
Generative Models
  & Probabilistic distribution over solutions
  & Problem-dependent
  & Moderate \\
\midrule
\multicolumn{4}{l}{\textit{Physics-Embedded Methods}} \\[2pt]
Physics-Informed NNs
  & PDE residual enforced in loss function
  & $O(N_{\text{pts}} \cdot P)$
  & Minimal \\[4pt]
Neural Operators
  & Function-to-function operator maps
  & $O(N \log N)$--$O(N^2)$
  & Moderate--High \\
\midrule
\multicolumn{4}{l}{\textit{Hybrid Methods}} \\[2pt]
ML--Solver Coupling
  & Learned components embedded in classical solvers
  & Varies
  & Moderate \\[4pt]
Neural Differential Equations
  & Continuous-time learned dynamics via neural ODE
  & $O(T \cdot P)$ (adjoint)
  & Low--Moderate \\[4pt]
Koopman / EDMD
  & Nonlinear dynamics lifted to linear latent space
  & $O(m^3)$ (online)
  & Moderate \\
\bottomrule
\end{tabular}
\vspace{0.5em}

\footnotesize{\textit{Note:} $P$~denotes network parameters; $N$~is grid or point resolution;
$|V|$,~$|E|$ are graph vertices and edges; $T$~is the number of integration steps for neural ODE solvers; $m$~is the Koopman embedding dimension.
For classical ROM baselines (POD, DMD, Reduced Basis) and their relationship to neural
operators, see Section~\ref{subsec:rom_comparison}.}
\end{table}

\subsection{Pure Data-Driven Surrogate Models}
\label{surrogate_models}
\subsubsection{Generic Deep Surrogate Architectures}
\label{subsubsec:generic_deep}

All deep neural architectures, as the most traditional class of ML-based approaches for PDE
modeling, treat numerical methods for PDE solution as a pure function-approximation problem.
Typically trained in a supervised manner, the input and solution output of these models are usually represented as image datasets. They learn a non-linear mapping from problem inputs (e.g., initial and boundary conditions, geometry, physical parameters, and source terms) to solution fields over space and/or time. Moreover, they do not explicitly consider conservative governing equations and operate as pure black-box surrogate models. 
These network architectures have been used effectively for a variety of elliptic and parabolic PDE problems. Early studies in this area relied on multilayer perceptrons (MLPs) to construct approximations of PDE solutions, especially for low-dimensional parameter spaces~\cite{SHIRVANY200920,gao2024energy,lagaris1998artificial}. However, they have minimal application for grid-based datasets, in which CNNs such as U-Nets gained popularity~\cite{lagaris1998artificial,smets2023pde,illarramendi2022performance}.
Another broad class of architectures is autoencoders and variational autoencoders
(VAEs)~\cite{seydi2025prediction,he2020unsupervised,tait2020variational,chen2024pseudo}.
In these architectures, the PDE solution is first mapped into a latent space by an encoder, after
which a decoder transforms this latent representation back into the physical space to obtain the
prediction of the problem.

\paragraph{Generative Models for PDE Solutions}
Generative models shift from deterministic predictions to \emph{distributions} over solutions, which is essential when uncertainty arises from sparse observations, unknown parameters, or intrinsic stochasticity. Three principal classes have been applied to PDE problems.
\emph{Score-based diffusion models} generate solutions by reversing a stochastic noise corruption process and learning the associated score function~\cite{song2020score}. A neural generator $G(u_t, t; \theta)$ is trained to approximate:
\begin{equation}
  G(u_t, t; \theta)
  \;\approx\;
  \nabla_{u_t} \log p(u_t),
  \label{eq:score_diffusion}
\end{equation}
where $u_t$ denotes the noisy intermediate state at diffusion time $t$. The learned score function enables iterative denoising to recover samples from the learned solution distribution $p(u)$. Score-based diffusion models have been applied both to forward surrogate modeling, where they learn mappings from $(x,\xi)$ to the PDE state $u$, and to inverse problems, where posterior sampling over uncertain parameter fields $\xi$ is required~\cite{rozet2023score,lippe2023pde}.
\emph{Flow-matching models} learn continuous normalizing flows that transport a base distribution to solution distributions via a straight-path vector field, achieving comparable sample quality to diffusion with significantly fewer function evaluations~\cite{lipman2022flow}.
\emph{Bayesian neural networks} (BNNs) quantify epistemic uncertainty through posterior
distributions over network weights, enabling calibrated confidence intervals, albeit at the cost
of multiple stochastic forward passes.
These approaches provide native uncertainty quantification but incur higher computational costs: diffusion requires many denoising steps, and BNNs require multiple stochastic passes. See Supplementary Section~\ref{sec:supp_generative} for extended formulations. Furthermore, generic deep neural network architectures can be adapted to time-dependent PDEs by employing recurrent neural networks (RNNs), such as Long Short-Term Memory
(LSTMs)~\cite{gao2025ml,zhang2023deep,sabate2020solving} and Gated Recurrent Units
(GRUs)~\cite{chen2024pgnm,long2024physics}.

\paragraph{Capabilities and Limitations}
Generic deep surrogate architectures, including MLPs, CNNs, and encoder-decoder networks, can
rapidly approximate PDE solutions directly from data, without explicit information about the
underlying governing PDE equations. Although they have a time-consuming training phase, once trained, they can provide very fast predictions~\cite{guo2016convolutional,wandel2020learning, }. This makes these surrogate models interesting for computationally intensive simulations as well as for tasks like sensitivity analysis, uncertainty quantification, and problem optimization, where the time-consuming numerical solver needs to be run many times \cite{mousavi2024machine,esfandi2024effect}. Moreover, these models are well-suited when the physics of the problem or the underlying numerical model is unknown, partially known, or costly to evaluate~\cite{peherstorfer2018survey,willard2020integrating}. When sufficient data are available for a fixed geometry, such generic surrogates can attain high predictive accuracy and serve as a replacement for classical numerical solvers.
Although they are flexible and easy to implement, generic deep learning models exhibit several
limitations when replacing PDE numerical solvers. They typically require very large training
datasets~\cite{karniadakis2021physics,willard2020integrating}, are highly restricted to the
grid resolution on which they are trained~\cite{li2020fourier,lu2021learning}, and cannot be
transferred to new grid systems or resolutions in multi-scale problems~\cite{kovachki2023neural,brandstetter2022message}. Moreover, these models offer no accuracy guarantee beyond the training dataset. In other words, they usually cannot carry over to unseen geometries and boundary conditions that were not represented in the training data. In short, their ability to generalize is limited.

\subsubsection{Graph Neural Networks}
\label{subsubsec:gnns}

GNNs leverage the natural graph structure underlying computational meshes or particle
systems. This approach is well-suited for PDEs defined on irregular geometries and/or unstructured
domains. GNNs naturally handle irregular geometries and unstructured meshes, though information
propagation requires multiple message-passing steps for global effects, and deep networks risk
over-smoothing.
Graph Network-based Simulators (GNSs) introduced the encode-process-decode
framework~\cite{sanchez2020learning}, while MeshGraphNets extended this to mesh-based
simulations with explicit handling of mesh and world-space edges~\cite{pfaff2020learning}:
\begin{equation}
m_{ij}^{(k)} = f_{\text{edge}}\!\left(z_i^{(k)}, z_j^{(k)}, e_{ij}\right), \quad
z_i^{(k+1)} = z_i^{(k)} + f_{\text{node}}\!\left(z_i^{(k)},
               \sum_{j \in \mathcal{N}(i)} m_{ij}^{(k)}\right)
\label{eq:meshgraphnets}
\end{equation}
where $z_i^{(k)}$ denotes the latent feature vector associated with node $i$ at message-passing layer $k$, representing local solution states or physical quantities at a mesh node or a particle. The edge attribute $e_{ij}$ encodes geometric or physical information associated with the interaction between nodes $i$ and $j$ (e.g., relative position, area, or material properties). The function $f_{\text{edge}}$ computes messages $m_{ij}^{(k)}$ passed along edges, while
$f_{\text{node}}$ updates node states by aggregating incoming messages from the neighborhood $\mathcal{N}(i)$. MP-PDE introduces temporal bundling to reduce autoregressive error accumulation~\cite{brandstetter2022message}.
Learned message functions often correspond to discretized differential operators, providing interpretability.
For PDEs defined on domains with known geometric symmetries, rotational and translational invariance in fluid dynamics, or SE(3)-equivariance in molecular simulations, equivariant graph networks enforce these symmetries architecturally, reducing effective sample complexity and improving out-of-distribution
generalization~\cite{thomas2018tensor,satorras2021n,batzner20223}. This connects directly to the hard architectural constraint design dimension discussed in Section~\ref{subsubsec:physics_integration}.

\paragraph{Capabilities and Limitations}
Overall, GNNs sit between purely black-box surrogate models (Section~\ref{surrogate_models}) and fully physics-embedded methods (Section~\ref{sec:physics-embedded_ml}). They offer a good balance between architectural flexibility and interpretable local interactions, which has led to their growing use.
In practice, they are especially effective for problems with complex geometries~\cite{franco2023deep,taghizadeh2025multifidelity} and unstructured meshes~\cite{wang2023graph}.
However, their ability to generalize to unseen data or substantially different parameter ranges
remains a significant challenge~\cite{sanchez2020learning,brandstetter2022message,xu2018powerful,oono2019graph}.
For further information, see Supplementary Section~S.1.2.

\subsubsection{Transformers}
\label{subsubsec:transformers}
Transformers introduce attention mechanisms enabling direct global interactions between spatial
locations~\cite{vaswani2017attention,li2022transformer}:
\begin{equation}
\text{Attention}(Q, K, V) = \text{softmax}\!\left(\frac{QK^T}{\sqrt{d_k}}\right)V
\label{eq:attention}
\end{equation}
where $Q \in \mathbb{R}^{N \times d_k}$, $K \in \mathbb{R}^{N \times d_k}$, and $V \in \mathbb{R}^{N \times d_v}$ denote the query, key, and value matrices, respectively, obtained through linear projections of the input features. Here, $N$ represents the number of spatial locations, grid points (or tokens), while $d_k$ and $d_v$ denotes the dimensionalities of the key and value feature spaces. 
The scaled dot-product $QK^T/\sqrt{d_k}$ measures pairwise similarity between queries and keys. The softmax operation normalizes these scores to produce attention weights that quantify the relative influence of different locations. The resulting weighted combination of values enables global information exchange across the computational domain.
Transformer-based PDE models typically require tokenization of the spatial domain, where tokens
may correspond to grid points or spatial patches of solution fields. Since transformers are permutation invariant, positional information must be explicitly incorporated through learned positional encodings.
In addition to solution variables and spatial coordinates, auxiliary embeddings are often introduced to encode boundary conditions and domain properties, enabling transformers to operate on parameterized domains.
Several transformer variants have been proposed to improve scalability and efficiency in PDE
applications. In the context of PDE modeling, attention weights can be interpreted as approximating Green's
functions or integral kernels~\cite{kovachki2023neural}. Variants include Galerkin Transformers with linear attention, Fourier Transformers operating in spectral space, and factorized attention reducing 3D complexity from $O(N^6)$ to $O(N^2)$.

\paragraph{Large-Scale Foundation Models for PDEs}
An emerging extension of the transformer paradigm is the development of \emph{foundation models} pretrained on diverse PDE families, analogous to large language models in natural language processing (NLP). Approaches such as Poseidon~\cite{herde2024poseidon}, MPP, and UnifiedSolver~\cite{sun2025towards} learn transferable solution operators across multiple PDE types, boundary condition regimes, and geometries. 
Preliminary results demonstrate limited zero-shot and few-shot generalization to unseen PDE
families. However, critical limitations remain: pretraining requires enormous, carefully curated datasets
spanning multiple PDE regimes; distribution shift between pretraining and deployment PDE families degrades generalization quality in ways that are difficult to predict; computational costs are model-dependent but typically very high; and rigorous error bounds for out-of-distribution generalization do not yet exist. Despite these limitations, foundation models represent a qualitative step toward general-purpose PDE solvers and are an active frontier of current research.

\paragraph{Capabilities and Limitations}
Transformers provide global receptive fields from the first layer, since each token can attend
to all others, but face quadratic scaling limitations~\cite{choromanski2020rethinking}.
Therefore, they face challenges in scaling to very high-resolution three-dimensional simulations.
Moreover, transformers can be viewed as a class of neural operator models (discussed in Section~\ref{sec:physics-embedded_ml}), where attention mechanisms learn mappings from input
functions to output functions.
Transformers offer a global view of the solution~\cite{vaswani2017attention,dosovitskiy2020image}
and are highly expressive for capturing non-local interactions in PDEs. However, they can struggle with stability over long time horizons~\cite{brandstetter2022message} and often generalize poorly to unseen resolutions, geometries, or physical regimes, unless augmented with physics-based regularization~\cite{zhuang2023survey,wu2023comprehensive,wen2022transformers}. See Supplementary Section~S.1.3 for further information.

\subsection{Physics-Embedded Methods}
\label{sec:physics-embedded_ml}
Physics-embedded ML methods incorporate governing physical laws directly into the learning
process. Thus, they do not rely solely on data-driven mappings. By enforcing PDE constraints through loss functions, architectural design, or operator formulations, these approaches improve physical consistency. Thus, they can generalize better than other ML approaches. Examples include PINNs and neural operators ; extended architectural details appear in Supplementary Section~\ref{sec:supp_architectures}.
The primary distinction between physics-embedded ML methods and hybrid physics-ML models lies in the role of classical PDE solvers during inference. Physics-embedded approaches incorporate physical laws directly into neural model training and do not use a separate numerical solver at prediction time. However, hybrid models keep classical PDE solvers within the solution loop. 

\subsubsection{Neural Operators}
\label{sec:neural_operators}
\label{subsec:rom_comparison}
Neural operators learn non-linear mappings between function spaces $\mathcal{G}: \mathcal{A} \rightarrow \mathcal{U}$, where inputs $x \in \mathcal{X}$ (e.g., initial and boundary conditions, source terms, material properties, and geometry) are mapped to solution fields over space or time $u \in \mathcal{U}$, i.e., function-to-function mapping ($u = \mathcal{G}(x;\theta)$). They solve PDEs by learning mappings directly between function spaces, which is in contrast to generic deep surrogate models (Section~\ref{surrogate_models}) and PINNs (Section~\ref{sec:pinns}), which approximate pointwise solutions, or classical numerical solvers (Section~\ref{sec:classicmethods}), which discretize differential operators on fixed computational grids. As the learned operator can be applied directly to unseen inputs, new problems can be evaluated without the need for retraining.
It is important to note that neural operators can be trained either purely data-driven or in a physics-informed framework. In this review, they are categorized as surrogate models (Section \ref{surrogate_models}) when the learned mapping directly approximates the solution operator without explicitly enforcing the governing operator during training. If residual-based constraints are incorporated, they conceptually move toward physics-embedded formulations (Section \ref{sec:physics-embedded_ml}). The classification, therefore, depends on the functional role and training strategy rather than the architectural design. 
Neural operators inherit the spirit of classical reduced-order modeling (ROM) methods, which
compress high-dimensional PDE solution manifolds into low-dimensional representations.
In POD-Galerkin methods, the solution is expanded over a set of $r$ empirical modes
$\{\phi_k\}_{k=1}^r$ learned from snapshot data, and the reduced-order dynamics are obtained by
Galerkin projection of the governing equations onto this subspace.
DeepONet can be understood as a nonlinear generalization of this principle: its trunk network constructs a set of $p$ learned basis functions $\{t_k(y;\theta)\}_{k=1}^p$, and the branch network computes 
input-dependent expansion coefficients $\{b_k(u;\theta)\}_{k=1}^p$, mirroring the POD structure but with \emph{nonlinear} bases learned end-to-end from data. Similarly, FNO is structurally analogous to spectral Galerkin methods: 
both operate via truncated spectral expansions, but FNO replaces analytically prescribed Fourier modes 
with learned spectral filters $R(k;\theta)$, enabling it to capture problem-specific spectral 
content beyond the capabilities of fixed-basis methods.
Key differences are that neural operators are entirely data-driven and do not require
projection of the governing equations, making them applicable when the PDE is only partially
known; however, this flexibility comes at the cost of the convergence guarantees and physical
interpretability inherent to projection-based ROMs.

\paragraph{Fourier Neural Operator}
FNO parameterizes integral operators through learnable filters in Fourier space~\cite{li2020fourier}:
\begin{equation}
(K_\theta v)(x) 
= \mathcal{F}^{-1}\!\left(R(k;\theta) \cdot \mathcal{F}(v)(k)\right)(x) 
+ W(\theta)\, v(x)
\label{eq:fno}
\end{equation}
where $v(x)$ denotes the input function evaluated at spatial location $x$, $\mathcal{F}$ and $\mathcal{F}^{-1}$ represent the forward and inverse Fourier transforms, and $k$ denotes the Fourier frequency variable. The learnable tensor $R(k;\theta)$ contains truncated Fourier coefficients parameterized by  $\theta$, enforcing a low-frequency spectral bias, while $W(\theta)$ denotes a learned pointwise linear operator acting in physical space. 
Fast Fourier transforms yield a computational complexity of $O(N \log N)$ for $N$ spatial 
degrees of freedom. Within trained parameter ranges, FNO achieves speedups of several orders 
of magnitude over classical solvers~\cite{li2020fourier}, though this advantage requires 
amortizing substantial training costs (see Section~\ref{sec:economics}).

\paragraph{DeepONet}
DeepONet employs a branch–trunk architecture grounded in the universal approximation theorem
for operators~\cite{lu2021learning,chen1995universal}. Let the trainable parameters be denoted by  $\theta = (\theta^{\mathrm{branch}}, \theta^{\mathrm{trunk}})$. The learned operator is expressed as:
\begin{equation}
\mathcal{G}(u)(y;\theta)
=
\sum_{k=1}^{p}
b_k\!\left(u;\theta^{\mathrm{branch}}\right)\,
t_k\!\left(y;\theta^{\mathrm{trunk}}\right)
\label{eq:deeponet}
\end{equation}
where $u$ denotes the input function, $y$ is the spatial location at which the solution is evaluated, and $p$ is the number of basis functions. The branch network maps discretized samples of $u$ to coefficients $b_k(u;\theta^{\mathrm{branch}})$, while the trunk network produces continuous basis functions $t_k(y;\theta^{\mathrm{trunk}})$ evaluated at arbitrary query points.
This formulation, the same as FNO, enables mesh-free evaluation of solutions and flexible handling of multiple input functions. However, prediction accuracy depends sensitively on the placement and number of sensors used to sample the input function.

\paragraph{Graph Neural Operator}
GNOs extend operator learning to irregular domains using message passing on graph discretizations~\cite{li2020neural}. They combine neural operators (Section~\ref{sec:neural_operators}) with GNN architectures,
enabling PDE solution learning on unstructured meshes.
Thus, GNOs maintain discretization invariance and can generalize across different mesh
topologies and resolutions.
A typical layer update is given by:
\begin{equation}
v^{(l+1)}_i 
=
\sigma\!\left(
W_1^{(l)}(\theta)\, v^{(l)}_i
+
\sum_{j \in \mathcal{N}(i)} 
\kappa(x_i, x_j;\theta)\,
W_2^{(l)}(\theta)\, v^{(l)}_j
\right)
\label{eq:gno}
\end{equation}
where $v_i^{(l)}$ denotes the feature vector at node $i$ and layer $l$, $x_i$ and $x_j$ represent the spatial coordinates of nodes $i$ and $j$,  $\mathcal{N}(i)$ denotes the neighborhood of node $i$, and $\sigma$ is a nonlinear activation function. The learnable kernel $\kappa(x_i, x_j;\theta)$ encodes geometric relationships between nodes, while $W_1^{(l)}(\theta)$ and $W_2^{(l)}(\theta)$ denote trainable linear transformations.  Here, $\theta$ collects all learnable parameters of the graph neural operator.
Several extensions have been proposed to address specific limitations of neural operators. Wavelet Neural Operators provide multi-resolution decomposition for multiscale problems~\cite{tripura2023wavelet}; U-FNO combines U-Net hierarchical processing with Fourier layers~\cite{wen2022accelerating}; Geometry-Informed Neural Operators incorporate geometric embeddings for complex domains~\cite{li2023geometry}. Foundation models pretrained on diverse PDE families represent an emerging direction with preliminary zero-shot generalization results~\cite{herde2024poseidon,sun2025towards}. See Supplementary Section~S.1.1 for further details.

\paragraph{Capabilities and Limitations}
Neural operators, including FNO, DeepONet, and GNO, are resolution-independent and efficient in
parametric evaluation of problems, even without need for retraining~\cite{li2020fourier,lu2021learning,kovachki2023neural,li2020neural}.
This capability makes these models well suited for parametric sensitivity analysis, inverse problem modeling, problem optimization, and especially uncertainty quantification. FNO captures correlations through spectral Fourier convolution, DeepONet provides nonlinear operator approximation via its branch-trunk architecture, and GNO extends operator learning to irregular unstructured meshes using graph-based message passing. Thus, neural operators have the potential to replace classical numerical models on multi-scale, multi-physics coupled PDE problems. When neural operators fully replace the numerical solver, they are considered as pure surrogate models. In contrast, if the learned operator is embedded within a classical PDE solver, the resulting formulation falls under hybrid physics-ML models. 
Despite these advantages, neural operators are still data-driven models. Their predictive performance  heavily depends on the availability and coverage of the training dataset. Training neural operators typically requires many input-output function pairs  (e.g., $10^3$-$10^5$). Compared to classical POD-based ROMs, neural operators require more training data but achieve superior expressiveness for strongly nonlinear problems where POD subspaces require prohibitively many modes for accurate representation. While neural operators generally exhibit improved resolution generalization compared to other pointwise surrogate models, their accuracy degrades for solutions with sharp gradients or discontinuities~\cite{lu2022comprehensive}.
Moreover, in contrast to PINNs (Section~\ref{sec:pinns}), these models do not enforce physical constraints derived from PDEs during training. Finally, training can be computationally expensive for high-resolution three-dimensional computational problems, and interpretability remains limited compared to classical numerical solvers~\cite{kovachki2023neural,li2020neural}.

\subsubsection{Physics-Informed Learning}
\label{sec:pinns}
PINNs embed governing equations directly into neural network training, enabling solution approximation with minimal simulation data~\cite{cuomo2022scientific,karniadakis2021physics}. Instead of relying purely on data-driven mapping of inputs to outputs in other subcategories of ML approaches, PINNs enforce PDEs as soft constraints during training. Detailed formulations of this approach are provided in Supplementary Section~\ref{sec:supp_pinns}.
The standard PINN formulation parametrizes the solution field $u(x,t)$ as a machine learning models $u(x,t;\theta)$ with trainable parameters $\theta$.  Training is performed by minimizing a composite loss function~\cite{lawal2022physics,cai2021physics}:
\begin{equation}
\mathcal{L}(\theta) = \lambda_{\text{BC}}\mathcal{L}_{\text{BC}}
                    + \lambda_{\text{IC}}\mathcal{L}_{\text{IC}}
                    + \lambda_{\text{PDE}}\mathcal{L}_{\text{PDE}}
                    + \lambda_{\text{data}}\mathcal{L}_{\text{data}}
\label{eq:pinn_loss}
\end{equation}
where $\mathcal{L}_{\text{BC}}$ and $\mathcal{L}_{\text{IC}}$ penalize violations of boundary and initial conditions, $\mathcal{L}_{\text{PDE}}$ enforces the governing PDE, and $\mathcal{L}_{\text{data}}$ uses available simulation data in the training dataset. The coefficients $\lambda_{\text{BC}}$, $\lambda_{\text{IC}}$, $\lambda_{\text{PDE}}$, and
$\lambda_{\text{data}}$ are weighting parameters that balance the different loss contributions.
For example, the PDE residual loss is defined by enforcing the governing operator $F_{\mathrm{PDE}}$ at a set of collocation points:
\begin{equation}
\mathcal{L}_{\mathrm{PDE}}(\theta) 
= \frac{1}{N_r}\sum_{i=1}^{N_r}
\left|
F_{\mathrm{PDE}}\!\left(u(x_i,t_i;\theta)\right)
\right|^2
\label{eq:pde_residual}
\end{equation}
where $N_r$ is the number of collocation points and $\{(x_i,t_i)\}_{i=1}^{N_r}$ are sampled from the spatio-temporal domain.
Automatic differentiation enables continuous evaluation of spatial and temporal derivatives to
machine precision, eliminating discretization errors inherent to finite-difference approximations.
Table~\ref{tab:pinn_variants} summarizes principal PINN variants addressing specific computational challenges. Further details of each variant can be found in Supplementary Section~\ref{sec:supp_pinns}.

\begin{table}[h!]
\centering
\caption{PINN variants: modifications and applications.}
\label{tab:pinn_variants}
\begin{tabular}{p{3cm}p{5cm}p{4cm}p{2.5cm}}
\toprule
\textbf{Variant} & \textbf{Key Modification} & \textbf{Use Case} & \textbf{Reference} \\
\midrule
Variational (VPINN) & Weak formulation via energy functional & Problems with variational structure & \cite{kharazmi2019variational,rojas2024robust} \\
Extended (XPINN) & Domain decomposition, interface conditions & Complex geometries, parallelization & \cite{jagtap2020extended,hu2021extended} \\
Conservative & Architectural conservation enforcement & Long-time integration, transport & \cite{jagtap2020conservative} \\
Deep Ritz & Energy minimization for elliptic PDEs & Variational problems & \cite{ji2024deep,xu2024refined} \\
Weak Adversarial & Min-max formulation avoiding high derivatives & High-order PDEs & \cite{zang2020weak,oliva2022towards} \\
Multi-fidelity & Hierarchical data fusion & Mixed-quality data sources & \cite{meng2020composite,taghizadeh2024multi} \\
Adaptive & Residual-based sampling refinement & Solutions with localized features & \cite{wu2023comprehensive,torres2025adaptive} \\
\bottomrule
\end{tabular}
\end{table}

\paragraph{Capabilities and Limitations}
PINNs offer a mesh-free formulation by approximating the solution via neural networks at
specific sampled collocation points within the problem domain~\cite{raissi2019physics,karniadakis2021physics}. This approach eliminates explicit geometric discretization in classical numerical approaches (FDMs, FEMs, and FVMs) and provides favorable scaling to multi-physics multi-dimension problems.
The computational cost of these methods scales with the number of collocation points (typically
$O(10^3$-$10^5)$) rather than with grid size ($O(N^d)$) in classical numerical approaches (Section~\ref{sec:classicmethods}). Moreover, because of incorporating physical constraints during training, they usually require less training data compared to other ML techniques~\cite{raissi2019physics,karniadakis2021physics,cuomo2022scientific,willard2020integrating}.
However, these physics-informed models have a complex optimization loss function
(Eq.~\ref{eq:pinn_loss}), in which competing loss terms of PDE residuals, boundary conditions,
and data point terms generate conflicting gradient signals~\cite{wang2022and,wang2021understanding}. This pathological training behavior can be analyzed through the lens of the Neural Tangent Kernel (NTK). Wang et al.~\cite{wang2021understanding} showed that stiff PDEs cause dramatically different NTK eigenvalue magnitudes for different loss components, leading to systematic imbalance during gradient descent.
This theoretical insight motivates the use of adaptive loss weighting schemes that normalize
loss contributions by their respective NTK spectra, though a fully principled solution remains
an open problem. In addition, spectral bias leads to preferential learning of low-frequency components, limiting performance for sharp gradients and boundary layers~\cite{rahaman2019spectral}. As a result, accuracy typically remains moderate compared to spectral methods for smooth problems.
Recent developments have begun to address these persistent challenges. Curriculum learning strategies progressively increase collocation point complexity, adaptive loss weighting schemes balance competing objectives using gradient statistics, and causal training approaches for time-dependent problems enforce temporal causality to prevent information leakage~\cite{toscano2025pinns}. Together, these advances reflect the evolution of PINNs from proof-of-concept models toward increasingly practical PDE solvers, although systematic design guidelines remain elusive. See Supplementary Section~S.2.2 for a detailed analysis of training dynamics.

\subsection{Hybrid Methods}
\label{sec:hybrid_ml}

The data-driven models (discussed in Sections \ref{surrogate_models} and \ref{sec:physics-embedded_ml}), although computationally efficient, still require a large amount of data, are difficult to generalize beyond the training domain, and lack physical consistency. In contrast, classical PDE solvers are based on strong theoretical foundations and can enforce conservation laws; however, they are computationally expensive.

To leverage both approaches, hybrid models use ML when and where necessary. They retain the core physical aspects of the scientific problem (i.e., PDEs) while adding data-driven models to accelerate the solution. Thus, hybrid models can be considered a robust and practical choice for solving scientific problems, as they combine the strengths of classical physics-based solvers and data-driven models while mitigating the inherent limitations of each approach.

\begin{figure}[t]
    \centering
    \includegraphics[width=0.95\textwidth]{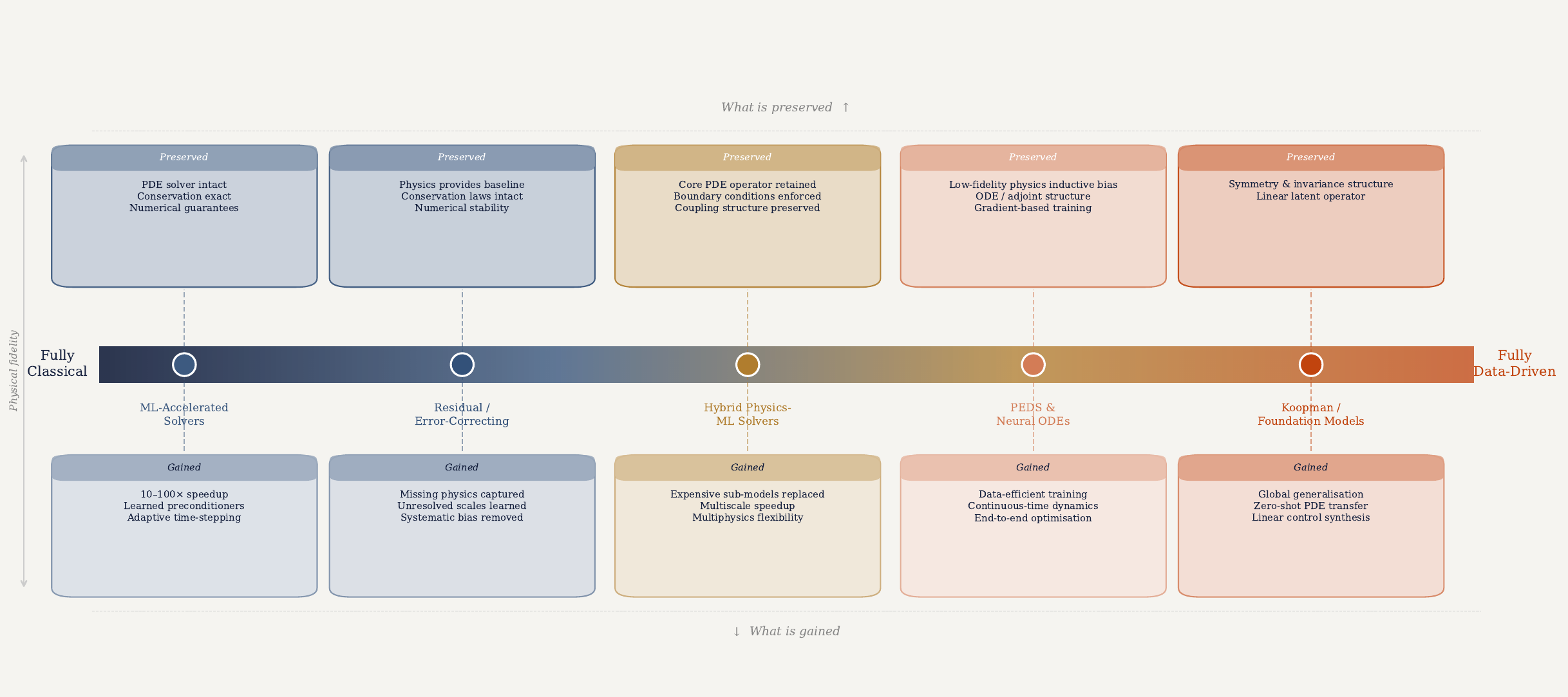}
    \caption{Spectrum of hybrid ML--PDE methods, ordered by retained physical fidelity (left: fully classical solvers; right: fully data-driven surrogates). Five representative strategy families are positioned along the axis:
    - \emph{ML-accelerated solvers}: preserve the full PDE operator, augmenting only numerical kernels with learned components for speed;
    - \emph{residual / error-correcting models}: maintain physics baseline while learning discrepancy terms for unresolved scales or missing closures;
    - \emph{hybrid physics-ML solvers}: replace costly sub-models (e.g., constitutive laws, fine-scale homogenization, geochemical modules) inside a structure-preserving PDE framework;
    - \emph{physics-enhanced deep surrogates \& neural differential equations} (PEDS, Neural ODEs): leverage low-fidelity physics as inductive bias, expressing dynamics via adjoint-friendly continuous architectures;
    - \emph{Koopman and foundation models}: linearize nonlinear dynamics in learned latent spaces or enable transfer across PDE families with minimal retraining.
    The upper boxes list provably preserved physical structures (conservation laws, symmetries, boundary conditions); the lower boxes highlight key computational gains over classical or purely data-driven baselines.}
    \label{fig:hybrid_spectrum}
\end{figure}

Figure~\ref{fig:hybrid_spectrum} depicts the design space as a continuous spectrum from purely classical to fully data-driven solvers, locating five representative hybrid strategy families by their retained level of physical fidelity. The spectrum exposes a fundamental trade-off: greater reliance on ML for physics modeling progressively sacrifices classical guarantees---exact conservation, certified boundary-condition enforcement, and numerical stability---in exchange for superior speed, broader generalization across parameter regimes, and the capacity to capture complex or incompletely understood processes that resist closed-form PDE representation. No single position on this spectrum is optimal universally. The ideal degree of hybridization depends on data availability, the fidelity of known physics, the query volume of the trained model, and the acceptable risk of unverified extrapolation. The following subsections examine each family in detail, ranging from the most conservative (ML-accelerated solvers, Section~\ref{subsubsec:ML_accelerated_solvers}) to the most expressive (Koopman and foundation models, Section~\ref{subsubsec:Neural_Differential_Models}).

This category represents a balance between purely data-driven surrogate ML models and fully physics-based methods. Instead of replacing the entire set of governing PDEs, these approaches selectively combine numerical solvers with data-driven models to enhance efficiency while preserving physical consistency. The hybrid category can be subclassed into different strategies (see Figure \ref{fig:ML_classification}). Hybrid physics-ML solvers (Section \ref{subsubsec:Hybrid_physics_ML_Solvers}) replace selected physical components or scales within a classical PDE framework. ML-accelerated solvers (Section \ref{subsubsec:ML_accelerated_solvers}) focuses on enhancing the numerical solution part. Residual, or error-correcting, models (Section \ref{subsubsec:Residual_ML_Models}) learn discrepancy terms that account for unresolved physical processes, whereas neural differential equation models (Section \ref{subsubsec:Neural_Differential_Models}) incorporate continuous-time dynamics into a trainable representation that stays connected to traditional numerical solvers. These approaches illustrate how ML can be integrated into PDE-based simulations without replacing their physical foundations.

\subsubsection{Hybrid Physics-ML Solvers}
\label{subsubsec:Hybrid_physics_ML_Solvers}
Hybrid physics-PDE solvers can be applied to scientific problems in which either at least a  governing PDE subsystem or specific components of it (e.g., scale- or physics-related terms,  source terms) are replaced by an ML model, while the remaining part of the physics is solved using traditional numerical approaches. Instead of approximating the entire solution with an ML approach (like the models in Section \ref{surrogate_models}), hybrid approaches retain parts of the physics-based solver and substitute computationally expensive,  highly nonlinear, or poorly resolved components of physical models. In other words, the hybrid system still enforces conservation laws and boundary conditions through the remaining physics-based solvers.
A general hybrid physics-ML system can be written as,
\begin{equation}
\mathcal{F}_{\text{PDE}}\!\left(
u,\,
\mathcal{M}_{\text{ML}}(z;\theta)
\right)
= 0,
\label{eq:hybrid_general}
\end{equation}
In Eq.~\eqref{eq:hybrid_general}, $u$ denotes the PDE  variable, $\mathcal{F}_{\text{PDE}}$ is the retained PDE operator, $\mathcal{M}_{\text{ML}}(\cdot;\theta)$ is the ML model parameterized by trainable weights $\theta$, and $z$ represents the input to the ML model.
By replacing only selected PDE components or subsystems rather than the entire system, hybrid surrogate-PDE solvers reduce computational cost while preserving interpretability and numerical robustness. They offer a balance  between fully data-driven approaches and direct PDE-based simulations, which makes them especially well suited for complex coupled systems. 

\paragraph{Hybrid Multi-Physics Surrogate Embedding}
Hybrid physics-PDE solvers can be widely applied in multi-physics problems in which the traditional nonlinear couplings of underlying PDEs have very high computational cost. A very good example is reactive transport modeling in porous media, where flow and transport equations are nonlinearly coupled with geochemical reactions through porous media. The traditional approach couples the transport step with a costly geochemical equilibrium or kinetic calculation. Recent hybrid surrogate-PDE solvers can replace the geochemical equilibrium module with trained neural networks that predict the reaction part of the model \cite{silva2024rapid, qiao2025hidden}. 
This hybrid approach significantly reduces computational time while maintaining the physical consistency of the transport model. Similar ideas have been explored in other areas of computational science, where learned components replace selected physical processes within a larger PDE framework. Examples include turbulent flow modeling \cite{ling2016reynolds, beck2019deep},  climate modeling \cite{rasp2018deep, brenowitz2018prognostic}, and computational solid mechanics \cite{klein2022polyconvex}.

\paragraph{Hybrid Multiscale Surrogate Embedding}
In multiscale problems, fine-scale problems must be solved at each macroscale integration point, leading to substantial computational cost. Hybrid physics-ML solvers replace this repeated fine-scale computation with a learned surrogate model,
\begin{equation}
y = \mathcal{M}_{\text{ML}}(x,\,\xi;\,\theta),
\label{eq:general_homogenization}
\end{equation}
where $x$ is the macroscale input variables, $y$ represents the corresponding macroscale response, $\xi$ is the  fine-scale descriptors , $\theta$ is the trainable parameters of the ML model, and $\mathcal{M}_{\text{ML}}$ is the learned mapping that approximates the classical fine-scale-to-coarse-scale operator.
Thus in multiscale problems, fine-scale processes can be learned from high-resolution data through surrogate models and subsequently integrated into macro-scale PDE solvers. Such strategies are widely used in upscaling and uncertainty quantification applications, where learned surrogates approximate fine-scale behavior while maintaining the governing macro-scale equations \cite{mo2019deep, moslemipour2025multi, zhu2018bayesian, jiang2023upscaling, menke2021upscaling}. The neural network can be trained offline and then evaluated online during direct solving of PDEs at macroscale. This approach preserves the macroscopic PDE structure while achieving orders-of-magnitude speedup compared to classical nested FEM~\cite{liu2023hierarchical}.  

\paragraph{Physics-Enhanced Deep Surrogates (PEDS).}
Physics-Enhanced Deep Surrogates (PEDS) are a subclass of hybrid solvers in which a neural generator is combined with a low-fidelity PDE physics solver~\cite{pestourie2023physics}. Instead of directly approximating the solution field, the ML model learns to generate solver inputs that yield accurate outputs after passing through a physics-based operator:
\begin{equation}
\tilde{u}
=
\mathcal{S}_{\text{LF}}\big(G(x;\theta)\big)
\label{eq:peds}
\end{equation}
where $G(x;\theta)$ denotes a neural generator parameterized by the trainable weights $\theta$, $x$ represents the problem input variables, and $\mathcal{S}_{\text{LF}}$ is a low-fidelity PDE solver encoding simplified but physically consistent dynamics. The generator learns to produce solver inputs yielding correct outputs rather than directly
approximating solutions.

\subsubsection{ML-Accelerated Solvers}
\label{subsubsec:ML_accelerated_solvers}
The aim of ML-accelerated solvers is to improve the computational efficiency of classical PDE solvers without modifying the underlying physical model. These hybrid approaches focus specifically on the numerical solver component rather than on the physical formulation of the scientific problem. In other words, ML is used to enhance numerical performance rather than replace physical PDEs.  
A very good example is the use of learned preconditioners or data-driven components to accelerate iterative linear and nonlinear solvers. Other examples include strategies that establish scale‑compatible and flexible interface relations enabling coherent information transfer between 1D and 3D domains~\cite{von2024machine}. Neural preconditioners learn approximate inverses for iterative solvers~\cite{luz2020learning}. Consider a linear system  ${A x = b}$, where $A \in \mathbb{R}^{n \times n}$ denotes the system matrix arising from the discretization of the underlying PDEs, $x \in \mathbb{R}^{n}$ is the unknown solution vector, and $b \in \mathbb{R}^{n}$ is the right-hand side vector. Neural preconditioners aim to learn an approximation of the inverse of $A$,
\begin{equation}
M_{\text{NN}}^{-1} \approx A^{-1}, \qquad
\kappa(M_{\text{NN}}^{-1}A) \ll \kappa(A),
\label{eq:neural_preconditioner}
\end{equation}
where $M_{\text{NN}}^{-1}$ denotes the learned preconditioning operator and $\kappa(\cdot)$ represents the spectral condition number of a matrix. By reducing the condition number of the preconditioned system $M_{\text{NN}}^{-1}A$, the number of Krylov iterations required for convergence is significantly decreased. This approach is particularly effective for problems with highly heterogeneous coefficients, where classical algebraic multigrid preconditioners often degrade in performance.
Beyond preconditioning, ML models have been proposed to predict adaptive time-step sizes based on solution dynamics, enabling larger stable steps while maintaining accuracy \cite{liu2022hierarchical, dellnitz2023efficient,riley2025reinforcement}.  Similarly, mesh strategies can be guided by ML approaches for local refinement or coarsening decisions based on learned error indicators, thereby reducing computational expense for multi-scale, multi-physics problems~\cite{chen2026neural, tingfan2022mesh, sharma2025accelerating}. Additional applications include learning improved initial guesses for nonlinear solvers, accelerating multigrid cycles, and predicting solver convergence behavior \cite{greenfeld2019learning,hsieh2019learning}. In all of the cases described above, the underlying PDEs and numerical scheme are left unchanged; ML is used purely as an accelerator within the solver framework to enhance runtime performance.

\subsubsection{Residual and Error-Correcting ML Models}
\label{subsubsec:Residual_ML_Models}
Residual ML models, also called  error-correcting models, offer a practical way to improve classical PDE solvers without changing their underlying physics. When a scientific problem has complex, missing, or unknown physics; unresolved scale; or systematic numerical errors (bias), this class of hybrid models can be used. They learn correction terms to PDE  solution via ML models. The physics-based solver (i.e., PDE) still provides the main prediction and ensures conservation laws as well as boundary conditions are satisfied, while the ML part tries to estimate the remaining difference between the PDE model prediction and supervised high-resolution solutions during the training phase. In this way, the overall framework becomes more accurate without sacrificing stability or interpretability.  
Let the classical PDE model be written in operator form as $\mathcal{F}_{\text{PDE}}(u, x) = 0$, where $\mathcal{F}_{\text{PDE}}$ denotes the known governing differential operator, $u$ is the primary PDE variable, and $x$ represents input variables. In residual correction approaches, the model is augmented by a learned discrepancy term:
\begin{equation}
\mathcal{F}_{\text{PDE}}(u, x)
+
\mathcal{M}_{\text{ML}}(u, x;\theta)
= 0,
\label{eq:residual_correction}
\end{equation}
where $\mathcal{M}_{\text{ML}}(u, x;\theta)$ denotes an ML-based correction parameterized by the trainable weights $\theta$. The classical PDE structure remains unchanged, while the learned residual term improves predictive accuracy by approximating unresolved scales, missing physics, or modeling errors.
For example, physics-informed supervised residual learning has been used to iteratively refine electromagnetic simulations \cite{shan2023physics}. In inverse scattering, researchers have begun adding learned update steps into traditional iterative methods to fix errors in the forward model, as shown by the neural Born iteration approach \cite{shan2022neural}. Similar concepts have been extended to computational fluid dynamics, where machine learning-CFD hybrid methods have been proposed to speed up long-term unsteady flow simulations \cite{jeon2024residual}. These studies demonstrate that learning correction terms provide a practical way for systematically improving predictive performance.
A closely related category within residual and error-correcting ML methods involves \textit{refinement learning}. In such model-form correction methods, implemented ML models learn closure terms to address missing or simplified physic ~\cite{duraisamy2019turbulence}. Refinement-based hybrids, such as DeePoly~\cite{liu2025deepoly}, combine neural approximation with high-order polynomial corrections. The neural network first captures global solution features with moderate accuracy, and the remaining approximation error is subsequently reduced through spectral polynomial refinement:
\begin{equation}
u(x) = u_{\text{ML}}(x; \theta) + \sum_{k=0}^{p} c_k\, P_k(x),
\label{eq:deepoly}
\end{equation}
where $u(x)$ denotes the solution field of the PDE at spatial location $x$. The term $u_{\text{ML}}(x; \theta)$ represents the neural-network-based approximation with trainable parameters $\theta$. The functions $P_k(x)$ are predefined polynomial basis functions of order $k$, and $c_k$ are the corresponding coefficients associated with each basis function. The integer $p$ denotes the highest polynomial order.
By learning only correction or refinement components rather than replacing the entire PDE system, hybrid physics-ML solvers provide a balanced compromise between fully data-driven models and classical numerical simulation. They retain interpretability and numerical robustness while achieving improved accuracy and computational efficiency, making them particularly attractive for complex coupled and multiscale systems.

\subsubsection{Neural Differential Models}
\label{subsubsec:Neural_Differential_Models}

Differentiable physics enables gradient-based optimization through simulations.  Adjoint methods compute gradients efficiently for time-dependent problems, while implicit differentiation handles iterative solvers via the implicit function theorem~\cite{blondel2022efficient}. 
Let the governing equation be written as $F_{\mathrm{PDE}}\!\left(u^*;\theta\right)=0$, implicit differentiation yields:
\begin{equation}
\frac{\partial u^*}{\partial \theta}
=
-\left(
\frac{\partial F_{\mathrm{PDE}}}{\partial u}
\right)^{-1}
\frac{\partial F_{\mathrm{PDE}}}{\partial \theta},
\label{eq:implicit_diff}
\end{equation}
provided that the Jacobian $\partial F_{\mathrm{PDE}}/\partial u$ is non-singular. 
This formulation enables inverse design, parameter estimation, and end-to-end training of neural components embedded within numerical solvers. Gradients through the solver can be computed via adjoint methods, keeping memory cost independent of solver depth.
Neural Ordinary Differential Equations (Neural ODEs) model hidden-state dynamics as a continuous-time ODE parameterized by a neural network 
$M_{\mathrm{ML}}(\cdot;\theta)$~\cite{chen2018neural}:
\begin{equation}
\frac{d z(t)}{dt}
=
M_{\mathrm{ML}}\!\left(z(t), t; \theta\right),
\qquad
z(t_1)
=
z(t_0)
+
\int_{t_0}^{t_1}
M_{\mathrm{ML}}\!\left(z(t), t; \theta\right)\, dt,
\label{eq:neural_ode}
\end{equation}
where $z(t)$ denotes the latent state and the integral is evaluated using an adaptive ODE solver. 
The key advantage over standard recurrent architectures is adaptive time-stepping: the ODE solver automatically adjusts step sizes to maintain accuracy without requiring a fixed temporal discretization.
Extensions to PDE settings include Neural PDE models, where selected terms in the spatial differential equation are learned from data~\cite{brandstetter2022message}, and Latent ODE approaches, where a low-dimensional latent trajectory captures the essential dynamics of a high-dimensional PDE solution field. 
Continuous normalizing flows (CNFs) further extend this framework to density evolution by modeling the transformation of probability distributions through differential dynamics.
A limitation of neural differential equation models is that implicit coupling between time steps can make parallelization more challenging than in explicit recurrent models. However, recent parallel-in-time training schemes partially address this issue~\cite{gunther2020layer,betcke2024parallel}.
Neural differential equation models parameterize parts of the governing dynamics using a neural network. 
In this formulation, the evolution of the state variable is described by a differential equation in which certain terms are replaced or augmented by a learned component. Rather than enforcing the full governing operator as a residual constraint during training, these models learn unknown or unresolved dynamics directly from data within the differential equation itself. 
While physics-based structure may still guide the formulation (e.g., through known conservation laws or partially specified operators), the governing equation is not necessarily imposed as a hard constraint. 
The defining characteristic of neural differential equation models is therefore the replacement or augmentation of system dynamics by a trainable neural component embedded within the differential equation.

\paragraph{Koopman Operator Methods}
Koopman operator theory offers an alternative linearization strategy for nonlinear PDE dynamics.
The Koopman operator $\mathcal{K}$ is an infinite-dimensional linear operator that acts on observable functions $g$ of the system state $u$:
\begin{equation}
(\mathcal{K} g)(u) = g\!\left(\Phi(u)\right),
\label{eq:koopman}
\end{equation}
where $\Phi$ denotes the (generally nonlinear) flow map of the dynamical system. 
Although $\Phi$ may be nonlinear, $\mathcal{K}$ is linear; the nonlinearity is lifted into an infinite-dimensional function space of observables.
In practice, Extended Dynamic Mode Decomposition (EDMD) and deep Koopman methods~\cite{lusch2018deep,li2017extended} learn a finite-dimensional approximation of the Koopman operator by identifying a set of $m$ observable functions through an encoder $\Psi(u;\theta): \mathcal{U} \rightarrow \mathbb{R}^m$,  such that the lifted dynamics are approximately linear:
\begin{equation}
\Psi(u_{t+1};\theta)
\;\approx\;
K(\theta)\,\Psi(u_t;\theta),
\qquad
K(\theta) \in \mathbb{R}^{m \times m},
\label{eq:koopman_approx}
\end{equation}
where $u_t$ denotes the system state at time $t$ and $K(\theta)$ is a learned finite-dimensional approximation of the Koopman operator. For PDE applications, the spatial field $u(\cdot,t)$ is treated as the system state, and the encoder–decoder architecture is often implemented using convolutional neural networks. 
Koopman-based surrogate models yield linear latent dynamics, enabling efficient prediction via repeated matrix multiplication or matrix exponentiation rather than explicit nonlinear time-stepping. 
They also facilitate linear control design for systems governed by nonlinear PDEs. 
A principal limitation is that strongly nonlinear dynamics (e.g., turbulence or sharp propagating fronts) may require prohibitively large embedding dimensions $m$ to obtain an accurate linear representation~\cite{brunton2021modern}. CoSTA (Corrective source term approximation) employs a linear high-fidelity FEM solver with trained residual corrections based on combining structure preservation and accuracy~\cite{blakseth2022deep,keilegavlen2024porotwin}.

\subsubsection{Selection Guidance}

Table~\ref{tab:hybrid_selection} provides method selection guidelines based on application
requirements.

\begin{table}[h!]
\centering
\caption{Hybrid and physics-coupled method selection guidelines.
The entries in this table are not mutually exclusive, as many applications benefit from combining multiple strategies.}
\label{tab:hybrid_selection}
\begin{tabular}{p{4.8cm}p{4.2cm}p{5.5cm}}
\toprule
\textbf{Application Scenario} & \textbf{Recommended Approach} & \textbf{Rationale} \\
\midrule
Unknown or complex material constitutive behavior
  & Neural constitutive models
  & Preserves equilibrium; learns material response from data while satisfying thermodynamic constraints \\[4pt]
Repeated microscale homogenization at many macroscale points
  & Homogenization surrogates (FE$^2$ replacement)
  & Amortizes expensive offline microscale training across online macroscale queries \\[4pt]
Scarce training data with known governing equations
  & PEDS (physics-enhanced deep surrogate)
  & Low-fidelity physics solver provides strong inductive bias, reducing labeled data requirements by 10--100$\times$ \\[4pt]
High accuracy required for smooth solutions
  & DeePoly (neural + polynomial refinement)
  & Achieves near-spectral convergence through polynomial correction; applicable to smooth problems only \\[4pt]
Parameter estimation or design optimization from observations
  & Differentiable physics
  & Enables gradient-based optimization through the solver via adjoint or implicit differentiation \\[4pt]
Highly heterogeneous or high-contrast coefficients
  & Neural preconditioners
  & Learns approximate inverse operators that reduce Krylov iteration count in regimes where algebraic multigrid may degrade \\[4pt]
Continuous-time dynamics with irregular or adaptive time sampling
  & Neural ODEs / Neural PDEs
  & Adaptive ODE solver adjusts step size automatically; adjoint-based gradient computation is memory-efficient \\[4pt]
Nonlinear dynamical systems requiring linear analysis or control design
  & Koopman operator methods
  & Globally linear latent representation enables matrix-exponential prediction and linear control synthesis with interpretable eigenstructure \\
\bottomrule
\end{tabular}
\end{table}

\paragraph{Advantages and Limitations}
Conservation guarantees are preserved exactly by physics solvers.
Data efficiency improves 10--100$\times$ through physics-based inductive bias.
Graceful degradation occurs when learned components perform poorly---physics provides reasonable
baselines.
Integration with established frameworks enables standard verification procedures.

Implementation requires expertise spanning numerical methods and machine learning.
Combined errors from discretization and neural approximation may interact non-additively.
Automatic differentiation through solvers requires memory for intermediate states.
Learned corrections may degrade outside training conditions.

\subsection{Design Dimensions and Method Selection}
\label{sec:design_dimensions}

ML models for solving computational scientific problems span a four-dimensional design space
whose axes determine computational complexity, data requirements, accuracy characteristics, and
problem applicability:

\subsubsection{Input Encoding Strategies}
\label{subsubsec:input_encoding}

The manner in which solution inputs (i.e., initial and boundary conditions, system domain
parameter fields, source terms, and geometry) are encoded fundamentally constrains the
flexibility of the method and its computational scalability.
\emph{Regular grid representations} (e.g., CNNs, U-Nets, ResNets, Transformers, FNOs) exploit
structured topology for efficient computation through convolution, achieving $O(N \log N)$ for
spectral methods or $O(N)$ for convolutions, but require rectangular domains or coordinate
transformations introducing the geometric complications discussed in
Section~\ref{subsec:geometry}.
\emph{Unstructured mesh representations} (e.g., GNNs, GNOs) naturally handle complex
geometries through message passing on arbitrary graph connectivity, incurring $O(|V| + |E|)$
cost where edge count $|E|$ depends on mesh topology.
\emph{Point cloud representations} (e.g., PINNs, DeepONet) eliminate connectivity requirements
entirely by evaluating solutions at arbitrary locations via coordinate-based networks, though
they may require more sophisticated architectures to capture spatial relationships.
\emph{Sensor-based representations} (e.g., DeepONets) encode functions through values at fixed
observation points, enabling integration of heterogeneous data sources but requiring careful
sensor placement to capture solution features adequately.

The representation choice directly impacts which Section~\ref{sec:challenges} challenges the
method can address: regular grids excel for periodic problems and smooth solutions, but struggle
with complex geometries; unstructured representations handle geometric complexity naturally but
face greater implementation complexity; mesh-free approaches offer maximum geometric flexibility
but may require careful architecture design to achieve competitive accuracy.

\subsubsection{Output/Solution Representation}
\label{subsubsec:output_rep}

The solution parameterization determines expressiveness, evaluation flexibility, and accuracy
characteristics.
\emph{Pointwise predictions} (e.g., CNNs, U-Nets, ResNets, Transformers, PINNs) output
solution values at input locations, requiring re-evaluation for different query points but
enabling straightforward training through supervised regression.
\emph{Global basis coefficient predictions} (FNOs) exploit low-dimensional solution structure
when it exists, achieving efficient representation for smooth solutions
(Section~\ref{subsec:discontinuities} on regularity), but face challenges for discontinuous or
highly irregular solutions.
\emph{Learned operator representations} (e.g., DeepONet, FNO, GNO) enable
resolution-invariant evaluation, i.e., trained models generalize across discretizations, but
require function-space datasets for training, raising the data generation costs discussed in
Section~\ref{sec:economics}.

The output representation determines the method's ability to handle the multiscale phenomena of
Section~\ref{subsec:multiscale}: global basis representations naturally capture smooth
multiscale structure via hierarchical coefficients, while pointwise representations require
architectural features (e.g., wavelets, hierarchical networks) to efficiently represent scale
separation.

\subsubsection{Physics Integration Mechanisms}
\label{subsubsec:physics_integration}

How physical laws constrain learned representations fundamentally affects data efficiency,
extrapolation capability, and satisfaction of conservation principles.
\emph{Soft physics constraints} incorporated through loss function penalties (e.g., PINN,
residual-based DeepONets, variational PINNs) enable flexible optimization but provide no
guarantees---learned solutions may violate conservation laws or boundary conditions if the
penalty terms are not properly balanced.
\emph{Hard architectural constraints} (divergence-free networks, symplectic architectures,
conservative convolution, equivariant GNNs) ensure exact satisfaction of
specific physical properties (incompressibility, energy conservation, symmetries) regardless of
optimization outcome, preserving the structure-preserving properties identified as critical in
Section~\ref{subsec:multiphysics}.
\emph{Purely data-driven approaches} (e.g., CNNs, U-Nets, Transformers) make minimal physical
assumptions, maximizing flexibility for problems with unknown or approximate governing
equations, but requiring extensive training data and lacking physical interpretability.

Physics encoding directly addresses the robustness requirements for coupled systems
(Section~\ref{subsec:nonlinearity}): hard constraints maintain physical feasibility during
optimization, while soft constraints face the training instabilities and competing gradients
documented for physics-informed networks.

\subsubsection{Supervision Data Strategies}
\label{subsubsec:supervision}

Available information fundamentally determines viable methodological approaches.
\emph{Equation-driven regimes} (e.g., PINNs, variational PINNs, residual-based neural
operators), assume known governing PDEs with minimal observation data and enable
physics-informed training where differential equations provide supervision, addressing scenarios
where classical methods are tractable but expensive---common in inverse problems and parameter
estimation.
\emph{Data-driven regimes} (CNNs, U-Nets, Transformers, DeepONets, FNOs, GNOs) assume
abundant simulation or experimental data with uncertain or partially known governing equations,
justifying purely supervised approaches, particularly when classical solvers have already
generated extensive datasets that can be repurposed for surrogate or operator learning.
\emph{Hybrid models} combining known physics with sparse observations enable multi-fidelity
approaches that leverage both sources of information to achieve superior data efficiency
compared to purely supervised methods while maintaining physical consistency.

The data regime determines which computational barriers from Section~\ref{sec:evaluationclassical}
can be addressed: equation-driven approaches circumvent expensive data generation but face
optimization challenges; data-driven methods achieve rapid inference for parametric studies but
require amortizing training costs over many queries (Section~\ref{sec:economics}); hybrid
approaches balance these trade-offs but increase implementation complexity.

\subsection{Computational Economics and Deployment Constraints}
\label{sec:economics}

Deployment decisions for ML-PDE methods ultimately hinge on a rigorous cost–benefit analysis performed prior to training. Figure~\ref{fig:computational_economics} maps the principal method families in this economic space using order-of-magnitude estimates from representative published benchmarks. Because reported costs depend strongly on dimensionality, resolution, hardware, and—crucially—the classical baseline, the figure should be read as a qualitative map of \emph{relative} computational character rather than precise absolute values.

Several structural patterns emerge consistently across studies. Physics-informed neural networks (PINNs) occupy a low-training-cost, moderate-speedup regime ideally suited to single-instance inverse problems and parameter estimation. Neural operators (FNO, DeepONet) and graph neural network simulators cluster in a high-training-cost, high-speedup region that becomes economically viable only in many-query regimes (hundreds to thousands of evaluations), such as uncertainty quantification, parametric studies, and real-time control. Foundation models lie at the extreme high-training-cost end and currently demand the largest query volumes for amortization, although their emerging zero-shot transferability across PDE families may eventually shift this balance.

Figure~\ref{fig:computational_economics}B explicitly illustrates the amortization structure: the break-even query count $N_{\rm be}$ (Eq.~\ref{eq:breakeven}) ranges from tens of queries for PINNs to hundreds for standard neural operators and potentially thousands for foundation models. Applications falling left of a method’s crossover point remain more efficiently served by a well-optimized classical solver. This section summarizes order-of-magnitude training compute, inference characteristics, and
amortization considerations.

\begin{figure}[t]
    \centering
    \includegraphics[width=0.95\textwidth]{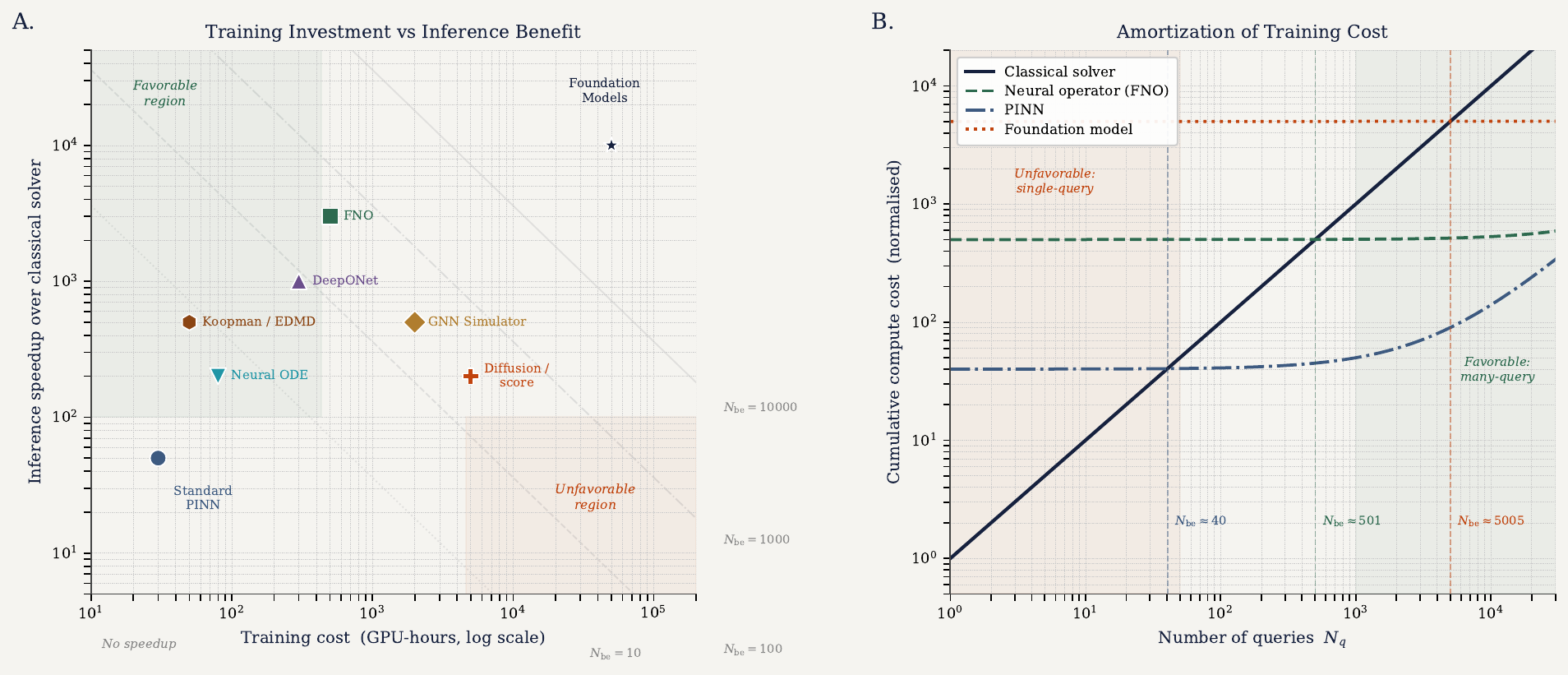}
    \caption{Computational economics of machine learning methods for PDE solution. All cost and speedup estimates are order-of-magnitude syntheses from representative published benchmarks~\cite{li2020fourier,lu2021learning,karniadakis2021physics,pfaff2020learning,song2020score,chen2018neural,herde2024poseidon} and illustrate the \emph{relative} computational character of each method family. Actual values vary significantly with problem dimensionality, resolution, rollout horizon, implementation quality, and classical baseline.
    \textbf{(A)} Log-log scatter of training cost (GPU-hours) versus inference speedup (relative to classical solver) for eight representative ML-PDE families. Diagonal iso-lines show break-even query count $N_{\rm be} = C_{\rm train}/(C_{\rm classical} - C_{\rm inference})$ (Eq.~\ref{eq:breakeven}): methods above the $N_{\rm be}=100$ line recoup training investment after fewer than 100 queries. Shaded quadrants separate the \emph{favorable} regime (high speedup, low training cost) from the \emph{unfavorable} one (low speedup, high training cost).
    \textbf{(B)} Cumulative cost versus query count for a classical solver and three representative neural methods, based on the amortization model of Eq.~\ref{eq:breakeven}. Vertical dashed lines indicate break-even crossovers: the minimum query count at which each neural method becomes cheaper overall than repeated classical solves. Shaded regions delineate unfavorable regimes (few-query problems, left) from favorable ones (many-query scenarios such as uncertainty quantification, parametric studies, and real-time control, right).}
    \label{fig:computational_economics}
    \label{fig:computational_economics}
\end{figure}

\subsubsection{Training Costs}
Table~\ref{tab:training_costs} reports \emph{order-of-magnitude} training compute and data
requirements by method family, based on representative studies.
Values vary substantially with PDE family, resolution, rollout horizon, implementation,
optimizer, and hardware; memory scaling reflects \emph{typical on-device (GPU) memory} during
training (per batch), not offline dataset storage.

\begin{table}[h!]
\centering
\caption{Order-of-magnitude training costs by method family.
Compute and memory are indicative; actual costs vary substantially with problem size, rollout
horizon, and hardware.
GPU memory reflects typical on-device usage per batch during training, not offline dataset
storage.}
\label{tab:training_costs}
\begin{tabular}{p{3.2cm}p{2.3cm}p{3.5cm}p{4.5cm}}
\toprule
\textbf{Method} & \textbf{GPU-Hours} & \textbf{Training Data} &
\textbf{Typical GPU Memory Scaling} \\
\midrule
Standard PINN~\cite{karniadakis2021physics}
  & $10^1$--$10^2$
  & Minimal: PDE + BC/IC; optional $O(10^2)$ sparse observations
  & $O(B \cdot P)$ for activations; $O(N_{\text{pts}})$ for collocation storage \\[3pt]
FNO~\cite{li2020fourier}
  & $10^2$--$10^3$
  & $10^3$--$10^4$ solution fields or rollouts (supervised)
  & $O(B \cdot C \cdot N^d)$ for grid-based operators \\[3pt]
DeepONet~\cite{lu2021learning}
  & $10^2$--$10^3$
  & $10^3$--$10^5$ input--output function pairs
  & $O(B \cdot (N_s + N_q) \cdot C)$ \\[3pt]
GNN simulators~\cite{pfaff2020learning}
  & $10^2$--$10^4$
  & $10^2$--$10^4$ trajectories (mesh or graph)
  & $O(B \cdot (|V|+|E|) \cdot C)$ \\[3pt]
Diffusion / score-based~\cite{song2020score}
  & $10^3$--$10^4$
  & $10^3$--$10^5$ samples (conditional on parameters or BCs)
  & $O(B \cdot N \cdot C)$; inference cost scales with $T_{\text{steps}}$ \\[3pt]
Neural ODEs~\cite{chen2018neural}
  & $10^1$--$10^3$
  & $10^2$--$10^4$ trajectories
  & $O(B \cdot P)$ via adjoint; independent of integration depth \\[3pt]
Koopman / EDMD~\cite{lusch2018deep,li2017extended}
  & $10^1$--$10^3$
  & $10^2$--$10^4$ trajectories
  & $O(B \cdot m \cdot N)$ for encoder; $O(m^2)$ for linear operator $K$ \\[3pt]
Foundation models~\cite{herde2024poseidon,sun2025towards}
  & ${\geq}10^4$ (model-dependent)
  & ${\geq}10^4$ solutions spanning multiple PDE families and regimes
  & Model-dependent; often dominated by parameter and optimizer states \\
\bottomrule
\end{tabular}
\vspace{0.5em}

\footnotesize{\textit{Notation:}
$B$~=~batch size;
$P$~=~number of network parameters;
$N$~=~spatial degrees of freedom per sample;
$d$~=~spatial dimension;
$C$~=~number of feature channels;
$N_s$~=~number of input sensors (DeepONet branch);
$N_q$~=~number of query points (DeepONet trunk);
$|V|$,~$|E|$~=~graph vertices and edges;
$T_{\text{steps}}$~=~number of diffusion denoising steps;
$m$~=~Koopman embedding dimension.
Data generation cost can dominate total training time for supervised neural operators
(Section~\ref{sec:economics}); the amortization argument requires a stable, well-characterized
deployment distribution.}
\end{table}

For supervised neural operators, \emph{data generation can dominate total cost}: benchmark
suites such as PDEBench are built on large collections of numerical simulations, and generating
high-resolution rollouts can be a major fraction of end-to-end
time-to-solution~\cite{takamoto2022pdebench}.

\subsubsection{Inference Speedups}
Neural methods achieve 2--4 orders of magnitude inference speedup over classical solvers once
trained~\cite{li2020fourier}.
FNO evaluates solutions in milliseconds versus minutes or hours for classical methods on
equivalent problems.
However, a 2024 study found systematic weaknesses in comparative
evaluation~\cite{mcgreivy2024weak}: 79\% (60 of 76) of ML-for-PDE papers claiming superiority
used weak baselines, comparing optimized neural networks against unoptimized solver
implementations.
Claims of ``three orders of magnitude faster'' often compared 12-millisecond neural inference
against 12-second unoptimized FEniCS code, when optimized direct solvers complete in
milliseconds.
This finding demands rigorous re-evaluation of published speedup claims.
This does not negate the value of neural surrogates, but underscores that speedups must be
reported relative to optimized, problem-appropriate numerical baselines.

\subsubsection{Amortization Analysis}
The computational break-even point depends on training cost $C_{\text{train}}$, per-query
classical solver cost $C_{\text{classical}}$, and per-query neural inference cost
$C_{\text{inference}}$:
\begin{equation}
N_{\text{break-even}} = \frac{C_{\text{train}}}{C_{\text{classical}} - C_{\text{inference}}}
\label{eq:breakeven}
\end{equation}

For typical neural operators, $N_{\text{break-even}}$ ranges from hundreds to thousands of
queries.
This fundamentally limits applicability to many-query scenarios:

\begin{itemize}
    \item \textbf{Favorable:} Parametric studies, optimization loops, uncertainty quantification
    with Monte Carlo sampling, real-time control requiring repeated evaluation.
    \item \textbf{Unfavorable:} Single-query problems, high-accuracy requirements ($<10^{-4}$),
    problems outside training distribution, safety-critical applications requiring certified
    error bounds.
\end{itemize}

Recent benchmarking efforts~\cite{takamoto2022pdebench,zhou2024unisolver,koehler2024apebench}
provide standardized comparisons across methods, though fair baseline selection remains
challenging.

\section{Capabilities and Limitations of Machine Learning Methods}

\label{subsec:ml_challenges}

Despite rapid algorithmic advances, ML approaches for solving PDEs remain confronted with persistent challenges, namely, approximation barriers, reliability gaps that impede safety-critical deployment, data and computational constraints that limit practical feasibility, and theoretical deficiencies that obstruct rigorous verification. Understanding which limitations reflect fundamental mathematical constraints versus addressable implementation issues are essential for a realistic assessment of ML-PDE capabilities and for selecting the appropriate approach for a given problem.

\begin{figure}[t]
    \centering
    \includegraphics[width=1\linewidth]{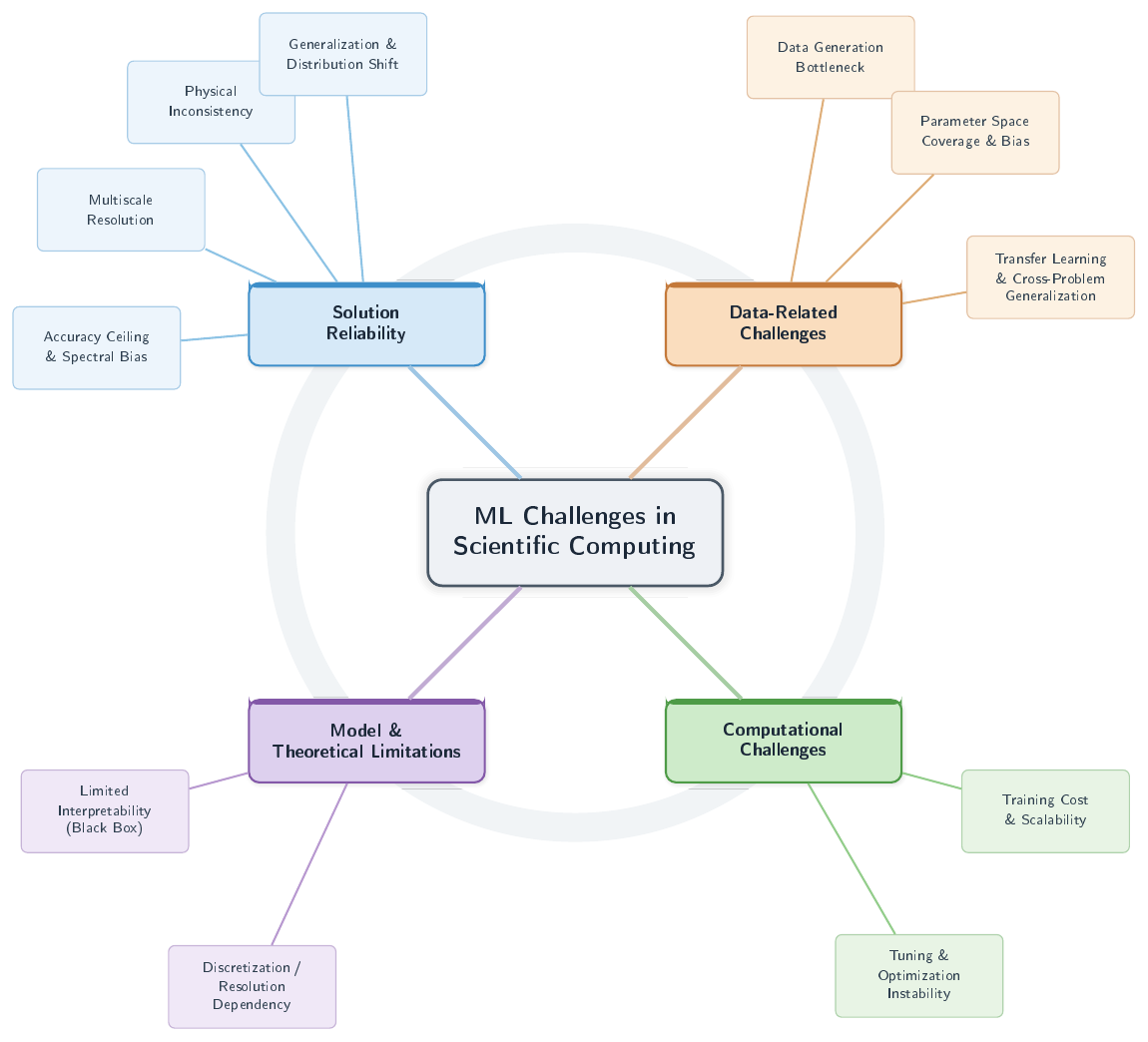}
    \caption{Hierarchical summary of major challenges in applying machine learning to scientific computing, organized into four major categories: solution reliability, data-related challenges, computational challenges, and model and theoretical limitations.}
    \label{fig:ml_challenges}
\end{figure}

Figure~\ref{fig:ml_challenges} organizes these limitations hierarchically across four
categories: \emph{solution reliability}, \emph{data-related challenges},
\emph{computational challenges}, and \emph{model and theoretical limitations}.
The following subsections address each category in turn, connecting observed failure modes back to the specific methods introduced in Sections~\ref{sec:surrogate
models}--\ref{sec:hybrid_ml}. Extended technical discussion and method-specific analyses appear in Supplementary Information (Section~\ref{sec:supp}).

\subsection{Solution Reliability and Accuracy Constraints}
\label{subsubsec:reliability}

Solution reliability encompasses the set of failure modes that arise when a trained ML model is queried in deployment, independently of whether training converged.
These include accuracy limitations intrinsic to neural representations, poor behavior under domain shift, and the production of physically inadmissible solutions.

\paragraph{Accuracy Ceiling.} Neural methods achieve accuracy levels on standard benchmarks that often, but not always, fall short of high-order classical methods on smooth problems, as documented in
Table~\ref{tab:method_comparison} and detailed in Table~\ref{tab:classical-pde-methods}.
This persistent gap across architectures---including recent variants with sophisticated
designs---reflects fundamental approximation characteristics of neural representations rather than implementation limitations.

The spectral and finite element methods discussed in  Section~\ref{sec:classicmethods} benefit from well-understood convergence hierarchies: increasing polynomial order or refining the mesh
produces predictable accuracy improvements with quantifiable rates.
No analogous refinement hierarchy exists for neural networks; increasing network width or depth does not guarantee monotone accuracy improvement, and the relationship between architecture complexity and solution quality remains problem-dependent. Although universal approximation theorems and empirical scaling law studies provide asymptotic
or statistical trends, these do not constitute the deterministic, a priori error bounds that
characterize classical refinement hierarchies.
For applications requiring rigorous error bounds or certified precision, classical methods
therefore remain essential.

\paragraph{Spectral Bias and High-Frequency Features.}
Neural networks preferentially learn low-frequency solution components due to initialization properties and gradient flow dynamics in parameter space~\cite{rahaman2019spectral}. This spectral bias is not an incidental artifact of particular architectures but a structural
property of gradient-based training on smooth loss landscapes, affecting MLPs, CNNs,
FNOs, and PINNs alike, albeit to different degrees.

This phenomenon bears a conceptual resemblance to the Gibbs phenomenon identified for
spectral methods in Section~\ref{subsec:eval_discontinuities}: both degrade accuracy in precisely the same problem class---sharp fronts, boundary layers, and high-frequency oscillations. While the mechanisms differ---the Gibbs phenomenon is an approximation-theoretic limitation of truncated global expansions, arising from representing discontinuities in a finite Fourier basis, whereas spectral bias reflects convergence rate disparities in gradient-based
optimization, mediated by the eigenspectrum of the neural tangent
kernel---both manifest as degraded accuracy for the same class of solution features.
This shared vulnerability underscores that the tension between global smooth representation and local solution features is mathematical rather than architectural: any global smooth representation, whether analytically prescribed or implicitly learned, faces inherent difficulties with localized structure.

Slow convergence or complete failure occurs for solutions with sharp gradients, boundary layers, or high-frequency oscillations---precisely the features characterizing the discontinuous solutions discussed in Section~\ref{subsec:discontinuities} and the multiscale
problems of Section~\ref{subsec:multiscale}.
Mitigation strategies, including Fourier feature mappings, modified activation functions, and
wavelet-based architectures, partially address this but require problem-specific tuning with
no universal solution.

\paragraph{Multiscale Phenomena.}
Standard architectures struggle when characteristic scales differ by orders of magnitude, directly linking to the multiscale challenge of Section~\ref{subsec:multiscale}. The global Fourier basis of FNO (Section~\ref{sec:neural_operators}) inadequately captures localized fine-scale features; standard PINNs (Section~\ref{sec:pinns}) require impractical
collocation density to simultaneously resolve disparate scales.
Wavelet-based operators and hierarchical GNNs partially address this by providing explicit multi-resolution structure~\cite{tripura2023wavelet,fortunato2022multiscale}, but these
architectural modifications embed classical multiscale concepts within neural frameworks---suggesting that genuine multiscale capability requires problem-specific inductive biases that mirror classical approaches rather than emerging solely from data.

\paragraph{Physical Inconsistency.}
A failure mode that is well-controlled in classical solvers but poorly characterized in purely data-driven methods is the production of physically inadmissible solutions: negative densities or pressures from fluid surrogate models, non-divergence-free velocity fields from incompressible flow surrogates, violated conservation balances from autoregressive rollouts, or thermodynamically inconsistent stress-strain responses from learned constitutive models.
Classical numerical methods can also produce physically inadmissible outputs under specific conditions---for example, negative densities from high-order schemes without limiters, spurious oscillations from unstabilized Galerkin FEM in convection-dominated regimes, or Gibbs-induced overshoots in under-resolved spectral methods---but these failure conditions are well characterized and systematically remediable through established techniques such as flux limiters, stabilization operators, and mesh refinement. 

In contrast, physically inconsistent outputs from neural models may be invisible to standard loss metrics during training, yet problematic in downstream use, with no systematic framework for predicting or preventing their occurrence.
Pure data-driven surrogates (Section~\ref{subsubsec:generic_deep}) and generic transformers (Section~\ref{subsubsec:transformers}) are most susceptible, as they impose no physical structure. Hard architectural constraints (divergence-free networks, symplectic integrators, energy-based constitutive models) as discussed in Sections~\ref{subsubsec:physics_integration} and~\ref{sec:hybrid_ml} can eliminate specific inconsistencies by design, but require knowing in advance which physical properties must be enforced. Soft constraints via loss penalization, as in PINNs (Section~\ref{sec:pinns}), reduce but do not eliminate the risk of violations when loss weighting is improperly balanced.

\paragraph{Generalization and Distribution Shift.}
Out-of-distribution (OoD) performance degrades rapidly for parameters, geometries, or
initial conditions outside the training distribution, which is a fundamental departure from classical methods that provide convergence guarantees. The convergence guarantees while dependent on problem regularity
and discretization parameters, are a priori computable from the problem description alone. Among the methods reviewed, neural operators (FNO, DeepONet, GNO;
Section~\ref{sec:neural_operators}) show improved resolution generalization compared to pointwise surrogates but share the same sensitivity to parameter-space OoD queries. PINNs (Section~\ref{sec:pinns}) are also trained per-problem and do not transfer across parameter regimes without retraining. Foundation models (Section~\ref{subsubsec:transformers}) represent the most ambitious attempt
to address this through broad pretraining, but principled frameworks for predicting generalization quality to unseen PDE families remain unavailable. This limitation particularly constrains inverse problems and parameter estimation (Section~\ref{subsec:eval_nonlinearity}), where exploration beyond observed
regimes is essential by definition.

\begin{figure}[h!]
    \centering
    \includegraphics[width=0.95\textwidth]{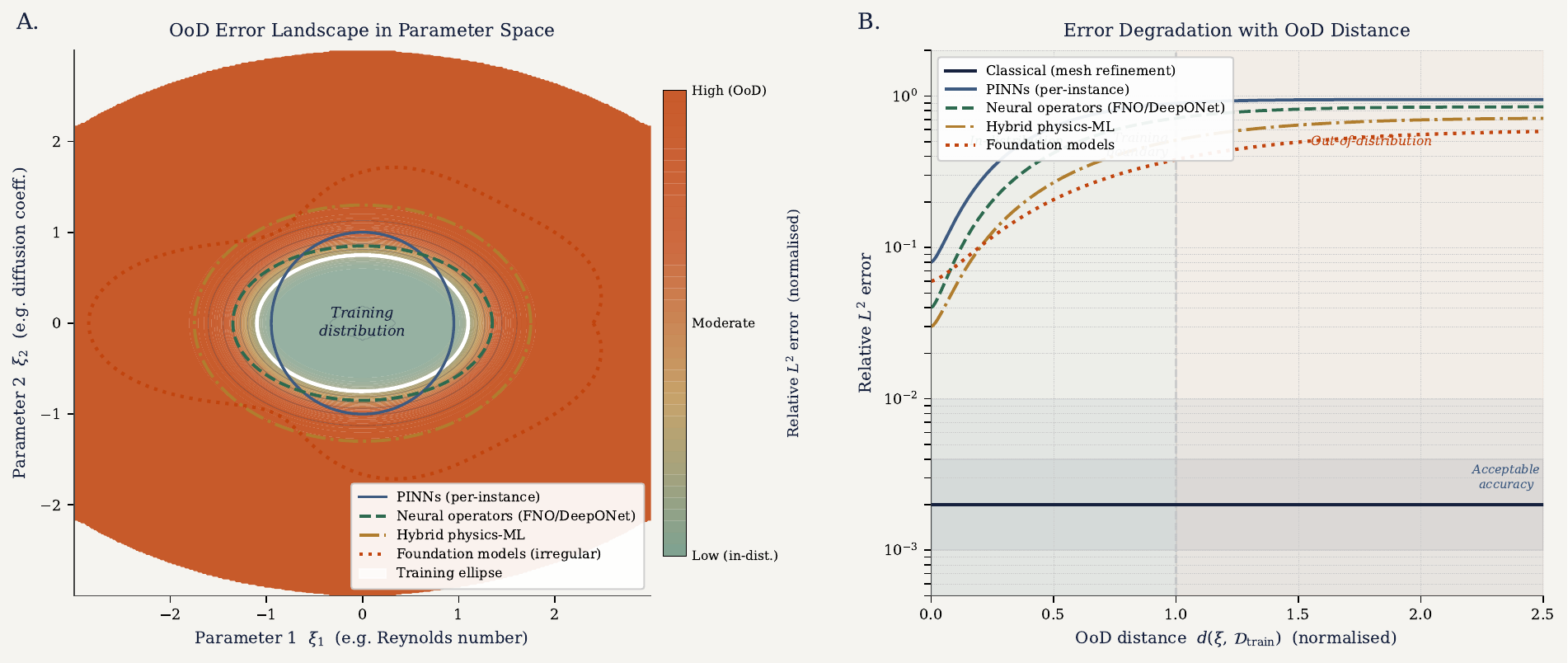}
    \caption{Out-of-distribution (OoD) generalisation and degradation for representative ML-PDE method families. All contours and curves reflect qualitative trends synthesised from published benchmarks and illustrate characteristic degradation behaviour rather than precise, problem-specific performance. Actual rates depend on the PDE type, the parameter-space geometry, the training-set coverage, and the architectural choices.
    \textbf{(A)} Two-dimensional slice of parameter space (e.g., Reynolds number vs.\ diffusion coefficient). The white ellipse bounds the training distribution $\mathcal{D}_{\mathrm{train}}$; the background colour field shows normalised relative $L^2$ error (green: near-zero inside training support; crimson: high far outside). Iso-accuracy contours indicate extrapolation range before unacceptable degradation:
    - PINNs (steel, solid): minimal extension beyond training boundary (per-instance training).
    - Neural operators FNO/DeepONet (emerald, dashed): moderate extension via resolution-invariant operator learning.
    - Hybrid physics-ML methods (gold, dash-dot): substantially larger reach, driven by PDE operator as inductive bias.
    - Foundation models (crimson, dotted): irregular/asymmetric boundary due to uneven cross-PDE-family transfer.
    Classical solvers (omitted) show no distribution-shift dependence; error is controlled solely by discretisation parameters.
    \textbf{(B)} Normalised $L^2$ error versus OoD distance $d(\xi,\mathcal{D}_{\mathrm{train}})$ for the four ML families plus classical reference (dark navy, flat). Vertical dashed line: training boundary. Shaded regions separate in-distribution from OoD regimes. Horizontal band: representative acceptable-accuracy threshold. PINNs degrade most steeply; hybrids degrade most slowly (physics structure persists); foundation models show intermediate rates with slower initial fall-off (benefit of broad pretraining) but higher in-distribution error. No ML family achieves the distribution-independent behaviour of classical solvers.}
\label{fig:ood_generalisation}
\end{figure}

Figure~\ref{fig:ood_generalisation} provides a geometric interpretation of this out-of-distribution (OoD) degradation. Figure~\ref{fig:ood_generalisation}A maps a representative two-dimensional parameter subspace around the training distribution $\mathcal{D}_{\mathrm{train}}$ (white ellipse): the normalised $L^2$ error field rises sharply outside the training support, while iso-accuracy contours reveal a clear hierarchy among ML-PDE families. PINNs (per-instance training) barely extend beyond $\mathcal{D}_{\mathrm{train}}$; neural operators (FNO, DeepONet) reach moderately farther thanks to resolution-invariant operator learning; hybrid physics-ML methods extend significantly further, with the retained PDE operator serving as a strong inductive bias that constrains extrapolation; foundation models display irregular, asymmetric boundaries reflecting uneven cross-PDE-family transfer success. Classical solvers are absent: their error depends solely on mesh resolution and exhibits no dependence on proximity to training data. Figure~\ref{fig:ood_generalisation}B further schematically quantifies degradation as error versus normalised OoD distance $d(\xi,\mathcal{D}_{\mathrm{train}})$. All ML families degrade monotonically beyond the training boundary (vertical dashed line), at markedly different rates, while the classical reference remains flat (error governed by discretisation only). The practical consequence is stark: every trained ML-PDE model has a finite, problem-specific effective operating radius in parameter space—within which accuracy can be trusted without further validation—yet this radius is currently neither predictable a priori nor comparable to the rigorous convergence guarantees of classical mesh refinement.

\subsection{Data Requirements and Generalization Limits}
\label{subsubsec:data_challenges}

Data requirements constitute a fundamental cost axis that distinguishes ML-PDE methods from classical solvers and from one another. Unlike classical methods, which solve a given problem from governing equations, data-driven surrogates and neural operators must be trained on datasets generated by the very numerical solvers they are intended to replace or accelerate, creating a circular
dependency that constrains practical feasibility.

\paragraph{The Data Generation Bottleneck.}
For supervised neural operators (FNO, DeepONet, GNO; Section~\ref{sec:neural_operators}) and generic deep surrogates (Section~\ref{subsubsec:generic_deep}), training datasets typically consist of large collections of input-output function pairs obtained by running classical
numerical solvers across a parameter space. As documented in Table~\ref{tab:training_costs}, this data generation cost can dominate
total end-to-end time-to-solution, particularly for high-resolution or long-horizon
problems where each classical solution is itself expensive. The amortization argument, that training cost is recovered across many subsequent
queries, only holds when the deployment distribution is stable and well characterized in advance; for exploratory or inverse problems, this assumption frequently fails.

\paragraph{Coverage and Sampling of the Parameter Space.}
Beyond the raw volume of data required, the \emph{distribution} of training samples
fundamentally determines deployment reliability. Data-driven methods inherit whatever biases are present in the training dataset: undersampled parameter regimes, atypical boundary configurations, or geometrically homogeneous training domains all create blind spots that are invisible during training but consequential during
deployment. Active learning and adaptive sampling strategies offer partial remedies by directing data generation toward under-represented or high-uncertainty regions, but these are less often employed in practice due to the complexity of closed-loop training pipelines. Multi-fidelity approaches, as discussed in Section~\ref{sec:hybrid_ml}, partially mitigate the data cost by combining inexpensive low-fidelity simulations with sparse high-fidelity
labels, but require careful design of the fidelity hierarchy.

\paragraph{Transfer Learning and Cross-Problem Generalization.}
Each new PDE problem---different geometry, boundary conditions, coefficient fields, or PDE type---generally requires a new training dataset and retraining cycle.
Unlike classical solvers, which solve any well-posed problem from its description alone, purely data-driven models carry no transferable knowledge of governing equations from one problem to the next. PINNs (Section~\ref{sec:pinns}) are partially exempt from this constraint, as they encode the governing equation directly into the loss and require only collocation point samples rather than pre-computed solution fields; however, they also require retraining per problem
instance. Neural ODEs and Koopman methods (Section~\ref{sec:hybrid_ml}) offer the prospect of transferable dynamical structure across similar systems, but demonstrated cross-problem generalization remains limited. Foundation models (Section~\ref{subsubsec:transformers}) represent the most direct attempt
to build cross-problem transferability, with early positive results, though the conditions under which zero-shot transfer is reliable are not yet established.

\subsection{Computational and Scalability Challenges}
\label{subsubsec:computational_challenges}

ML-PDE methods introduce a distinct computational cost structure relative to classical
solvers, with implications for both the feasibility of training and the practicality of
deployment.

\paragraph{Training Cost and Amortization.}
As formalized in the amortization analysis of Section~\ref{sec:economics}, the
computational investment in training a neural surrogate is only recovered if the model is
queried sufficiently many times.
This fundamentally restricts ML surrogates to many-query scenarios, such as parametric
sweeps, optimization loops, and uncertainty quantification by Monte Carlo sampling.
For single-query or low-query-count problems, which are common in engineering design,
where a single high-fidelity simulation is the objective, the training overhead renders
ML approaches computationally inferior to well-optimized classical solvers, particularly
given the weak baseline issue identified in Section~\ref{sec:economics}.
Among the methods reviewed, PINNs (Section~\ref{sec:pinns}) have the lowest data cost
but comparable or higher training cost per problem instance; neural operators
(Section~\ref{sec:neural_operators}) have high upfront cost but low per-query cost once
trained; hybrid methods (Section~\ref{sec:hybrid_ml}) vary depending on the extent to
which the learned component replaces classical computation.

\paragraph{Memory and Scaling to High-Dimensional Problems.}
As indicated in Table~\ref{tab:training_costs}, GPU memory requirements scale with problem
resolution and batch size in ways that limit direct application to high-resolution
three-dimensional problems.
Transformers (Section~\ref{subsubsec:transformers}) face quadratic memory scaling with
token count, restricting resolution; GNN simulators (Section~\ref{subsubsec:gnns}) scale
with graph size and require storing all edge features; score-based diffusion models
require storing intermediate denoising states.
While model parallelism and reduced-precision training partially alleviate these
constraints, the memory footprint of training large neural surrogates on fine
three-dimensional meshes remains a practical barrier that classical solvers, leveraging
decades of memory-efficient algorithm design, do not face to the same degree.

\paragraph{Hyperparameter Burden and Reproducibility.}
Classical numerical methods are governed by a small set of well-understood parameters---mesh
spacing, time step, polynomial order---with established guidelines for their selection
derived from stability analysis, convergence theory, and decades of empirical experience.
ML-PDE methods introduce a qualitatively different challenge: hyperparameter optimization
is itself a black-box optimization problem over a high-dimensional space where each
function evaluation requires training a complete neural network to
convergence~\cite{Franceschi_2025}.
The hyperparameter vector typically includes architectural choices (depth, width,
activation functions), optimization settings (learning rate schedules, batch size,
optimizer selection), physics-embedding parameters (loss weights for PINNs, collocation
point distributions), and regularization strategies (dropout rates, weight decay,
data augmentation).
For a moderately complex PINN or neural operator, the effective hyperparameter dimension
easily exceeds twenty, and the objective function---validation set performance---is
non-convex, noisy due to stochastic initialization, and expensive to evaluate.

\begin{figure}[t]
\centering
\includegraphics[width=0.48\textwidth]{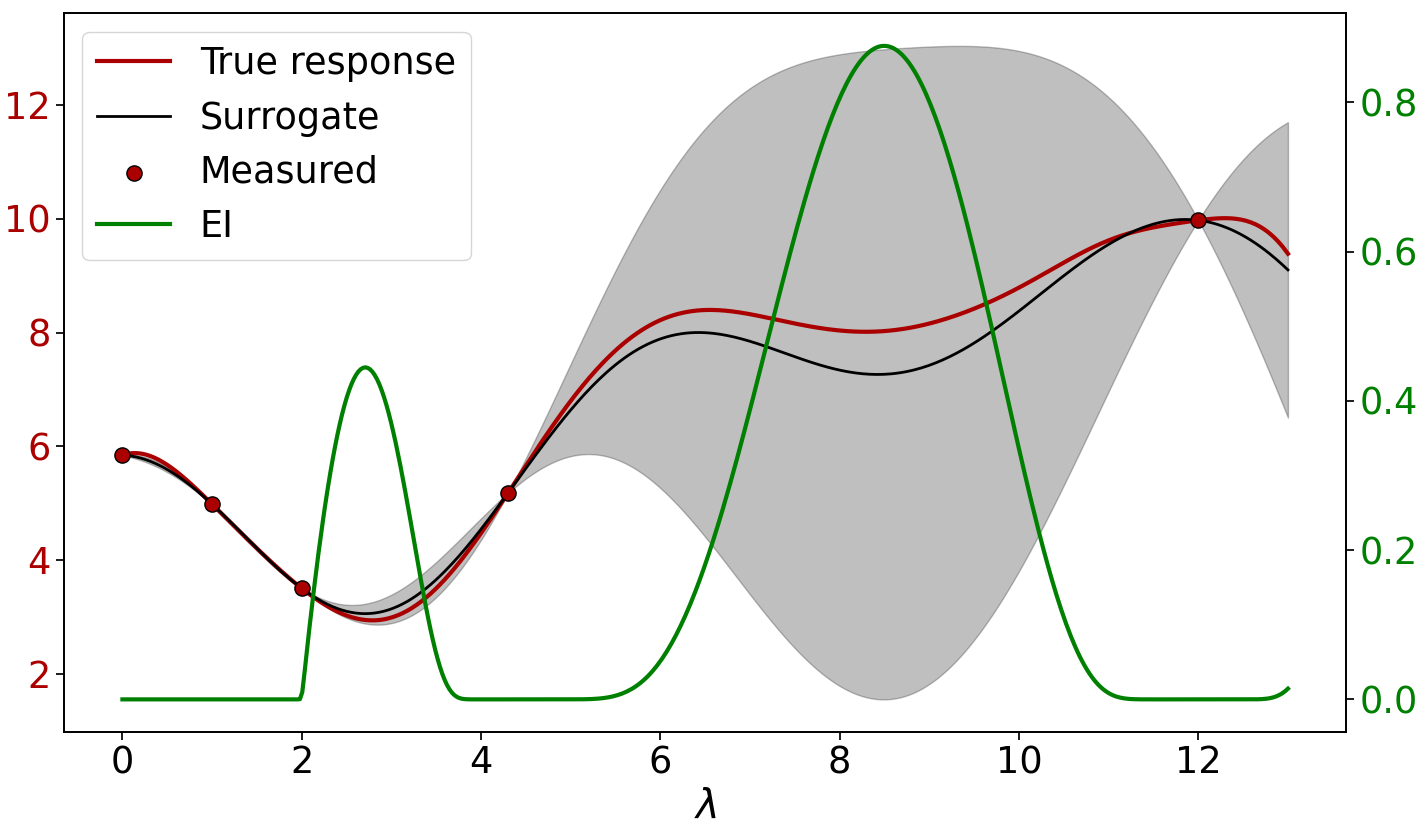}
\includegraphics[width=0.48\textwidth]{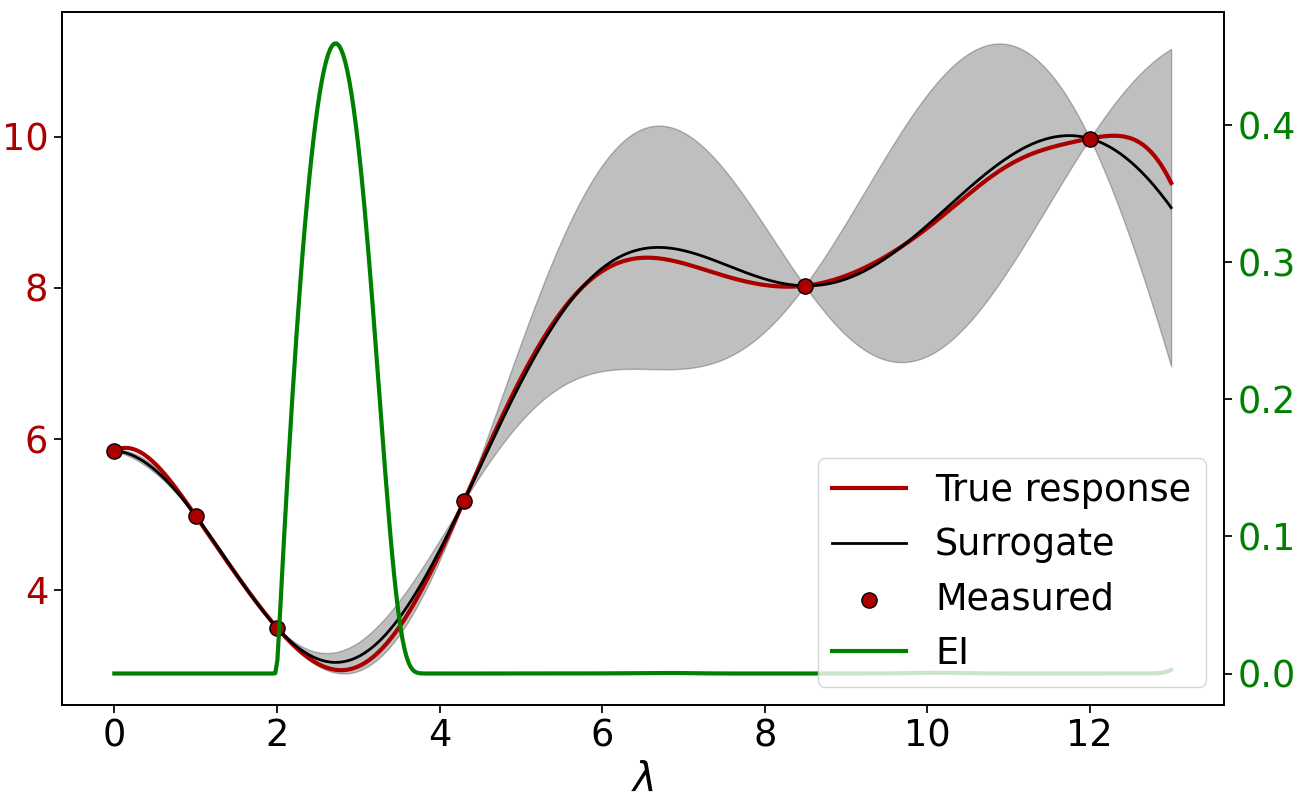}
\caption{Bayesian optimization for hyperparameter selection.
\textbf{Left:} After evaluating validation performance (red curve, unknown to the
optimizer) at a sparse set of hyperparameter configurations, a Gaussian process surrogate
(black line, with $\pm1\sigma$ uncertainty in gray) models the objective landscape.
The expected improvement acquisition function (green) identifies the next configuration
to evaluate by balancing exploitation of the current best region (left peak) against
exploration of high-uncertainty regions (right peak).
In this iteration, exploration dominates and the next candidate is selected near
$\lambda = 8.5$.
\textbf{Right:} After observing the objective at the new configuration, the surrogate
model is updated with substantially reduced uncertainty in the explored region.
The acquisition function now concentrates around the promising region near $\lambda = 2.7$,
demonstrating convergence toward the true optimum.
For neural PDE solvers, $\lambda$ represents a high-dimensional hyperparameter vector and
each evaluation requires training a complete model to convergence, making systematic
optimization computationally expensive.
Adapted from~\cite{Franceschi_2025}.}
\label{fig:bayesian_optimization}
\end{figure}

Bayesian optimization (BO) has emerged as the state-of-the-art approach for
navigating this space~\cite{Franceschi_2025}, superseding earlier strategies
such as grid search, random search, and evolutionary algorithms, which either
require prohibitively many evaluations or lack the sample efficiency necessary
when each evaluation costs hours of GPU time. BO constructs a probabilistic surrogate model (typically a Gaussian process) of the validation performance surface and uses an acquisition function to balance exploration of uncertain regions against exploitation of known high-performing configurations. A detailed analysis is performed in ~\cite{Franceschi_2025}. Figure~\ref{fig:bayesian_optimization} illustrates this process: after observing the
objective at a small set of hyperparameter configurations, the surrogate model's
uncertainty (shaded region) is highest where data are sparse, and the acquisition function
(expected improvement, shown in green) identifies the next configuration to evaluate by
trading off predicted performance against epistemic uncertainty.
This iterative procedure converges to high-performing configurations more efficiently than
grid search or random search, but each iteration still requires a full model training
cycle, and convergence is not guaranteed for the high-dimensional, non-smooth objective
landscapes characteristic of neural PDE solvers.

The practical consequence is that published performance figures for ML-PDE methods often
reflect extensive, undisclosed hyperparameter search whose computational cost is not
reported alongside training and inference costs.
This hidden cost makes fair comparison across methods difficult and introduces a
reproducibility challenge distinct from that faced by classical methods: run-to-run
variability from stochastic initialization and optimization, combined with the
combinatorial explosion of the hyperparameter space, prevents the systematic convergence
studies---varying one parameter at a time while holding others fixed---that are standard
practice in classical numerical analysis.
The lack of principled default configurations or problem-class-specific initialization
guidelines means that practitioners must either invest substantial computational resources
in hyperparameter optimization or accept potentially suboptimal performance.
Until community-wide BO protocols and benchmark-derived default configurations emerge for
major method families (PINNs, FNO, DeepONet), the hyperparameter burden will remain a
barrier to the routine deployment of ML-PDE methods in production scientific computing
workflows.

\paragraph{Validation and Certification Cost.}
Classical numerical solvers carry established verification and validation (V\&V)
frameworks: convergence tests, manufactured solutions, a posteriori error estimators, and
comparison to analytical benchmarks provide structured evidence of solution quality.
ML-PDE models require an analogous but substantially more expensive validation regime:
since model quality cannot be certified from training loss alone, an independent test set
spanning the intended deployment distribution must be constructed and evaluated.
For safety-critical applications, this validation burden may equal or exceed the cost of
running the classical solver that the ML model is intended to replace, substantially
eroding the amortization advantage.
No standardized V\&V protocol for ML-PDE models currently exists, though emerging
benchmark platforms such as PDEBench~\cite{takamoto2022pdebench} and
APEBench~\cite{koehler2024apebench} represent steps toward systematic evaluation.

\subsection{Theoretical Foundations and Verification Gaps}
\label{subsubsec:theoretical_limits}

Beyond practical challenges of data and computation, ML-PDE methods face deeper limitations rooted in the current state of approximation theory, optimization theory, and interpretability. These limitations are particularly consequential for scientific computing, where solution quality must be understood and certified rather than merely assessed empirically.

\paragraph{Absent Theoretical Foundations.}
Rigorous a priori error bounds that simultaneously account for network architecture, training dynamics, and PDE-specific properties remain largely unavailable.
This gap has three largely independent components. \emph{Approximation theory} asks: can neural networks of a given architecture represent the target PDE solution to a desired accuracy? Universal approximation theorems provide existence results, but neither convergence rates nor constructive guidance for architecture design.
\emph{Optimization theory} asks: does training find a good approximation?
The non-convex loss landscapes of deep networks permit multiple local minima and saddle points; convergence to a globally good solution is not guaranteed, and the relationship between gradient-based training and the quality of the found solution is poorly understood for PDE problems. \emph{Generalization theory} asks: does low training error imply low test error? Classical statistical learning theory provides bounds via VC dimension or Rademacher complexity, but these bounds are typically vacuous for the large networks used in practice. Taken together, these three gaps mean that a trained ML-PDE model carries no certified relationship between its empirical training performance and its reliability on a new problem instance, which is a stark contrast to the convergence hierarchies of classical numerical methods
reviewed in Section~\ref{sec:evaluationclassical}.

\paragraph{Uncertainty Quantification Deficit.}
Only generative models (diffusion, VAE, flows, BNNs) and Gaussian process-based PDE
solvers~\cite{he2025survey,raissi2017machine,bohm2019uncertainty,gawlikowski2023survey} provide native uncertainty quantification through probabilistic formulations. Gaussian process methods for PDEs occupy a structurally unique position: they encode PDE constraints directly as properties of the covariance function, enabling closed-form posterior inference over solution fields from sparse observations with analytically tractable uncertainty estimates~\cite{raissi2017machine,swiler2020survey}. While their cubic scaling with observation count limits applicability in high dimensions, they
represent the most theoretically principled probabilistic approach and serve as a useful reference for calibration. Other methods in the reviewed family---FNO, DeepONet, GNNs, transformers---provide no native
UQ and require post-hoc modifications such as ensemble averaging or Monte Carlo dropout,
which lack theoretical calibration guarantees and do not decompose uncertainty into epistemic
and aleatoric components.
This deficit represents a fundamental barrier to safety-critical applications that demand
quantified confidence intervals, as emphasized in Section~\ref{subsec:eval_final}, and
stands in stark contrast to classical methods that provide rigorous a posteriori error
estimates and certified bounds on quantities of interest.

\paragraph{Interpretability and Scientific Transparency.}
Classical numerical methods operate through mathematically explicit discretization schemes
whose error sources---truncation error, round-off, geometric approximation---are individually
identifiable and addressable.
ML-PDE models, by contrast, encode physical relationships implicitly in distributed network
weights that resist direct interpretation.
This opacity creates two distinct scientific problems.
First, when a model produces an unexpected result, it is generally not possible to trace the
failure to a specific physical mechanism or model component, making debugging and systematic
improvement substantially harder than for classical methods.
Second, for scientific discovery applications (inferring constitutive laws, identifying
governing equations, learning turbulence closures), the ability to extract interpretable
physical relationships from the trained model is a primary objective; current architectures
provide limited support for this.
Hybrid approaches (Section~\ref{sec:hybrid_ml}), particularly learned constitutive models and
Koopman operator methods, offer partial interpretability through their structured coupling
with physics-based components, but fully interpretable neural PDE models remain an open
research direction.

\paragraph{Inverse Problem Reliability.}
Inverse problems, inferring parameters, initial conditions, or forcing fields from sparse observations, represent one of the most cited application domains for ML-PDE methods, particularly PINNs and differentiable physics. However, the critical evaluation framework applied to forward problems applies here with
additional force.
Ill-posedness is intrinsic to most inverse PDE problems: multiple parameter configurations
may be consistent with available observations, and the solution landscape contains multiple
plausible minima.
Classical regularization frameworks (Tikhonov, Bayesian inference) provide principled
approaches to this ill-posedness with quantified solution uncertainty.
PINN-based inverse solvers, by incorporating the PDE as a soft constraint in the loss,
implicitly regularize the problem, but the form of this implicit regularization is not
user-controlled and may introduce unintended biases.
Purely data-driven approaches for inverse problems require training data that covers the
inverse operator, which is typically far harder to generate than forward-problem data.
The combination of soft constraint enforcement, non-convex optimization, and limited UQ
means that ML-based inverse problem solutions should currently be treated as physically
motivated initial estimates requiring validation rather than as certified solutions.

\subsection{Overall Assessment: Addressable and Fundamental Limitations}
\label{subsubsec:ml_assessment}

The four challenge categories reviewed above---Reliability, Data, Computational, and Theoretical Limits---establish a structured assessment of where ML-PDE methods currently stand relative to classical approaches and delineate the practical boundaries within which they can be responsibly deployed. Table~\ref{tab:challenge_tractability} summarizes this evaluation across the four categories and fourteen identified failure modes.

\begin{figure}[h!]
    \centering
    \includegraphics[width=0.95\textwidth]{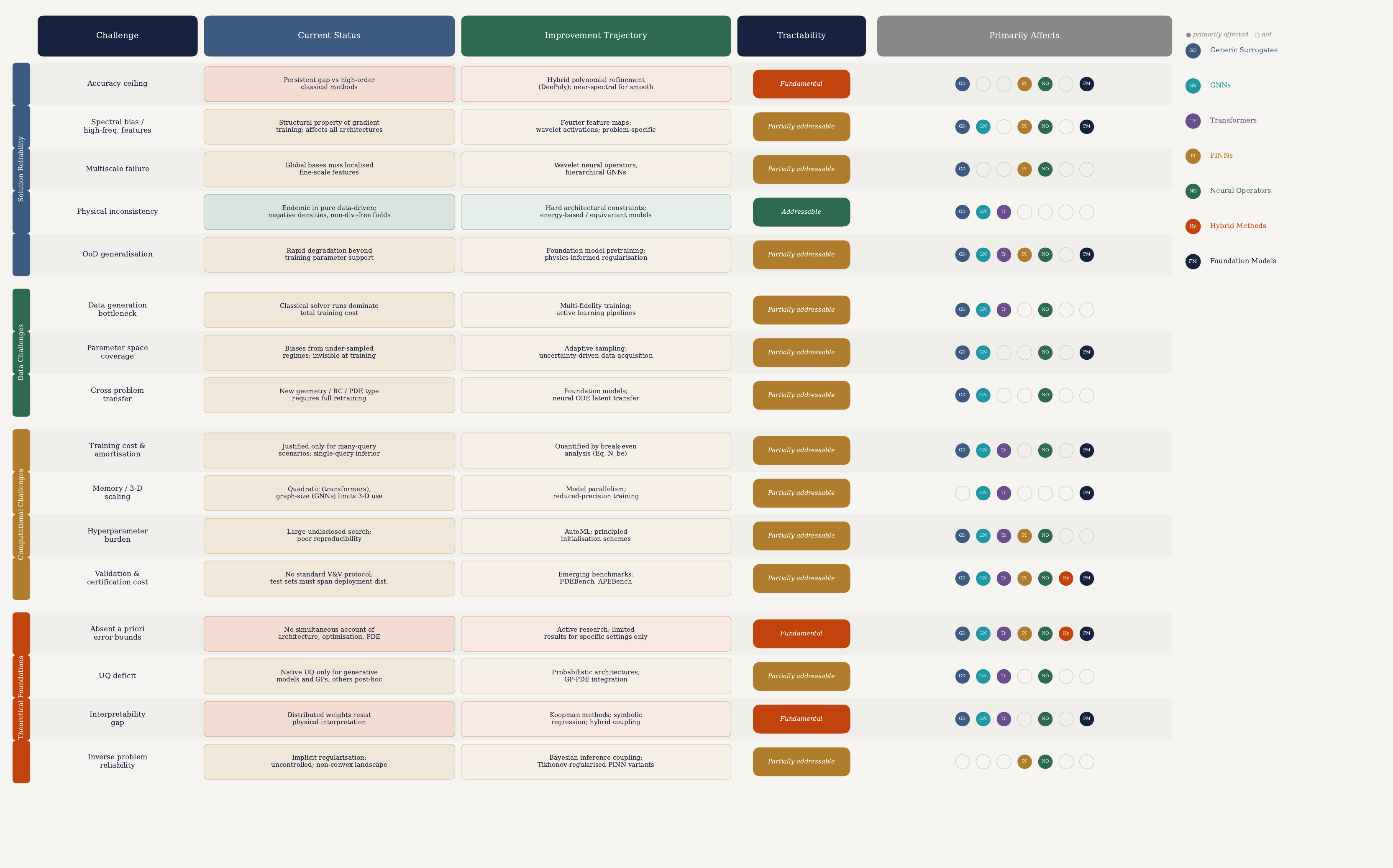}
    \caption{Tractability assessment of persistent challenges facing ML-based PDE solvers, organised by the four categories of Sections~\ref{subsubsec:reliability}--\ref{subsubsec:theoretical_limits}. Each row details one failure mode across three columns: \emph{Current Status} (how the challenge manifests in practice today), \emph{Improvement Trajectory} (most credible near- to medium-term mitigation pathways from the literature), and \emph{Tractability} (\textit{Addressable}, \textit{Partially addressable}, or \textit{Fundamental}), reflecting whether clear technical solutions exist or the limitation is intrinsic to neural approximation and stochastic optimisation. It also indicates tractability level: green (addressable, solutions already demonstrated), gold (partially addressable, requiring further conceptual progress), red (fundamental, unlikely to yield without new mathematical foundations). The rightmost column uses filled coloured markers (following the taxonomy of Figure~\ref{fig:ML_classification}) to show which method families are primarily affected; unfilled circles indicate minor or no impact. No \textit{Fundamental} challenge is confined to a single architecture: the accuracy ceiling and lack of a priori error bounds affect every ML-PDE family reviewed, confirming these as intrinsic properties of the neural paradigm rather than specific design choices.}
    \label{fig:tractability_heatmap}
\end{figure}

Figure~\ref{fig:tractability_heatmap} synthesizes the same analysis into a unified tractability map, revealing three robust structural patterns. First, the two \emph{Fundamental} challenges marked across every method family---the accuracy ceiling and the absence of a priori error bounds---are intrinsic to the neural approximation paradigm itself, not mere architectural flaws; no current architecture evades them, and they are unlikely to be resolved through incremental engineering alone. Second, the cluster of \emph{Partially addressable} challenges concentrated in the Data and Computational categories (data generation cost, parameter coverage, training amortisation, validation overhead) are primarily economic and logistical rather than mathematically intractable; credible mitigation pathways, including multi-fidelity pipelines, active learning, and emerging benchmark standards, are already emerging at small scale. Third, physical inconsistency emerges as the sole \emph{Addressable} challenge of high deployment severity: targeted architectural constraints (divergence-free networks, equivariant operators, energy-based models) eliminate specific inconsistencies by construction, with several practical demonstrations already reported.

The method-impact column further underscores a consistent pattern: pure data-driven surrogates and transformers exhibit the broadest exposure to failure modes, whereas hybrid physics-ML methods evade several entirely by preserving core classical solver structure---a recurring theme across Section~\ref{sec:hybrid_ml} that directly underpins the deployment guidance developed in the following sections.

\textcolor{blue}{%
\begin{table}[h!]
\centering
\caption{Assessment of ML-PDE challenge tractability. Challenges are classified by whether
improvement trajectories are clear (Addressable), partially available (Partially addressable),
or reflect fundamental properties of neural approximation and stochastic optimization
(Fundamental).}
\label{tab:challenge_tractability}
\begin{tabular}{p{3.5cm}p{4cm}p{4cm}p{2.5cm}}
\toprule
\textbf{Challenge} & \textbf{Current Status} & \textbf{Improvement Trajectory} &
\textbf{Tractability} \\
\midrule
Physical inconsistency & Frequent in pure data-driven models & Hard architectural constraints; energy-based models & Addressable \\
Training instability & Significant for PINNs & NTK-based weighting; curriculum learning & Addressable \\
Hyperparameter burden & Pervasive & AutoML; principled initialization & Partially addressable \\
Multiscale handling & Architecture-dependent & Wavelet operators; hierarchical GNNs & Partially addressable \\
Data generation cost & Dominates for supervised methods & Multi-fidelity; active learning & Partially addressable \\
OoD generalization & Rapid degradation & Foundation models; physics-informed training & Partially addressable \\
UQ for non-generative models & Post-hoc only & Probabilistic architectures; GP integration & Partially addressable \\
Accuracy ceiling & Persistent across architectures & Hybrid polynomial refinement (DeePoly) & Fundamental \\
Theoretical foundations & Largely absent & Active research; limited to specific settings & Fundamental \\
Validation/certification & No standard protocol & Emerging benchmarks (PDEBench, APEBench) & Partially addressable \\
\bottomrule
\end{tabular}
\end{table}
}

The addressable challenges---physical inconsistency, training instability, hyperparameter
burden---have clear improvement trajectories through architectural innovation and algorithmic
refinement, and several recent advances (NTK-informed loss balancing, equivariant
architectures, energy-based constitutive models) have already demonstrated meaningful
progress.
The partially addressable challenges---multiscale capability, data generation cost, OoD
generalization---require more substantial conceptual advances, with foundation models and
active learning representing the most promising current directions.
The fundamental challenges---the accuracy ceiling and the absence of theoretical
foundations---appear to reflect inherent properties of neural approximation and stochastic
optimization rather than fixable implementation deficiencies, and are unlikely to be resolved
without new mathematical frameworks that lie beyond the current state of the field.

\paragraph{Mapping the ML Landscape Against the Five Classical Requirements.}
The requirements of Section~\ref{subsec:eval_final} provide a structured summary of
where each ML method family succeeds, falls short, and remains an open question, beyond the per-challenge tractability analysis above.

\emph{Structure preservation.}
Hybrid methods (Section~\ref{sec:hybrid_ml}), particularly energy-based constitutive models
derived from convex potentials and differentiable physics solvers, come closest to satisfying
this requirement by architectural design.
Pure data-driven surrogates (Section~\ref{sec:surrogate_models}) generally fail it:
conservation violations and thermodynamically inconsistent outputs are endemic without
structural constraints, as documented in the physical inconsistency paragraph above.
PINNs (Section~\ref{sec:pinns}) enforce structure softly through loss penalization, which
reduces but cannot eliminate violation risk.
Hard architectural constraints---divergence-free networks, symplectic integrators, equivariant
GNNs---achieve exact satisfaction for specific properties but require advance knowledge of
which structure to enforce and do not transfer automatically to new PDE families.

\emph{Verifiability.}
No current ML-PDE method provides a priori error bounds comparable to the convergence
hierarchies of classical methods.
Gaussian process solvers provide analytically tractable posterior uncertainty but scale
cubically with observation count.
Hybrid DeePoly achieves near-spectral accuracy for smooth problems, but only within the
polynomial correction regime.
The absence of any refinement-like reliability hierarchy---in which tighter approximation is
achievable by systematic resource increase with predictable rates---remains the most
consequential theoretical gap relative to classical methods.

\emph{Problem-aware robustness.}
Classical methods achieve robustness through problem-specific analysis: inf-sup stable pairs,
physics-based preconditioners, parameter-robust coupling schemes derived from understanding
the governing operator structure.
ML methods have no analog mechanism: architectures are fixed before deployment, and their
robustness properties across parameter regimes are unknown until empirically tested.
Foundation models and physics-embedded hybrids represent the most promising directions, but
principled robustness guarantees remain absent.

\emph{Reduced human-in-the-loop burden.}
ML methods offer genuine relief on one dimension: mesh-free formulations (PINNs, DeepONet,
Neural ODEs) eliminate the geometry preprocessing bottleneck identified in
Section~\ref{subsec:eval_geometry} as the dominant workflow cost in classical mesh-based
methods.
However, this reduction is partially offset by a new burden: architecture selection,
hyperparameter tuning, training pipeline design, and validation dataset construction
collectively constitute a different but comparable human investment.
The form of expertise required shifts from mathematical-geometric discretization knowledge
toward ML engineering and statistical validation knowledge---a trade-off whose net benefit
depends strongly on the application community and available expertise.

\emph{Scalability by design.}
Classical solvers leverage decades of multigrid, domain decomposition, and
communication-aware algorithm design to achieve mesh-optimal scaling on HPC architectures.
ML training pipelines are GPU-native and benefit from data parallelism, but model inference
does not exploit multilevel structure: neural operator evaluations at inference are
essentially fixed-resolution feedforward passes without adaptive coarsening.
Hybrid methods that retain the classical solver for global structure and employ ML only for
targeted subproblems---constitutive models, preconditioners, closures---are currently the
most credible path toward HPC-scalable ML-PDE deployment that genuinely satisfies this
requirement.

\begin{table}[h!]
\centering
\caption{ML-PDE method comparison.
Accuracy$^\dagger$ reflects indicative relative $L^2$ errors; values vary substantially
across problem settings.
UQ column indicates native support (Native), approximate support via ensemble or dropout
(Approx.), or absence (No/Limited).}
\label{tab:method_comparison}
\begin{tabular}{p{2.2cm}p{2.5cm}p{1.8cm}p{1.8cm}p{1.8cm}p{3.2cm}}
\toprule
\textbf{Method} & \textbf{Architecture} & \textbf{Complexity} & \textbf{UQ} &
\textbf{Accuracy}$^\dagger$ & \textbf{Best Application} \\
\midrule
\multicolumn{6}{l}{\textit{Physics-Informed}} \\
Standard PINN     & Residual MLP      & $O(N_{\text{pts}} \cdot P)$  & Limited  & $10^{-3}$--$10^{-1}$ & Inverse problems, sparse data \\
VPINN             & Energy functional & $O(N_{\text{quad}} \cdot P)$ & Limited  & $10^{-4}$--$10^{-1}$ & Variational problems \\
XPINN             & Domain decomp.    & $O(K \cdot N \cdot P)$       & No       & $10^{-3}$--$10^{-1}$ & Complex geometries \\
\midrule
\multicolumn{6}{l}{\textit{Neural Operators}} \\
FNO               & Fourier layers    & $O(N \log N)$                & Approx.\ (ensemble) & $10^{-2}$--$10^{-1}$ & Periodic, parametric \\
DeepONet          & Branch-trunk      & $O(N_s \cdot N_q)$           & Approx.\ (dropout)  & $10^{-2}$--$10^{-1}$ & Multiple inputs, flexible \\
GNO               & Graph kernel      & $O(|E|)$                     & Limited  & $10^{-2}$--$10^{-1}$ & Irregular domains \\
\midrule
\multicolumn{6}{l}{\textit{Graph and Attention}} \\
MeshGraphNets     & Message passing   & $O(|V|{+}|E|)$               & Limited  & $10^{-2}$--$10^{-1}$ & Unstructured mesh dynamics \\
Transformer       & Self-attention    & $O(N^2)$                     & Not native & $10^{-2}$--$10^{-1}$ & Long-range interactions \\
\midrule
\multicolumn{6}{l}{\textit{Generative and Probabilistic}} \\
Diffusion         & Score-based       & $O(T \cdot N)$               & Native   & $10^{-2}$--$10^{-1}$ & UQ, stochastic PDEs \\
Gaussian Process  & Kernel-based posterior & $O(n^3)$              & Native   & $10^{-3}$--$10^{-1}$ & Sparse data, calibrated UQ \\
\midrule
\multicolumn{6}{l}{\textit{Hybrid}} \\
Neural-FEM        & FEM + NN          & Varies                       & Limited  & $10^{-4}$--$10^{-2}$ & Conservation-critical \\
PEDS              & LF solver + NN    & $O(N_{\text{LF}} + P)$       & Limited  & $10^{-3}$--$10^{-1}$ & Data-limited scenarios \\
DeePoly           & NN + polynomial   & $O(P + p^d)$                 & No       & $10^{-10}$--$10^{-4}$$^\ddagger$ & High accuracy required \\
Neural ODE/PDE    & Adjoint ODE       & $O(T \cdot P)$               & Limited  & $10^{-3}$--$10^{-1}$ & Irregular-time dynamics \\
Koopman / EDMD    & Encoder + linear $K$ & $O(m^3)$                 & Limited  & $10^{-2}$--$10^{-1}$ & Nonlinear dynamics, control \\
\bottomrule
\multicolumn{6}{l}{\footnotesize $^\dagger$Relative $L^2$ error; varies significantly with PDE family, coefficient regime (ID/OoD), and rollout.}\\
\multicolumn{6}{l}{\footnotesize $^\ddagger$For sufficiently smooth solutions admitting high-order polynomial refinement.}
\end{tabular}
\end{table}

For practitioners, these distinctions translate into concrete deployment guidance.
ML-PDE methods are well-matched to many-query scenarios (parametric studies, optimization,
UQ), problems with high-dimensional parameter spaces where classical methods face the curse
of dimensionality (Section~\ref{subsec:dimensionality}), mesh-free or complex-geometry
settings where GNNs and PINNs reduce discretization burden, and applications where moderate
accuracy is sufficient and the primary cost driver is inference speed.
They are poorly matched to single-query problems where training cost cannot be amortized,
applications requiring certified error bounds below the accuracy floor of current
architectures, safety-critical inverse problems where solution uncertainty must be formally
quantified, and long-horizon rollouts where accumulated unphysical errors compound.
Hybrid approaches (Section~\ref{sec:hybrid_ml}) occupy the most practically viable position
for many applications: by coupling learned components with classical solvers, they preserve
conservation guarantees, gracefully degrade to physics-based baselines when the ML component
fails, and substantially reduce the data requirements imposed on purely data-driven methods.

ML methods, therefore, are most powerful not as replacements for classical numerical methods but as complements, deployed strategically where
their distinct capabilities---high-dimensional approximation, rapid parametric inference, mesh-free geometry handling, and probabilistic solution characterization---address the specific barriers where classical methods face the hardest structural limits. The convergence between classical rigor and neural flexibility---rather than the replacement of one by the other---emerges as the most defensible path forward.

\section{Synthesis of Deductive and Inductive Paradigms}
\label{sec:synthesis}

This section does not revisit the individual methods of Sections~\ref{sec:classicmethods}
and~\ref{sec:mlmethods} nor rehearse the per-challenge evaluations of
Sections~\ref{sec:evaluationclassical} and~\ref{subsec:ml_challenges}.
Instead, it isolates the fundamental distinctions between the paradigms at the level of
mathematical structure, error, and scientific role; identifies conditions under which
they are genuinely complementary; and argues that progress depends not on competition
but on a principled discipline of hybrid design—one whose foundations remain only
partially understood.

\subsection{A Comparative Anatomy: Structure Versus Statistics}
\label{subsec:anatomy}

The deepest distinction between classical numerical methods and machine learning
approaches for PDEs is epistemological rather than computational.
Classical methods are \emph{deductive}: algorithms follow from the mathematical
structure of the governing equations, and errors are bounded by quantities derived
from PDE regularity and discretization parameters.
Machine learning methods are \emph{inductive}: they infer solution behavior from data
and generalize through learned representations whose accuracy depends on statistical
assumptions about the training distribution.
This distinction is not evaluative but operational: it specifies exactly what each
paradigm can certify—and what it fundamentally cannot. Despite this epistemological contrast, the two paradigms have undergone a striking
convergent evolution toward similar computational structures.
This convergence is not superficial: it reflects shared mathematical constraints on
efficient representation of PDE solution operators.

A closely related distinction concerns how prior knowledge is encoded.
Classical methods encode prior knowledge \emph{structurally}: the choice of discretization
space encodes regularity assumptions; the choice of flux formulation encodes physical
conservation; the choice of basis functions encodes expected spectral content.
When this prior is correct, accuracy is guaranteed; when it is wrong---a smooth
approximation applied to a discontinuous solution, for instance---the failure mode is
well understood and quantifiable (Gibbs oscillations, loss of convergence rates).
Machine learning methods encode prior knowledge \emph{probabilistically}: through the
training distribution, the architecture's inductive bias, and the loss function's
regularization.
When this prior is correct, generalization follows empirically; when it is wrong, the
failure mode is a distributional mismatch whose severity is generally not detectable
from the training loss.

This epistemological asymmetry has direct consequences for each of the six challenges
of Section~\ref{sec:challenges}.
Where the challenge is characterizable in terms of mathematical structure---conservation,
inf-sup stability, symplectic geometry, entropy conditions---classical methods exploit
that structure precisely and neural methods can only approximate it through soft
constraints or architectural choices that must be specified in advance.
Where the challenge resists structural characterization---where the relevant prior is
statistical rather than mathematical---ML methods adapt and classical methods cannot:
learning from heterogeneous experimental data, approximating unknown constitutive
relations, amortizing parametric evaluations over distributions that are empirically
described but not analytically characterized.

Table~\ref{tab:anatomy_comparison} makes this comparative anatomy explicit across
five axes that are more discriminating than the standard accuracy-versus-speed framing.

\begin{table}[h]
\centering
\caption{Comparative anatomy of classical numerical methods and machine learning approaches
for PDE solution across five discriminating axes. Entries characterize paradigm-level
behavior, not every individual method within each paradigm.}
\label{tab:anatomy_comparison}
\begin{tabular}{p{3.0cm}p{5.8cm}p{5.8cm}}
\toprule
\textbf{Axis} & \textbf{Classical Numerical Methods} & \textbf{Machine Learning Methods} \\
\midrule
Knowledge encoding
  & Structural: conservation laws, variational principles, and inf-sup conditions
    encoded by construction into discretization and solver design
  & Statistical: prior knowledge enters via training distribution, architecture
    inductive bias, and loss function regularization \\[4pt]
Error characterization
  & Deductive: a priori and a posteriori bounds derived from PDE regularity and
    discretization parameters; convergence hierarchy is systematic and refinable
  & Empirical: accuracy assessed on held-out data; no refinement hierarchy;
    training loss does not certify deployment accuracy \\[4pt]
Generalization guarantee
  & Convergence for any well-posed input satisfying regularity assumptions,
    independent of training history
  & Statistical: degrades for inputs outside training distribution; rate of
    degradation generally unpredictable without additional assumptions \\[4pt]
Failure mode character
  & Identifiable and quantifiable: spurious oscillations (Gibbs), pressure
    locking, non-physical modes, CFL violation---each with known cause and remedy
  & Often opaque: physically inadmissible outputs, silent distributional failure,
    accumulated rollout errors not detectable from internal model quantities \\[4pt]
Human-in-the-loop form
  & Geometric and mathematical: mesh generation, solver parameter selection,
    stability analysis, preconditioner design
  & Statistical and engineering: architecture search, hyperparameter tuning,
    training pipeline design, validation dataset construction \\
\bottomrule
\end{tabular}
\end{table}

One important corollary deserves emphasis: the two paradigms do not differ primarily
in the \emph{class of problems they target}.
Both can in principle be applied to the same PDEs.
They differ in \emph{what they can certify about the result}.
The choice between them---or the decision to combine them---should therefore be driven
first by the certification requirements of the application.
A structural integrity simulation for which a certification authority requires error
bounds is not simply a harder problem than an uncertainty quantification study for which
approximate ensemble statistics suffice; it is a problem with a different epistemological
requirement that determines which paradigm is appropriate.

\subsection{Convergent Evolution: Structural Parallels Between Paradigms}
\label{subsec:convergent_evolution}

Despite their epistemological differences, classical PDE solvers and machine learning approaches have undergone striking convergent evolution: independently developed methods have converged on structurally similar computational designs. This convergence highlights where successful hybrid strategies are most likely to emerge and where theoretical insights can transfer across paradigms. Figure~\ref{fig:convergent_evolution} maps this convergence explicitly, focusing on the four most structurally significant parallels documented in the literature. The figure is structured as two vertical trunks—classical methods ascending from the left, machine learning methods from the right—both rooted in the same three foundational mathematical domains: operator and functional analysis, variational principles and Sobolev theory, and spectral/harmonic analysis. Four horizontal cross-connections link independently invented method pairs; each level is colour-coded by its shared mathematical core, and the central box for each parallel presents three layers: a structural descriptor, the governing equations, and the key distinction that separates the two approaches despite their architectural similarity.

Convergent evolution in this space is not coincidental. It reflects deep mathematical constraints on the efficient approximation of PDE solution manifolds—specifically, the prevalence of low effective dimensionality, hierarchical organisation, and rich spectral content—which reward designs that exploit these properties, regardless of whether the framework is classical or data-driven.

\paragraph{Hierarchical information propagation.}
Geometric multigrid (Section~\ref{subsec:crosscutting}) and graph neural network message
passing (Section~\ref{subsubsec:gnns}) both propagate corrections across spatial scales
through hierarchical sequences of coarser representations.
In multigrid, the hierarchy is geometric and the transfer operators are derived from
the differential operator structure; in GNNs, the pooling hierarchy is learned from data.
The mathematical purpose---efficiently transporting long-range information while managing
local computation---is identical, and recent work has formalized multigrid V-cycles as
special cases of message-passing architectures with fixed operator-derived weights~\cite{luz2020learning}.

\paragraph{Spectral representation and learned filtering.}
Spectral Galerkin methods (Section~\ref{subsec:spectral}) and Fourier Neural Operators
(Section~\ref{sec:neural_operators}) both operate through truncated spectral expansions.
The key distinction is that spectral methods use analytically prescribed basis functions
with approximation properties characterized by classical harmonic analysis, while FNO
learns spectral filters from data.
This parallel makes explicit that FNO inherits both the advantages of spectral
representation---efficient global solution structure---and the same Gibbs-type degradation
for non-smooth solutions that limits classical spectral methods
(Section~\ref{subsec:eval_discontinuities}), for precisely the same mathematical reasons.

\paragraph{Low-dimensional solution manifolds.}
The Reduced Basis method (Section~\ref{sec:model-order-reduction}) and neural operators
(Section~\ref{sec:neural_operators}) both exploit the observation that the solution
manifold of a parametric PDE is often low-dimensional, even when individual solutions
require many degrees of freedom to represent.
The structural relationship is precise: DeepONet's branch-trunk decomposition is a
nonlinear generalization of the POD-Galerkin ansatz, and FNO's spectral layers realize a
learned coarse space analogous to $V_0$ in the subspace correction framework of
equation~\eqref{eq:subspace-correction}.

\begin{figure}[H]
    \centering
    \includegraphics[width=0.9\textwidth]{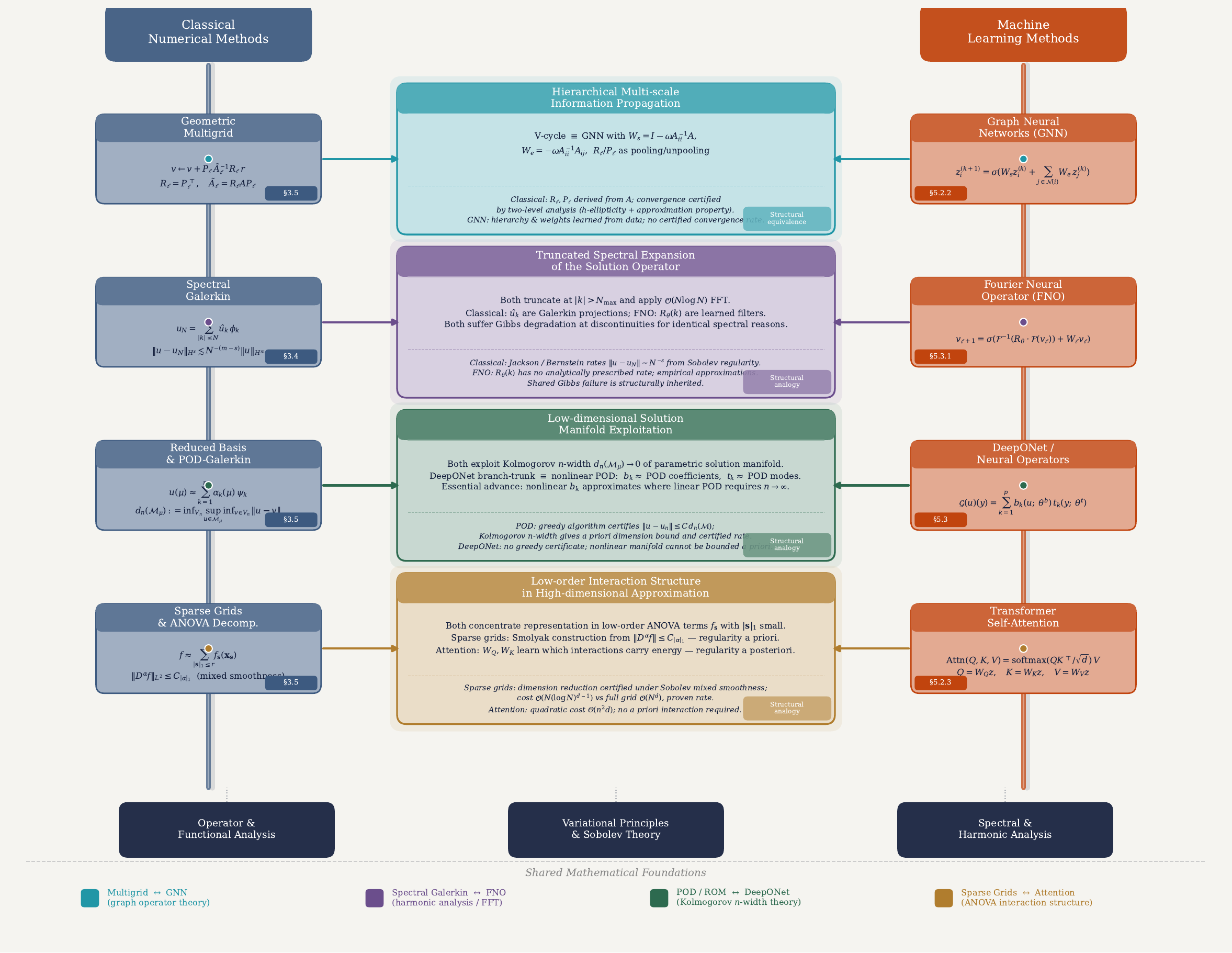}
    \caption{Convergent evolution of classical numerical methods and machine learning approaches toward shared mathematical structures for PDE approximation. The figure depicts classical solvers (left) and ML methods (right), both rooted in three common mathematical foundations: operator/functional analysis, variational principles and Sobolev theory, and spectral/harmonic analysis. Four horizontal levels connect independently developed method pairs. Each level is colour-coded by its shared mathematical domain. Central convergence boxes comprise three tiers: the structural parallel, the precise governing equations/mathematical equivalence, and the key distinction that separates the approaches despite their shared architecture.
    \textbf{Row 1:} Geometric multigrid $\leftrightarrow$ graph neural networks (GNNs). Both propagate corrections hierarchically across scales. A multigrid V-cycle is mathematically equivalent to a GNN with fixed aggregation weights $W_s$, $W_e$ determined by the operator $A$ rather than learned. Classical multigrid guarantees convergence via two-grid analysis ($h$-ellipticity, approximation property); GNNs learn the hierarchy from data without certified rates.
    \textbf{Row 2:} Spectral Galerkin $\leftrightarrow$ Fourier Neural Operator (FNO). Both rely on truncated spectral expansions computed in $\mathcal{O}(N \log N)$ via FFT, yielding identical Gibbs ringing at discontinuities. Classical methods certify convergence via Jackson/Bernstein theory; FNO's learned filter $R_\theta(k)$ lacks prescribed approximation rates.
    \textbf{Row 3:} Reduced Basis / POD-Galerkin $\leftrightarrow$ neural operators (DeepONet, FNO). Both exploit low Kolmogorov $n$-width of parametric solution manifolds. DeepONet's branch-trunk decomposition nonlinearly generalises the POD ansatz (trunk $\approx$ modes, branch $\approx$ coefficients), but learned rather than SVD-derived. This enables approximation in regimes requiring many POD modes, at the expense of the greedy certificate and a priori dimension bound from $n$-width theory.
    \textbf{Row 4:} Sparse grids / Smolyak construction $\leftrightarrow$ transformer self-attention. Both capitalise on energy concentration in low-order ANOVA terms of high-dimensional functions. Sparse grids analytically derive interactions from mixed Sobolev smoothness $\|D^\alpha f\|_{L^2} \leq C_{|\alpha|_1}$, reducing cost from $\mathcal{O}(N^d)$ to $\mathcal{O}(N (\log N)^{d-1})$ with certification. Transformer attention learns informative interactions from data, forgoing a priori regularity assumptions at the cost of $\mathcal{O}(n^2 d)$ scaling and no certified dimension reduction.
    Convergent evolution across these parallels is driven by a single mathematical imperative: efficient PDE approximation---classical or learned---must exploit the low effective dimensionality, hierarchical organisation, and spectral content of solution manifolds.}
\label{fig:convergent_evolution}
    \label{fig:convergent_evolution}
\end{figure}

The essential advance of neural operators over classical ROMs is that the learned
representation can be nonlinear, enabling accurate approximation where POD subspaces
require prohibitively many modes---but this comes at the cost of the projection-error
bounds and greedy selection theory that make classical ROMs certifiable.

\paragraph{Sparse approximation and attention.}
Sparse grid methods (Section~\ref{subsec:eval_dimensionality}) and transformer attention
mechanisms (Section~\ref{subsubsec:transformers}) both exploit the observation that
high-dimensional functions often have most of their energy concentrated in low-order
interaction terms.
Sparse grids do this analytically using mixed derivative regularity assumptions;
attention mechanisms do it statistically, learning which spatial dependencies carry
the most information from the training distribution.
The former provides rigorous dimension-reduction guarantees under smoothness hypotheses;
the latter adapts to unknown interaction structure at the cost of generalization bounds.

These parallels are neither coincidental nor superficial. They stem from a single, deep mathematical imperative: efficient approximation of PDE solution manifolds—whether classical or learned—must exploit their intrinsic low effective dimensionality, hierarchical organisation, and rich spectral content. Methods that harness these properties succeed across frameworks; those that do not remain fundamentally constrained by the curse of dimensionality. The most promising hybrid designs, therefore, lie at the precise intersection of both paradigms: they combine the structural guarantees and deductive certification of classical methods with the adaptive, data-driven exploitation of machine learning. This convergence directly informs the disciplined hybrid programme of Section~\ref{subsec:hybrid_discipline}.

The enduring distinction between the two approaches is epistemological rather than computational: classical methods certify their exploitation of these manifold properties through rigorous analytic theorems; machine learning methods exploit the same properties statistically from data, trading deductive guarantees for greater flexibility and generalisation power. Far from a limitation, this difference constitutes the exact basis of their complementarity—the foundation that Section~\ref{subsec:complementarity} formalises.

\subsection{Genuine Complementarity and Its Boundaries}
\label{subsec:complementarity}

\begin{figure}[H]
    \centering
    \includegraphics[width=\textwidth]{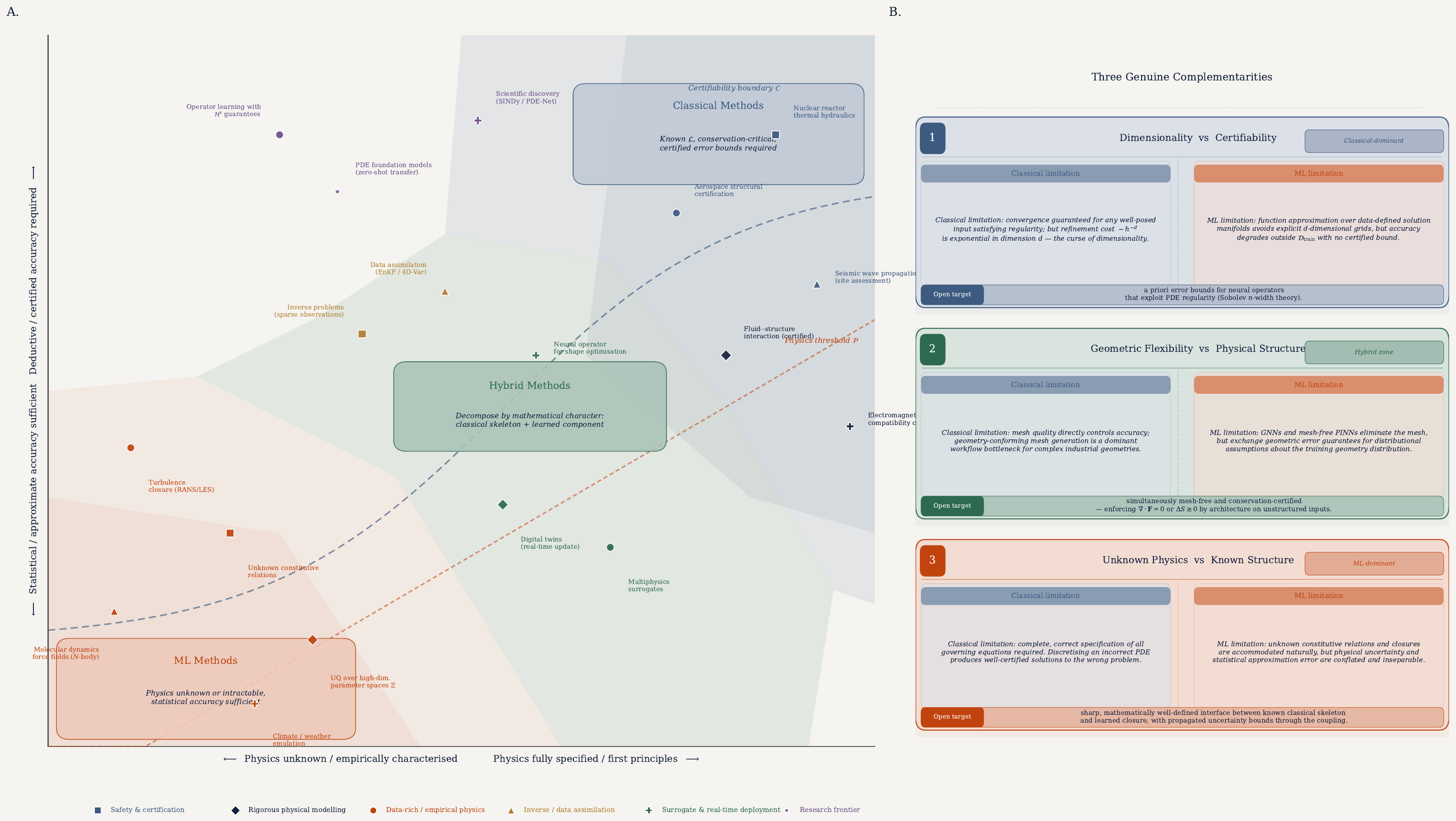}
    \caption{Complementarity phase space for classical numerical methods and
    machine learning approaches to PDE solution.
    \textbf{A.}~Two-dimensional deployment space spanned by the degree
    of physics specification (horizontal axis, from empirically
    characterised to be fully specified from first principles) and the
    certification requirement (vertical axis, from statistical accuracy
    sufficient to deductive certified accuracy required).
    Three shaded regions demarcate the natural domain of each paradigm:
    the classical-dominant zone where the governing operator $\mathcal{L}$ is fully known and certification is required; the ML-dominant zone where physics     is unknown or intractable and statistical accuracy suffices; and the hybrid transition where the problem
    decomposes naturally into a classical skeleton and a learned
    component.
    Two dashed boundary curves mark the certifiability threshold
    $\mathcal{C}$ (above which deductive error bounds are required) and
    the physics specification threshold $\mathcal{P}$ (to the right of
    which the governing equations are sufficiently characterised for
    classical discretisation); the hybrid zone is the region between
    these boundaries.
    Eighteen representative application domains are plotted as labelled
    points, colour-coded by discipline cluster: safety and certification, rigorous physical modelling, data-rich and empirical physics, inverse and
    data assimilation problems, surrogate and real-time deployment, and the research frontier
    methods. The deployment decisions depend simultaneously
    on both axes.
    \textbf{B.}~The three genuine complementarities identified in
    Section~\ref{subsec:complementarity}, each corresponding to a zone
    of panel~A.
    Each box presents the intrinsic limitation of each paradigm
    and the \emph{open target} — the capability that a well-designed hybrid
    method must achieve to address both limitations simultaneously.
    Complementarity~1: classical refinement guarantees convergence but costs $\mathcal{O}(h^{-d})$ in dimension
    $d$; ML avoids the curse of dimensionality but provides no certified
    bound outside the training distribution $\mathcal{D}_{\mathrm{train}}$.
    Complementarity~2: classical methods enforce
    conservation and inf-sup conditions by construction, but require
    geometry-conforming meshes; ML methods are mesh-free but trade
    geometric error guarantees for distributional assumptions on the
    training geometry.
    The open target for Complementarity~2 — simultaneously mesh-free
    and conservation-certified, enforcing $\nabla \cdot \mathbf{F} = 0$
    or $\Delta S \geq 0$ by architecture on unstructured inputs — is a
    core unsolved problem of hybrid design.
    Complementarity~3: classical methods
    certify solutions to completely and correctly specified problems; ML
    methods accommodate unknown constitutive relations and empirical
    closures, but at the cost of conflating physical uncertainty and
    statistical approximation error with no mechanism for separating
    the two.
    These limitations are intrinsic in the sense of
    Section~\ref{subsec:complementarity}: they follow from the
    foundational assumptions of each paradigm and cannot be resolved by
    engineering refinement within that paradigm alone.}
    \label{fig:complementarity_phase_space}
\end{figure}

The claim that classical and ML methods are complementary is made frequently but rarely
made precise. Complementarity has a specific meaning that distinguishes it from mere coexistence: two methods are genuinely complementary when each addresses a difficulty that is
\emph{intrinsic} for the other---not merely inconvenient, but rooted in its
foundational assumptions. Three genuine complementarities can be identified. Figure~\ref{fig:complementarity_phase_space} provides a geometric interpretation of this tripartite structure. Panel~A organises PDE deployment scenarios in a two-axis phase space that emphasises epistemological character over mere computational difficulty: the horizontal axis represents the degree to which the governing physics is known and expressible from first principles, while the vertical axis captures the stringency of required accuracy certification. The classical-dominant regime occupies the upper-right quadrant (fully specifiable physics + strict certification), the natural domain of finite element, finite volume, and spectral methods. The ML-dominant regime lies in the lower-left quadrant (intractable or unknown physics + statistical accuracy sufficient), encompassing turbulence closures, molecular force fields, climate emulation, and high-dimensional uncertainty quantification. The hybrid transition band forms a distinct diagonal zone—not merely an overlap, but a region defined by problems with partially known operators, geometries or parameter spaces that defeat classical mesh refinement, or empirically accessible yet analytically non-derivable closure terms. The application domains plotted in Panel~A are positioned according to real deployment constraints that dictate the appropriate paradigm or hybrid decomposition. Figure~\ref{fig:complementarity_phase_space}B sharpens the argument: for each of the three complementarities, it states both paradigms' limitations in precise mathematical terms and explicitly identifies the open target that a principled hybrid method must achieve.

\paragraph{Dimensionality versus certifiability.}
Classical methods are fundamentally limited by the curse of dimensionality: accuracy is
guaranteed through refinement hierarchies, but refinement costs grow exponentially with
dimension.
ML methods address this by replacing explicit discretization with function approximation
over data-defined solution manifolds.
Conversely, ML methods are fundamentally limited by distributional generalization:
accuracy is guaranteed only within statistical neighborhoods of the training distribution,
with no mechanism for certified extrapolation.
Classical methods address this through deductive error bounds that hold for any input
satisfying regularity assumptions, without reference to historical data.
These are genuine limitations: classical discretizations typically suffer from exponential or rapidly increasing complexity with dimension, and no purely statistical learner provides deductive
guarantees for inputs not covered by its training distribution.

\paragraph{Geometric flexibility versus physical structure.}
Classical methods require geometric conformity: mesh quality directly affects accuracy
(Section~\ref{subsec:eval_geometry}), and labor-intensive mesh generation remains a
dominant workflow bottleneck.
ML methods---particularly GNNs and mesh-free PINNs---relax the requirement for geometry-conforming meshes,
but exchange geometric error guarantees for distributional assumptions about the training
geometry set.
Conversely, classical methods encode physical structure precisely and certifiably:
conservation laws, inf-sup conditions, symplectic geometry, entropy dissipation are
satisfied by construction.
ML methods enforce these properties approximately through soft constraints or not at all
in purely data-driven architectures.
A method that is simultaneously mesh-free and conservation-certified would address
fundamental limitations of both---and designing such methods is a core challenge of the
hybrid research agenda.

\paragraph{Unknown physics versus known structure.}
Classical methods require complete and correct specification of governing equations.
When constitutive relations are unknown, when closure terms are empirically inaccessible,
or when the governing equations themselves are uncertain, classical discretization of an
incorrect PDE produces well-certified solutions to the wrong problem. ML methods accommodate unknown or partially known physics naturally, at the cost of
confounding physical uncertainty with statistical approximation error.

These three complementarities are genuinely interdependent rather than independent. Classical methods certify solutions to fully known problems, while machine learning excels at representing unknown or intractable components---but only when the boundary between known and unknown physics is sharply defined and the interface between classical and learned sub-components is mathematically rigorous. Figure~\ref{fig:complementarity_phase_space} reveals why this interface is non-negotiable: the certifiability threshold $\mathcal{C}$ and physics-specification threshold $\mathcal{P}$ in Panel~A intersect precisely within the hybrid transition band, and the three open targets in Panel~B share a single common requirement---a verifiable, structure-preserving interface that propagates classical guarantees in one direction while bounding statistical approximation error in the other. Without such an interface, hybrid methods remain largely empirical rather than theoretically grounded.

\subsection{Hybrid Design as a Principled Discipline}
\label{subsec:hybrid_discipline}

The genuine complementarities above motivate hybrid approaches.
But the discipline of hybrid design is not merely a catalog of specific instantiations.
It requires principles governing \emph{which} components to learn, \emph{how} to couple
learned and classical components without compromising the guarantees of either, and
\emph{what} mathematical properties the combined system can be shown to possess.

\subsubsection{The Decomposition Principle}

The first design question for any hybrid method is the decomposition of the problem into
components that are handled classically and components that are learned.
This decomposition should be guided by the mathematical character of each component.

Components that are appropriate targets for learning share identifiable characteristics:
they represent constitutive or closure relationships that depend on material-specific
or configuration-specific information not encoded in first principles; their mathematical
form is known but their parameters are uncertain or spatially variable; their
computational cost dominates relative to a well-understood classical skeleton; or their
solution manifold is demonstrably low-dimensional relative to its ambient representation
space.
Components that should remain classical are those whose mathematical character is
precisely known (linear operator structure, conservation form, inf-sup constraints),
whose error must be certified independently of data, or whose failure modes in the
classical setting are well understood and manageable.

This principle can be stated operationally: \emph{learn what cannot be derived from
first principles; compute classically what can.}
The sharpness of this boundary directly determines how much of the classical method's
guarantees the hybrid system inherits.
When the boundary is sharp---as in neural constitutive models integrated into an
otherwise standard FEM solver with intact variational structure and conservation
properties---the hybrid system inherits well-posedness, stability, and convergence from
the classical skeleton while the learned component contributes expressiveness in the
subspace of constitutive behavior.
When the boundary is diffuse---as in end-to-end neural solvers that learn the full
mapping from PDE coefficients to solutions---no classical guarantees propagate and the
system's reliability is purely statistical.

\subsubsection{The Structure Inheritance Problem}

Even when the decomposition principle is respected, coupling learned and classical
components raises a mathematical question not yet fully resolved: under what conditions
does the structure of the governing equations propagate through the hybrid
coupling to certify properties of the combined system?

For hybrid methods where the learned component is a constitutive model embedded in
a variational framework, partial answers exist.
If the neural constitutive model is derived from a convex potential
$\Psi_{\mathrm{NN}}(\varepsilon)$ with $\sigma = \partial\Psi_{\mathrm{NN}}/\partial\varepsilon$,
thermodynamic consistency and positive dissipation are guaranteed by architecture
regardless of training outcome, and the classical solver inherits well-posedness from
the material stability of the learned model~\cite{klein2022polyconvex}.
More generally, if the learned component is constrained to lie in a class of operators
that preserves the relevant mathematical structure---monotone operators for elliptic
problems, dissipative operators for parabolic problems, symplectic maps for Hamiltonian
systems---then the hybrid system inherits the corresponding qualitative properties.

For hybrid formulations involving tightly coupled PDE-learning loops---differentiable
physics, Neural ODEs embedded in multistep solvers, online-coupled adaptive mesh
controllers---structure inheritance is far less understood.
Combined errors from discretization and neural approximation may interact non-additively,
and classical stability analysis does not extend to the coupled system without new
theoretical work.
This is the \emph{structure inheritance problem}: developing mathematical frameworks that
track how much classical structure is preserved as learned components are introduced, and
providing design guidelines that ensure this preservation when required.

\subsubsection{The Error Budget Framework}

A practical tool for hybrid design is the explicit decomposition of total solution error
into distinct and separately controllable contributions:
\begin{equation}
\|u - u_{\mathrm{hybrid}}\|
\;\lesssim\;
\underbrace{\|u - u_h\|}_{\text{discretization}}
\;+\;
\underbrace{\|u_h - \hat{u}_h\|}_{\text{neural approximation}}
\;+\;
\underbrace{\|\hat{u}_h - u_{\mathrm{hybrid}}\|}_{\text{coupling}},
\label{eq:error_budget}
\end{equation}
where $u_h$ denotes the solution of the classically discretized problem, $\hat{u}_h$ the
solution with the learned component evaluated at its training-distribution accuracy, and
$u_{\mathrm{hybrid}}$ the output of the coupled system.

\begin{figure*}[h!]
  \centering
  \includegraphics[width=0.95\textwidth]{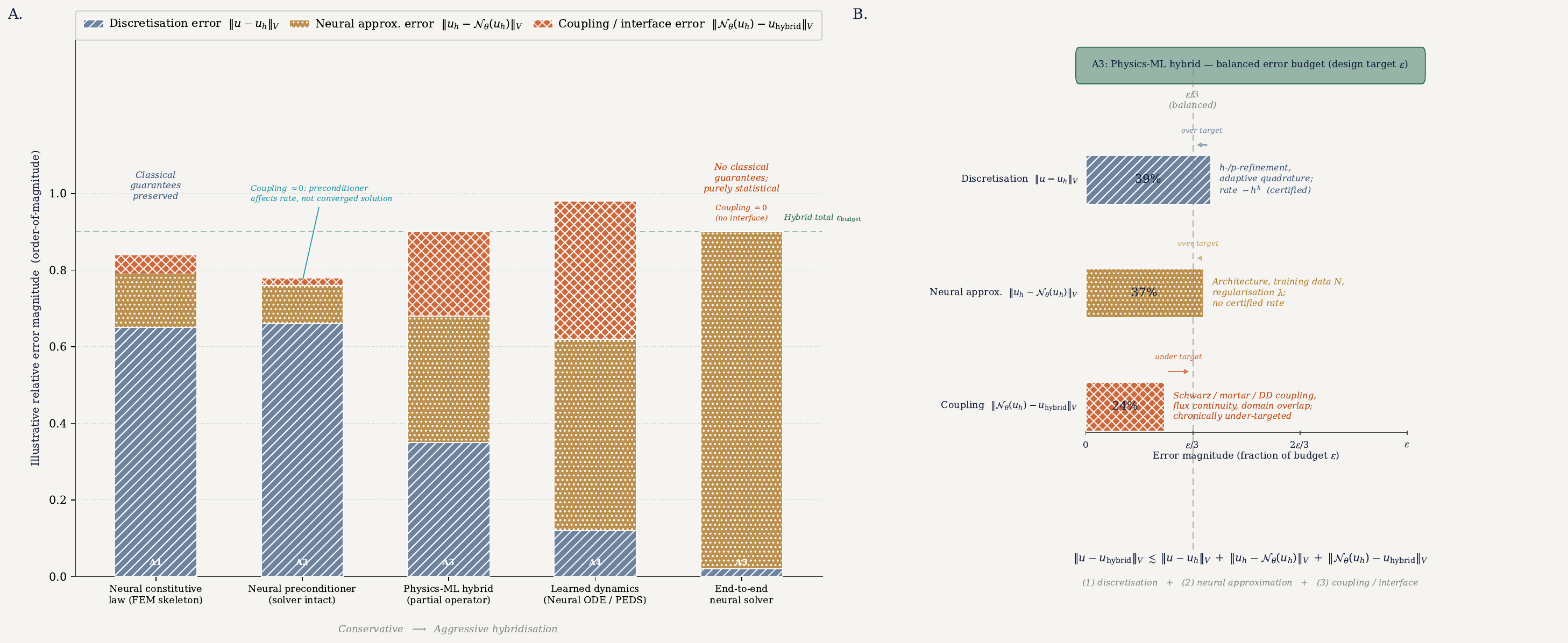}
  \caption{%
    Error budget decomposition across hybrid design archetypes
    (Eq.~\eqref{eq:error_budget}: terms (1)~discretisation,
    (2)~neural approximation, (3)~coupling).
    \textbf{A.}~Illustrative relative magnitudes of all three terms across five archetypes, ordered conservative to aggressive.
    \emph{A1~Neural constitutive law}: FEM skeleton intact; classical
    $h$-$p$ certificates preserved; coupling small.
    \emph{A2~Neural preconditioner}: coupling structurally near-zero,
    as the preconditioner affects convergence rate only, not the
    converged solution.
    \emph{A3~Physics-ML hybrid}: all three terms contribute; closest
    to a balanced budget.
    \emph{A4~Learned dynamics} (Neural ODE\,/\,PEDS): coupling
    accumulates per autoregressive step and dominates alongside
    neural approximation error.
    \emph{A5~End-to-end neural solver} (e.g.\ FNO, DeepONet, PINN):
    no classical skeleton; coupling is identically zero; effectively
    all error is neural approximation with no computable bound.
    Dashed line marks the A3 total budget~$\varepsilon$ used as
    normalisation in panel~B.
    \textbf{B.}~Balanced error budget for A3.
    Bars show the fraction of~$\varepsilon$ per term; dashed line
    at~$\varepsilon/3$ is the equal-investment target; arrows indicate
    over- or under-targeting.
    Control mechanisms (right): mesh refinement at certified
    rate~${\sim}\,h^{k}$ for~(1); architecture, data~$N$,
    regularisation~$\lambda$ with no certified rate for~(2);
    Schwarz\,/\,mortar\,/\,DD coupling and flux continuity for~(3).
    Term~(3) is chronically under-targeted because no computable
    a~posteriori estimator for
    $\|\mathcal{N}_{\theta}(u_h) - u_{\mathrm{hybrid}}\|_{V}$
    yet exists (cf.\ open problem~P4,
    \S\,\ref{subsec:open_problems}).%
  }
  \label{fig:error_budget}
\end{figure*}

The three terms in~\eqref{eq:error_budget} are separately controllable
by distinct mechanisms.
Term~(1) is governed by classical analysis: mesh refinement and
high-order approximation reduce it at a certified rate $\sim h^{k}$,
and a posteriori estimators provide computable upper bounds.
Term~(2) is governed by statistical learning: architecture capacity,
training-set size~$N$, and regularisation strength~$\lambda$ collectively
determine the neural approximation quality within the training distribution,
but no certified convergence rate exists.
Term~(3) is governed by the coupling formulation---interface conditions,
domain overlap, and flux continuity at the classical--ML boundary---yet
it is the least understood of the three: no computable a posteriori
estimator for $\|\mathcal{N}_{\theta}(u_h) - u_{\mathrm{hybrid}}\|_V$
currently exists.
Designing with an explicit error budget serves two purposes.
First, it enforces a sharp decomposition: if classical and neural
contributions cannot be separated, the decomposition is not sufficiently
precise.
Second, it provides a principled basis for resource allocation: a
\emph{balanced} budget---in which no single term dominates---is the
natural analogue of classical mesh optimisation and is the operative
design target for any efficient hybrid method.
Figure~\ref{fig:error_budget} maps the three terms across five
representative hybrid archetypes and illustrates the
balanced-budget target for the intermediate hybrid case.

Three observations from Fig.~\ref{fig:error_budget} are consequential
for hybrid design.
First, the two extreme archetypes are epistemologically clean but
practically limited in opposite directions.
The neural constitutive law~(A1) preserves all classical $h$-$p$
convergence certificates and represents the lowest-risk hybrid strategy;
the end-to-end neural solver~(A5) eliminates coupling error by
eliminating the classical interface entirely, but surrenders all
deductive error control in exchange.
Neither extreme can exploit the complementarity described in
\S\ref{subsec:complementarity}: the former does not leverage ML
flexibility where it is most needed; the latter sacrifices the
certification capability that the classical paradigm provides.
Second, the neural preconditioner~(A2) occupies a structurally
distinguished position: it is the only archetype for which coupling
error is \emph{provably} negligible, because a preconditioner affects
only the convergence rate and not the solution to which the iterative
solver converges.
This makes A2 the safest form of ML augmentation for
certification-critical applications---a property that is
under-exploited in the literature relative to its theoretical standing.
Third, and most consequentially, term~(3) is chronically under-targeted
across all intermediate archetypes.
Terms~(1) and~(2) each have established control mechanisms with
quantifiable effect; term~(3) has neither a computable a posteriori
estimator nor a systematic control strategy.
This is precisely the gap that open problem~P4
(\S\ref{ssubsec:open_problems}) must fill: without a principled
indicator for term~(3), the balanced budget of panel~B remains a
design aspiration rather than an achievable specification.

The error budget framework of Fig.~\ref{fig:error_budget} does not
resolve the open problems of \S\ref{subsec:open_problems}---it makes
their stakes precise.
Without structure inheritance theory~(P1) to bound coupling by
construction, without PDE-aware generalisation bounds~(P2) to certify
term~(2), and without a computable indicator for term~(3)~(P4), the
balanced budget of panel~B can be targeted empirically but never
certified.
Until these gaps are closed, hybrid design remains a collection of
engineering practices rather than a mathematical theory.

\subsubsection{Validation as a First-Class Design Requirement}

Classical methods have established verification and validation (V\&V) protocols that are
not incidental to method development but part of the mathematical framework itself:
the method of manufactured solutions~\cite{roache2002code} tests convergence rates;
a posteriori error estimators certify solution quality; community benchmarks provide
reproducible, consensus‑based reference solutions without the existence of formal or fine-scale solutions~\cite{spe11}.
Methods are not considered mature until V\&V infrastructure is in place.

Hybrid methods require a richer V\&V framework that simultaneously addresses both
classical and statistical failure modes, and current practice falls short.
Four types of testing are necessary.
\emph{Convergence testing} verifies that the classical skeleton converges at expected
rates when the learned component is replaced by its exact value, confirming that no
classical guarantees have been silently compromised by the coupling.
\emph{Generalization testing} verifies that the learned component behaves reliably for
problem instances outside its training distribution, including corner cases near
mathematical boundaries such as phase transitions, bifurcations, and singularities.
\emph{Physical consistency testing} verifies that the hybrid system satisfies discrete
conservation laws and structural constraints to the same tolerance as the classical
skeleton alone.
\emph{Uncertainty propagation testing} verifies that both discretization uncertainty and
neural approximation uncertainty are correctly propagated through the hybrid coupling to
provide certified bounds on quantities of interest.

Developing standardized V\&V protocols for hybrid methods is among the highest-priority
practical needs of the field.
Nascent benchmark platforms such as PDEBench~\cite{takamoto2022pdebench} and
APEBench~\cite{koehler2024apebench} provide initial infrastructure for statistical
evaluation; extending them to include convergence metrics and physical consistency checks
would substantially reduce the validation burden on individual research groups.

\subsection{Consequential Open Problems}
\label{subsec:open_problems}

The following open problems are not isolated technical gaps but structural questions
whose resolution would define the theoretical foundations of the combined paradigm. The preceding analysis identifies specific mathematical questions whose resolution would
have broad impact across method families and application domains.

\paragraph{Generalization bounds with PDE structure.}
Classical statistical learning theory provides generalization bounds in terms of hypothesis
class complexity (VC dimension, Rademacher complexity), but these bounds are vacuous for
the large networks used in practice.
The key open question is whether PDE structure---regularity theory, compact solution
manifolds, spectral properties---enables tighter generalization bounds for neural PDE
solvers than for general function approximation.
Preliminary results establishing approximation rates for neural operators in Sobolev
spaces~\cite{kovachki2023neural} suggest that PDE structure can be exploited, but a
complete theory connecting training error, architecture choice, and deployment accuracy
through PDE regularity theory does not yet exist.

\paragraph{Long-time stability theory for learned dynamics.}
Classical stability theory for time-dependent PDEs provides systematic tools---energy
estimates, Lyapunov analysis, CFL conditions---that bound solution growth over arbitrary
time horizons.
For neural surrogates in autoregressive rollout, the analogous question is: under what
conditions on the trained model do solution trajectories remain bounded, accurate, and
physically consistent over long integration times?
This question is particularly acute for chaotic systems, where classical stability
analysis yields statistical rather than pointwise bounds (Lyapunov exponents, attractor
characterization), and where neural surrogates must capture not just short-time dynamics
but the system's attractor structure.
Structure-preserving architectures---symplectic integrators, energy-dissipating recurrent
networks---address this partially, but a general stability theory for learned dynamics
analogous to classical energy analysis remains open.

\paragraph{Adaptive hybrid decomposition.}
Current hybrid methods use fixed decompositions: the boundary between classical and
learned components is set at design time and does not adapt during computation.
A natural extension---with precedent in classical AMR, hp-adaptivity, and adaptive
time-stepping---would identify, at runtime, regions or time intervals where classical
guarantees are required, due to proximity to discontinuities, physical constraint
violations, or loss of learned component confidence, and automatically revert to
classical computation in those regions.
The mathematical challenge is defining a reliable, efficiently evaluable indicator
function for this switching decision, ideally based on a posteriori estimators for the error budget decomposition~\eqref{eq:error_budget}.
Calibrated uncertainty quantification from the learned component is a natural candidate,
but calibrated UQ for deterministic neural surrogates---distinct from probabilistic
generative models---remains technically challenging.

\paragraph{Certified transfer for PDE foundation models.}
The scientific question raised by foundation models pretrained on diverse PDE families
is what constitutes a meaningful \emph{distribution} over PDE problems and under what
conditions transfer is reliable.
For large language models, the pretraining distribution is a corpus of human text and
transfer is grounded in approximately shared linguistic structure.
For PDE foundation models, the analogous hypothesis---that solution operators across
different PDE families share transferable structure---is physically motivated by shared
mathematical ancestry (Lagrangian mechanics, conservation forms, elliptic operator theory)
but has not been formalized.
Establishing a theory of \emph{certified transfer}---characterizing which structural
properties of a target problem allow zero-shot application of a pretrained model with
bounded error, and which require fine-tuning or retraining---would transform foundation
models from an empirical phenomenon into scientifically grounded capability.

\paragraph{Scientific discovery versus scientific computing.}
A distinction that runs beneath much ML-for-PDE research but is rarely made explicit
separates two fundamentally different scientific goals: \emph{scientific computing}
(solving known equations efficiently) and \emph{scientific discovery} (identifying
governing equations or constitutive relationships from data).
These goals have different validation requirements: computing requires accuracy and
certification relative to a known ground truth; discovery requires the identified
model to be interpretable and generalizable beyond observed data.
Many current ML-PDE methods conflate the two---using neural surrogates simultaneously
as solvers and implicit model identifiers---in ways that can compromise both objectives.
A clearer taxonomy distinguishing computing, discovery, and hybrid objectives, together
with validation criteria appropriate to each, would improve both research design and the
evaluation and communication of results.

\subsection{A Grounded Vision for the Combined Paradigm}
\label{subsec:vision}

The trajectory of the field does not point toward a single dominant paradigm, but toward
a unified computational framework in which classical structure-preserving methods and
adaptive learned representations are combined according to the mathematical character
of each component.

The boundary between classical and learned components in any hybrid system would be
mathematically sharp and verifiable: the classical skeleton would carry identifiable
convergence rates and conservation properties, the learned component would carry
calibrated uncertainty bounds, and their coupling would be governed by mathematical
conditions analogous to the inf-sup stability conditions that govern multiphysics
coupling in classical methods (Section~\ref{subsec:multiphysics}).
The structure inheritance problem would be answered for the most important problem classes,
and hybrid methods designed against its criteria would inherit classical guarantees in
the domains where those guarantees are required.

The V\&V framework for hybrid methods would be standardized and routinely applied:
convergence testing, physical consistency verification, and out-of-distribution
robustness assessment would be as expected as mesh refinement convergence studies
are for classical methods today.
Community benchmarks spanning both accuracy metrics and physical consistency metrics
would provide shared ground truth for method comparison that does not depend on the
choice of baseline.

The human-in-the-loop burden---currently redistributed between paradigms rather than
genuinely reduced---would decrease through methods that automate the most expertise-intensive
steps of both: geometry-to-simulation pipelines that reduce the classical preprocessing
bottleneck (Section~\ref{subsec:eval_geometry}) and architecture-selection protocols
that reduce the ML hyperparameter burden (Section~\ref{subsubsec:computational_challenges}),
both guided by mathematical problem analysis rather than empirical search.

Most fundamentally, the field would develop a mathematical language adequate to the
combined paradigm---a framework in which discretization errors, approximation errors, and
statistical generalization errors are commensurable, in which stability analysis applies
to coupled classical-learned systems, and in which design principles for hybrid methods
derive from the same operator-theoretic foundations that underpin classical numerical
analysis.
The convergent evolution documented in Section~\ref{subsec:convergent_evolution} suggests
this language is not entirely new but a generalization of existing frameworks that must
be extended to accommodate the statistical character of learned components.

The path toward this vision passes through the hybrid methods of Section~\ref{sec:hybrid_ml},
but does not end there.
Each open problem in Section~\ref{subsec:open_problems} represents a specific research
program with clear success criteria.
Progress on any of them advances not just the directly studied methods but the broader
project of building a rigorous, verifiable, and practically powerful computational science
of PDE solution---one that deserves the confidence of the scientific computing community
and of the safety-critical engineering applications that community serves.

\section{Emerging Frontiers and Open Problems}
\label{sec:emerging}

The landscape described in the preceding sections is not static. Several technological
and algorithmic developments are already shifting the boundaries of what is
computationally feasible for PDE solution, and they are likely to shape how classical
and machine-learning methods interact over the coming decade. This section examines
four such developments—foundation models for scientific computing, quantum algorithms,
differentiable programming for inverse design, and exascale architectures—with a focus
on their current status, the barriers limiting near-term impact, and the conditions
under which each could materially change the capability landscape. The aim is not to
predict which direction will dominate, but to distinguish structural barriers from
those that are likely to yield to continued engineering progress.

\subsection{Foundation Models: From Task-Specific Surrogates to General Operators}
\label{subsec:foundation_models_emerging}

Section~\ref{subsubsec:transformers} introduced PDE foundation models as an emerging
extension of the transformer paradigm, with early demonstrations of limited zero-shot
and few-shot generalization \cite{herde2024poseidon,ye2025pdeformer,sun2025towards,liu2024few,chen2025flow}.
The central scientific question is whether one can define a meaningful distribution
over PDE problems such that pretraining on this distribution yields transferable
solution operators, analogous to the way pretraining on natural language corpora
produces transferable linguistic representations.

There are genuine reasons to think the answer could be positive. Many PDE families
share deep mathematical structure, including conservation forms, elliptic operator
structure, and Lagrangian or Hamiltonian formulations. Solution manifolds arising from
superficially different equations may therefore share geometric properties such as
smoothness, low effective dimensionality, or spectral concentration that a sufficiently
expressive pretrained model could exploit. The central open question, however, is
whether this shared structure is rich enough to support reliable zero-shot transfer
across PDE families with qualitatively different solution behavior, including smooth
elliptic problems, hyperbolic shock dynamics, and stiff reaction--diffusion systems
\cite{menon2026scientific,subramanian2023towards}.

Before foundation models can serve as general-purpose PDE infrastructure rather than
specialized surrogates for narrow equation families, three conditions must be met.
First, \emph{dataset curation at scientific scale} requires solution libraries spanning
diverse equation classes, parameter regimes, geometries, and physical scales, all
generated and validated to the accuracy standards of classical numerical methods. This
is itself a substantial computational investment whose cost must be amortized across a
broad enough application portfolio. Second, \emph{structure-respecting architectures}
must encode PDE-specific properties such as symmetries, conservation laws, and
variational principles without sacrificing the flexibility that makes pretraining
valuable; that balance between universality and structure remains unresolved. Third,
\emph{principled transfer metrics} must replace empirical held-out accuracy as the
primary validation criterion. A model should be deployable on a new PDE family only
when mathematical conditions for reliable transfer can be verified, and such
conditions do not yet exist in a formal form.

Until these conditions are met, foundation models are best understood as tools for
rapid prototyping and scientific exploration rather than replacements for the certified
high-fidelity solvers required in safety-critical applications.

\subsection{Quantum Algorithms: Structural Advantages and Practical Constraints}
\label{subsec:quantum}

Quantum algorithms for linear systems and differential equations offer striking
theoretical complexity improvements. In particular, the Harrow--Hassidim--Lloyd (HHL)
algorithm achieves exponential speedup for certain classes of linear systems under
specific assumptions \cite{childs2021high,jin2023quantum,liu2021efficient,krovi2023improved}.
Translating such asymptotic advantages into practical tools for PDE solution, however, requires confronting constraints rooted less in hardware immaturity than in the mathematical structure of current quantum algorithms.

The HHL speedup applies only under restrictive conditions
\cite{childs2021high,jin2023quantum,liu2021efficient,krovi2023improved}, including
efficient quantum state preparation, access to quantum random access memory, and
well-conditioned sparse systems. These requirements are rarely satisfied
simultaneously by PDE discretizations arising in practice. In particular, state
preparation from classically computed right-hand sides can reintroduce much of the
classical computational cost that the algorithm is meant to avoid. Without efficient
encoding, the speedup applies only when the problem data are already available in
quantum form, a prerequisite that does not hold generically
\cite{jin2023quantum,harrow2009quantum,duan2020survey,jin2022time}. These are not
merely engineering bottlenecks that will disappear with larger devices; they reflect
a deeper mismatch between classical PDE data structures and current quantum
computational models.

Variational quantum algorithms relax some of these assumptions by tolerating
approximate solutions and operating on near-term hardware, but they introduce a
different set of difficulties. Most notably, the barren plateau phenomenon causes
gradient magnitudes to decay exponentially with system size, severely limiting the
depth and expressivity of circuits that can be trained reliably by gradient-based
optimization \cite{mcclean2018barren,cerezo2021variational,lubasch2020variational,larocca2025barren,cerezo2021cost}.
For this reason, hybrid classical--quantum workflows remain the most credible
near-term pathway: classical methods handle discretization, meshing, and solution
assembly, while quantum subroutines are used selectively to accelerate specific
bottlenecks such as eigenvalue estimation, Monte Carlo sampling, or high-dimensional
optimization \cite{cranganore2024paving,ye2024hybrid,jaffali2024h,mazzola2024quantum}.

Achieving quantum advantage for PDE solution in practically relevant regimes will
require both fault-tolerant hardware at scales not yet realized and algorithms that
avoid the restrictive assumptions of current proposals. Quantum computing remains a
scientifically important long-term investment, but near-term PDE research is not
materially limited by the absence of quantum resources.

\subsection{Differentiable Programming and the Inverse Design}
\label{subsec:differentiable_programming}

Differentiable programming---the capacity to propagate gradients through entire
simulation pipelines via automatic differentiation---represents a genuine near-term
capability shift whose impact on inverse problems and optimal design is already
measurable \cite{innes2019differentiable,farrell2013automated,moyner2025jutuldarcy,liang2019differentiable}.
By making the forward simulation a differentiable function of its inputs, this paradigm
enables gradient-based optimization over parameter spaces that were previously accessible
only through finite differences, adjoint derivations, or surrogate-based approximations.

The scientific impact is clearest in settings where the governing equations are well
established but the inputs---initial conditions, material parameters, boundary forcing,
geometric configuration---must be inferred from indirect observations \cite{innes2019differentiable,liang2019differentiable}.
Differentiable physics formulations in this context allow the inverse problem to be
posed as an unconstrained or constrained optimization over the input space, with
gradients computed at the cost of a small multiple of the forward solve \cite{holl2020learning,karnakov2024solving,de2018end}.
This capability had classical analogs in adjoint methods (Section~\ref{sec:hybrid_ml}),
but differentiable programming frameworks generalize adjoint computation to arbitrary
computational graphs without problem-specific derivations, substantially reducing the
expert burden of applying optimal control and parameter estimation to new problem classes.

Several structural challenges constrain reliability. Inverse PDE problems are inherently ill-posed: multiple input configurations may be
consistent with available observations, and gradient-based optimization over non-convex
loss landscapes may converge to physically spurious local minima \cite{ye2019optimization,kabanikhin2008definitions}.
Classical regularization frameworks (Tikhonov, Bayesian inference) address ill-posedness
with principled uncertainty quantification \cite{calvetti2018inverse,mohammad2021regularization,}; differentiable physics approaches require
analogous regularization design, and the implicit regularization introduced by neural
network parameterization or early stopping lacks the theoretical interpretability of
classical approaches \cite{habring2024neural,nguyen2024tnet}.
Multi-fidelity strategies that combine inexpensive differentiable surrogate models for
exploration with high-fidelity evaluations for refinement are well-positioned to address
the computational burden of large-scale inverse design, and represent an active
integration point between classical numerical methods and learned differentiable models \cite{grbcic2024efficient,li2024multi,deng2023differentiable,behrou2025physics}.

The longer-term potential of differentiable programming extends to \emph{scientific
discovery}: using differentiable PDE simulations as components in optimization problems
that identify governing equations or constitutive laws from experimental data.
Here, the distinction between computing (solving a known equation) and discovery
(identifying an unknown equation) introduced in Section~\ref{subsec:open_problems}
becomes operationally important.
Conflating the two objectives can compromise both: a model optimized simultaneously as
a solver and as an equation identifier may do neither task well.
Architectures and validation protocols that explicitly separate these objectives are
a productive direction.

\subsection{Exascale Architectures and Algorithmic Co-Design}
\label{subsec:exascale}

The realization of exascale computing systems has shifted the algorithmic bottleneck
from peak floating-point throughput to data movement, communication latency, and
load balance \cite{unat2025persistent,markidis2015solving}.
Classical PDE solvers have been engineered over decades for strong and weak scaling
on distributed memory architectures, with domain decomposition, multigrid hierarchies,
and communication-aware data structures providing the algorithmic foundation
(Section~\ref{subsec:crosscutting}).
Achieving efficient utilization of exascale resources requires extending this
engineering to heterogeneous node architectures---combining multicore CPUs with GPU
accelerators and high-bandwidth memory---while managing the increased probability of
silent data corruption that accompanies the scale of component counts \cite{snir2014addressing,atchley2023frontier}.

Neural methods exhibit structurally different scaling characteristics.
Inference over batched inputs parallelizes with minimal communication, making neural
surrogates natural candidates for embarrassingly parallel many-query workflows at
exascale \cite{yin2022strategies,ward2021colmena}.
Training, however, introduces gradient synchronization requirements that create
communication bottlenecks at extreme parallelism, and model parallelism introduces
pipeline overhead that classical solvers do not face \cite{romero2022accelerating,singh2025optimizing}.
The complementarity between classical and neural scaling profiles suggests that
exascale workflows will increasingly be \emph{heterogeneous by design}: classical
solvers providing high-fidelity single-query results, neural operators providing
rapid parametric sweeps, and hybrid methods combining both in adaptive workflows
that allocate resources according to per-query accuracy requirements \cite{tanneru2026exascale,govett2024exascale,ward2025employing}.

Algorithmic resilience is a dimension of exascale computing that receives less
attention than performance but is equally consequential for scientific reliability \cite{snir2014addressing,asch2018big,agullo2022resiliency}.
Classical iterative solvers exhibit natural resilience to isolated hardware faults:
a corrupted communication round degrades convergence rate rather than producing
silently incorrect results, and residual monitoring provides a runtime reliability
indicator.
Neural methods' resilience characteristics are substantially less understood: how
distributed weight corruption affects output quality, whether physics-based training
provides implicit error resistance, and how to design runtime monitors for neural
surrogate reliability are open questions with direct implications for long-running
exascale simulations.

The deepest challenge at exascale is not hardware utilization but algorithmic
co-design: ensuring that the mathematical structure of PDE discretizations, solvers,
and learned components is matched to the communication and memory access patterns
of the hardware, rather than treating the hardware as a generic platform on which
existing algorithms are accelerated \cite{asch2018big,sarkar2009software,anzt2020preparing,bergman2008exascale}.
This co-design philosophy, well established in classical high-performance scientific
computing, must be extended to encompass the training and inference pipelines of
neural PDE components if hybrid methods are to achieve their potential on future
architectures.

\section*{Concluding Remarks}

This review has examined computational methods for partial differential equations across two mature paradigms, classical numerical analysis and machine learning, through a unified evaluative framework grounded in six fundamental challenges. The central finding is not that one paradigm supersedes the other, but that the two are genuinely complementary in a precise sense: each addresses difficulties that are fundamental to the other, rooted in foundational assumptions rather than mere engineering limitations. Classical methods are deductive: their errors can be bounded by quantities derived from PDE structure and discretization parameters, independently of any training history. That deductive character underlies both their greatest strength—certified convergence and structure preservation—and their most persistent limitations, including severe scaling with dimension and labor-intensive geometric preprocessing. Machine learning methods, by contrast, are inductive: their accuracy depends on statistical proximity to the training distribution. This inductive character explains both their adaptability to high-dimensional and parametric problems and their inability to provide deductive guarantees for deployment instances outside that distribution. Recognizing this asymmetry is not defeatist; it is the necessary starting point for designing hybrid systems that are stronger than either paradigm alone.

The evaluation of classical methods reveals solvers of remarkable reliability and mathematical depth. These are not legacy technologies awaiting displacement; they are the certifying infrastructure that any alternative methodology must either match or genuinely surpass in applications where certification matters. The evaluation of machine learning methods, in turn, shows that the field has produced capabilities inaccessible to classical methods—resolution-invariant operator learning, mesh-free approximation in high-dimensional parameter spaces, and amortized inference over parametric families—while also exposing persistent limitations whose tractability differs in kind. Physical inconsistency and training instability appear at least partially addressable through architectural and algorithmic innovation. By contrast, persistent accuracy limitations and the absence of firm theoretical foundations appear more fundamental, reflecting structural properties of neural approximation and stochastic optimization that are unlikely to disappear through scaling alone.

The critical synthesis presented here argues that the most productive scientific trajectory lies in principled hybridization rather than paradigm competition. Three developments are necessary for this trajectory to mature. First, the \emph{structure inheritance problem}—determining when classical guarantees propagate through hybrid couplings to certify properties of the combined system—must be resolved for the most important problem classes, including constitutive learning within variational frameworks, differentiable physics with classical solvers, and neural closure models in conservation-law systems. Second, the \emph{error budget framework} should become standard practice in hybrid method design, enforcing a sharp decomposition of discretization, neural approximation, and coupling errors so that computational resources are allocated against identified bottlenecks rather than tuned empirically. Third, \emph{verification and validation protocols} for hybrid systems must attain the same institutional status that manufactured-solution convergence testing holds in classical numerical analysis—not optional good practice, but a prerequisite for credible claims of reliability.

The emerging frontiers discussed in the preceding section—foundation models, differentiable programming, quantum algorithms, and exascale co-design—represent genuine opportunities, but their realization depends on progress at the foundational level. Foundation models for PDEs will become scientifically credible when the conditions for certified transfer can be stated and verified, not merely when benchmark performance is encouraging. Differentiable programming is already transforming inverse-design workflows; its full potential depends on regularization frameworks that render the implicit structure of neural parameterizations as interpretable as classical Tikhonov or Bayesian formulations. Quantum advantage for PDE solution remains structurally constrained by state-preparation requirements that are mathematical, not merely hardware, limitations and will not be resolved by qubit scaling alone.

The grand challenges motivating this analysis are defined precisely by the concurrent presence of the six difficulties that opened this review: high dimensionality, nonlinearity, geometric complexity, discontinuities, disparate scales, and multiphysics coupling. No current method, classical or neural, addresses all of these simultaneously at the levels of accuracy and certification that demanding applications require. The honest conclusion, therefore, is not that the problem is solved, but that the field now possesses two complementary frameworks whose principled integration offers a credible path forward. Realizing that path will require sustained research into mathematical foundations that neither paradigm has yet established for their combination. Recognition of this foundational work, although less visible than benchmark improvements, is precisely what will determine whether the present era of computational research yields enduring scientific infrastructure or merely a collection of impressive demonstrations.

\subsection*{Statements \& Declarations}
\subsubsection*{Acknowledgments}
This work was supported by the Multiphysics of Salt Precipitation
During CO$_2$ Storage (Saltation) project, grant nr. 357784,
funded by the Research Council of Norway. JWB acknowledges support from the FRIPRO project “Unlocking maximal geological CO$_2$ storage through experimentally validated mathematical modeling of dissolution and convective mixing (TIME4CO2)”, grant nr. 355188, funded by the Research Council of Norway.  

\subsubsection*{Competing Interests}
The authors declare no competing interests.


\subsubsection*{Data Availability Statement}
No new data were generated or analysed in this study.

\printbibliography

\clearpage
\begin{landscape}
\small
\begin{longtable}{p{3.5cm} p{2.5cm} p{2cm} p{1.8cm} p{2.2cm} p{1.8cm} p{7cm}}
\caption{Comprehensive comparison of classical PDE solution methods: computational characteristics, adaptivity capabilities, nonlinearity handling, theoretical guarantees, implementation requirements, and key distinguishing features.} 
\label{tab:classical-pde-methods} \\
\toprule
\textbf{Method} & \textbf{Computational} & \textbf{Mesh} & \textbf{UQ} & \textbf{Nonlinearity} & \textbf{Theory} & \textbf{Key Distinguishing Characteristics} \\
 & \textbf{Cost Class} & \textbf{Adaptivity} & \textbf{Support} & \textbf{Handling} & \textbf{Guarantees} & \\
\midrule
\endfirsthead
\toprule
\textbf{Method} & \textbf{Computational} & \textbf{Mesh} & \textbf{UQ} & \textbf{Nonlinearity} & \textbf{Theory} & \textbf{Key Distinguishing Characteristics} \\
 & \textbf{Cost Class} & \textbf{Adaptivity} & \textbf{Support} & \textbf{Handling} & \textbf{Guarantees} & \\
\midrule
\endhead
\midrule
\multicolumn{7}{r}{\textit{Continued on next page}} \\
\endfoot
\bottomrule
\multicolumn{7}{p{\textwidth}}{\textbf{Notation and Definitions:} \textbf{Computational Cost Class:} \textit{Low} = optimal $O(N)$ or near-optimal complexity; \textit{Medium} = $O(N \log N)$ to $O(N^{1.3})$; \textit{High} = $O(N^{1.3})$ to $O(N^{1.8})$; \textit{Very High} = $O(N^2)$ or worse without acceleration. Cost includes setup; some methods have expensive offline phases with fast online evaluation. \textbf{Theory Guarantees:} \textit{Strong} = rigorous a priori/posteriori error estimates with proven convergence rates; \textit{Partial} = theoretical results for restricted problem classes or asymptotic bounds; \textit{Weak} = heuristic analysis without rigorous proof; \textit{Emerging} = active theoretical development. \textbf{UQ Support:} \textit{No} = requires external Monte Carlo sampling; \textit{Limited} = parametric sensitivity analysis; \textit{Yes} = native uncertainty quantification with error control. \textbf{Adaptivity:} Types include h-refinement (mesh geometry), p-refinement (polynomial degree), hp-refinement (combined), r-refinement (node relocation), AMR (adaptive mesh refinement with hierarchy). Performance characteristics assume appropriate problem regularity; singularities, discontinuities, or insufficient smoothness degrade all methods. Software: FEniCS/FEniCSx, deal.II, DUNE, MFEM (FEM); OpenFOAM, CLAWPACK (FVM); PETSc, Trilinos (general frameworks); NekRS, Dedalus (spectral); BoomerAMG, HYPRE (AMG).} \\
\endlastfoot

\multicolumn{7}{l}{\textbf{\large FINITE DIFFERENCE METHODS}} \\
\midrule
Standard FDM & Low & None & No & Explicit time & Partial & Optimal memory efficiency; structured grids only; von Neumann stability analysis; CFL restrictions $\Delta t \sim h$ (hyperbolic), $h^2$ (parabolic); 2nd-order typical \\
\addlinespace
Compact FD & Low-Medium & None & No & Implicit stencils & Strong & Padé-type schemes; tridiagonal systems; spectral-like resolution; 4th-6th order standard; excellent for DNS/LES \\
\addlinespace
High-Order Compact & Medium & None & No & Implicit stencils & Strong & 8th-10th order achievable; requires high solution regularity; recent advances in dispersion-relation-preserving schemes \\
\addlinespace
WENO & Medium & r-adapt & No & Conservative & Strong & Nonlinear adaptive stencils; TVD/ENO properties; robust shock capturing; 5th order standard; improved variants address critical points \\
\addlinespace
Unstructured WENO & Medium-High & r-adapt & No & Conservative & Strong & Complex stencil reconstruction; recent algorithmic improvements yield 40-80\% speedup; extended geometric capability \\

\midrule
\multicolumn{7}{l}{\textbf{\large FINITE ELEMENT METHODS}} \\
\midrule
h-FEM (linear) & Medium & h-refine & Limited & Newton-Raphson & Strong & $O(h^2)$ convergence; mature software ecosystem (FEniCS, deal.II); geometric flexibility; widespread adoption \\
\addlinespace
h-FEM (quadratic) & Medium & h-refine & Limited & Newton-Raphson & Strong & $O(h^3)$ convergence; higher assembly cost; better accuracy per DOF; standard for many applications \\
\addlinespace
p-FEM & Medium & p-refine & Limited & Newton-Raphson & Strong & Exponential convergence for smooth solutions; hierarchical basis; conditioning degrades at high polynomial degree \\
\addlinespace
hp-FEM & Low-Medium & hp-adapt & Limited & Newton-Raphson & Strong & Optimal complexity proven; exponential convergence even with singularities; automatic strategy selection; state-of-art adaptivity \\
\addlinespace
Mixed FEM & Medium-High & h-adapt & No & Saddle point & Strong & Simultaneous approximation of primal/dual variables; inf-sup stability required; Taylor-Hood, Raviart-Thomas elements; essential for incompressible flow \\
\addlinespace
Discontinuous Galerkin & Medium-High & hp-adapt & No & Explicit/Implicit & Strong & Local conservation; numerical flux coupling; hp-adaptivity without hanging nodes; excellent parallel scaling; higher memory than continuous FEM \\
\addlinespace
Asynchronous DG & Medium-High & hp-adapt & No & Asynchrony-tolerant & Strong & Eliminates global synchronization barriers; proven stability bounds; designed for massively parallel architectures \\

\midrule
\multicolumn{7}{l}{\textbf{\large SPECTRAL AND HIGH-ORDER METHODS}} \\
\midrule
Fourier Spectral & Low & None & No & Pseudospectral & Strong & FFT-based $O(N \log N)$; exponential convergence for smooth periodic problems; Gibbs phenomenon for discontinuities \\
\addlinespace
Chebyshev Spectral & Medium & None & No & Collocation & Strong & Gauss-Lobatto points; boundary clustering; non-periodic domains; dense matrices unless preconditioned \\
\addlinespace
Legendre Spectral & Medium & None & No & Galerkin & Strong & More uniform accuracy distribution than Chebyshev; Gauss-Lobatto integration; better conditioning \\
\addlinespace
Spectral Element (SEM) & Medium & p-adapt & No & Newton-Raphson & Strong & $C^0$ inter-element continuity; GLL quadrature; tensor-product structure enables matrix-free; geometric flexibility via decomposition \\
\addlinespace
GPU-Accelerated SEM & Medium & p-adapt & No & Newton-Raphson & Strong & Exascale CFD capabilities; GPU-native implementation; overset grids; high-order time splitting; production-ready for complex flows \\
\addlinespace
Nonlocal SEM & Medium-High & p-adapt & No & Convolution & Strong & Efficient treatment of nonlinear/nonlocal operators; convolution integrals on complex 2D domains; recent development \\

\midrule
\multicolumn{7}{l}{\textbf{\large FINITE VOLUME METHODS}} \\
\midrule
Cell-Centered FV & Medium & AMR & No & Godunov/MUSCL & Partial & Exact local conservation by construction; Riemann solvers; slope limiters; 1st-3rd order typical; dominant in CFD \\
\addlinespace
Vertex-Centered FV & Medium-High & Unstructured & No & Conservative & Partial & Median dual control volumes; FEM-compatible; complex flux evaluation; larger stencils than cell-centered \\
\addlinespace
High-Order FV & Medium-High & AMR & No & WENO/DG-like & Strong & 4th-5th order via reconstruction; maintains conservation; DG-level accuracy with FV conservation structure \\

\midrule
\multicolumn{7}{l}{\textbf{\large ADAPTIVE AND MULTIGRID METHODS}} \\
\midrule
h-Adaptive Refinement & Low-Medium & Dynamic h & No & Base method & Partial & Gradient/residual error indicators; dynamic data structures; 2-10× efficiency vs uniform refinement; automatic feature resolution \\
\addlinespace
Block-Structured AMR & Low-Medium & Hierarchical & No & Conservative & Partial & Berger-Colella framework; refluxing for conservation; temporal subcycling; vectorization-friendly; 10-100× speedup for localized features \\
\addlinespace
Geometric Multigrid & Low & Hierarchy & No & V/W/F cycles & Strong & Grid-independent convergence; $O(N)$ optimal complexity; requires geometric hierarchy; robust for elliptic problems \\
\addlinespace
Algebraic Multigrid & Low-Medium & Graph-based & No & Galerkin & Strong & No geometric information needed; strength-of-connection coarsening; essential for unstructured problems; BoomerAMG, HYPRE \\
\addlinespace
Compatible Relaxation AMG & Low-Medium & Graph-based & No & Modified coarsening & Strong & Improved compatibility with standard smoothers; reduced communication overhead; grammar-guided cycle optimization \\

\midrule
\multicolumn{7}{l}{\textbf{\large MESHLESS AND PARTICLE METHODS}} \\
\midrule
Method of Fundamental Solutions & High (dense) & None & No & Direct solve & Strong & Boundary collocation; exponential accuracy; requires fundamental solution; severe conditioning; FMM acceleration to $O(N \log N)$ \\
 & Low (FMM) & & & & & \\
\addlinespace
Global RBF & Very High & None & No & Global interpolation & Weak & Spectral accuracy possible; severe ill-conditioning limits to $N < 10^4$; shape parameter selection critical \\
\addlinespace
Local RBF (RBF-FD) & Medium & Nodal & No & Local interpolation & Partial & Compact support; sparse matrices; alleviates conditioning; better parallelization than global RBF \\
\addlinespace
Meshless Local Petrov-Galerkin & High & Nodal & No & Moving Least Squares & Partial & Local weak forms; MLS approximation; integration difficulties; free of mesh constraints; large deformation applications \\
\addlinespace
Smoothed Particle Hydrodynamics & Medium (tree) & Lagrangian & No & Explicit particle & Weak & Free surfaces; large deformation; consistency challenges; artificial viscosity for stability; tree algorithms essential \\
 & Very High (naive) & & & & & \\

\midrule
\multicolumn{7}{l}{\textbf{\large BOUNDARY AND INTEGRAL METHODS}} \\
\midrule
Boundary Element Method & High-Very High & Surface only & No & Integral equation & Strong & Dimensionality reduction; infinite domains natural; dense matrices; singular integral evaluation; linear problems only \\
\addlinespace
Fast Multipole BEM & Low-Medium & Surface only & No & Hierarchical FMM & Strong & Tree algorithms; multipole expansions; millions of unknowns feasible; complex but enables large-scale boundary problems \\
\addlinespace
Hierarchical Matrices BEM & Low-Medium & Surface only & No & Low-rank compression & Strong & $\mathcal{H}$-matrix compression; reduced storage; fast solvers; black-box acceleration for BEM \\

\midrule
\multicolumn{7}{l}{\textbf{\large SPECIALIZED DISCRETIZATION METHODS}} \\
\midrule
Isogeometric Analysis & Medium & k-refinement & No & Newton-Raphson & Strong & NURBS/T-splines from CAD; exact geometry; higher continuity ($C^1$, $C^2$); eliminates geometry approximation; shape optimization \\
\addlinespace
Extended FEM & Medium-High & Enrichment & No & Level sets/Heaviside & Partial & Partition of unity enrichment; discontinuities without remeshing; crack propagation; conditioning challenges at enriched nodes \\
\addlinespace
Virtual Element Method & Medium & h-adapt & No & Newton-Raphson & Strong & General polygonal/polyhedral elements; virtual shape functions; mimetic properties; recent development with growing adoption \\
\addlinespace
Mimetic Finite Difference & Medium & Unstructured & No & Conservative & Strong & Discrete calculus theorems; conservation by construction; natural treatment of complex geometries \\

\midrule
\multicolumn{7}{l}{\textbf{\large STRUCTURE-PRESERVING METHODS}} \\
\midrule
Energy Quadratization & Medium & Coarse meshes & No & Variational & Strong & Unconditional free energy dissipation; thermodynamic consistency; phase-field models; recent theoretical advances \\
\addlinespace
Symplectic Integrators & Low-Medium & Spectral basis & No & Hamiltonian & Strong & Long-time energy conservation; multi-symplectic formulations; wave equations; Fourier pseudospectral coupling \\
\addlinespace
SBP-Relaxation Runge-Kutta & Medium & Structured grids & No & Conservative entropy & Strong & Summation-by-parts framework; provable energy/entropy bounds; arbitrary high order; nonlinear wave stability \\
\addlinespace
Geometric Integrators & Medium & Problem-dependent & No & Structure preservation & Strong & Preserve Lie group structure; manifold constraints; long-time accuracy for geometric PDEs \\

\midrule
\multicolumn{7}{l}{\textbf{\large REDUCED-ORDER AND MULTISCALE METHODS}} \\
\midrule
Proper Orthogonal Decomposition & Low (online) & Parameter space & Yes & POD projection & Strong & Snapshot-based optimal basis; fast online evaluation; expensive offline; limited extrapolation; linear/mildly nonlinear \\
 & Very High (offline) & & & & & \\
\addlinespace
Reduced Basis Method & Low (online) & Parameter space & Yes & Greedy selection & Strong & Certified a posteriori error bounds; affine parameter decomposition; offline-online separation; parametric PDEs \\
 & Very High (offline) & & & & & \\
\addlinespace
Multiscale FEM and FV & Medium-High & Coarse grid & No & Implicit upscaling & Strong & Offline local basis construction; scale separation assumption $\epsilon \ll 1$; periodic/random microstructure \\
\addlinespace
Generalized Multiscale FEM & High & Coarse grid & No & Spectral basis & Strong & Multiple basis functions per coarse node; local eigenvalue problems; oversampling; adaptive enrichment strategies \\
\addlinespace
Heterogeneous Multiscale & High & Macro-micro & Limited & Constrained coupling & Strong & General multiscale framework; macro-micro coupling; micro solver can be black-box; dimensional scaling challenges \\
\addlinespace
Localized Orthogonal Decomposition & Medium & Local corrector & No & Corrector problems & Strong & Exponential decay of correctors; quasi-local; optimal error estimates; robust to high contrast \\
\addlinespace
Variational Multiscale & Medium & Coarse grid & No & Stabilized residual & Strong & Fine-scale Green's function modeling; residual-based stabilization; SUPG/PSPG; turbulence subgrid modeling \\
\addlinespace
Equation-Free Methods & High & Micro simulator & No & Projective integration & Emerging & No explicit macroscopic equations needed; coarse projective integration; legacy-code coupling; active theoretical development \\

\end{longtable}
\end{landscape}

\appendix
\renewcommand{\thesection}{S.\arabic{section}}
\renewcommand{\thesubsection}{S.\arabic{section}.\arabic{subsection}}
\renewcommand{\thetable}{S.\arabic{table}}
\renewcommand{\theequation}{S.\arabic{equation}}
\renewcommand{\thefigure}{S.\arabic{figure}}
\setcounter{section}{0}
\setcounter{table}{0}
\setcounter{equation}{0}
\setcounter{figure}{0}

\newpage
\part*{Supplementary Material}
\addcontentsline{toc}{part}{Supplementary Material}
\label{sec:supp}

\section{Extended Neural Architecture Details}
\label{sec:supp_architectures}

This section provides detailed formulations and extended discussion for neural architectures summarized in the main text.


\subsection{Neural Operator Variants}
\label{sec:supp_neural_operators}

\subsubsection{Theoretical Foundations}

Universal approximation results establish that neural operators can approximate continuous operators to arbitrary accuracy given sufficient capacity \cite{kovachki2023neural}. Approximation rates for FNO achieve $O(N^{-s/d})$ convergence for functions with Sobolev regularity $s$ in dimension $d$, comparable to classical spectral methods under smoothness assumptions.

Resolution invariance---the ability to evaluate trained models at resolutions different from training---stems from learning in function space rather than on fixed grids. However, accuracy typically degrades for resolutions significantly finer or coarser than training data, and theoretical guarantees require regularity assumptions often violated in practice.

The universal approximation theorem for operators \cite{chen1995universal} provides the foundation for DeepONet: any continuous operator can be approximated by:
\begin{equation}
\mathcal{G}[u](y) \approx \sum_{k=1}^p b_k(u(x_1), \ldots, u(x_m)) \cdot t_k(y)
\label{eq:supp_universal_operator}
\end{equation}
where $\{x_1, \ldots, x_m\}$ are sensor locations and $p$ is the approximation rank.

\subsubsection{Wavelet Neural Operator}

The Wavelet Neural Operator replaces Fourier transforms with wavelet decomposition for natural multi-resolution representation \cite{tripura2023wavelet}:
\begin{equation}
(Kv)(x) = \mathcal{W}^{-1}\left(R_\phi \cdot \mathcal{W}(v)\right)(x) + W_\phi v(x)
\label{eq:supp_wavelet}
\end{equation}
where $\mathcal{W}$ denotes the wavelet transform. This yields near-linear complexity (often reported as $O(N)$--$O(N\log N)$ depending on implementation) while capturing both local and global features through the wavelet's inherent scale separation. Wavelet bases provide compact support, avoiding Gibbs phenomena that plague Fourier methods near discontinuities.

\subsubsection{U-FNO}

U-FNO combines U-Net's hierarchical encoder-decoder structure with Fourier layers \cite{wen2022accelerating}:
\begin{equation}
v^{(l+1)} = \sigma\left(\text{FourierLayer}(v^{(l)}) + \text{Skip}(v^{(L-l)})\right)
\label{eq:supp_ufno}
\end{equation}
The encoder progressively coarsens resolution while increasing channel depth; the decoder reconstructs fine-scale features using skip connections from corresponding encoder levels. This architecture improves accuracy for problems with scale separation, achieving lower errors than standard FNO on turbulent flow benchmarks.

\subsubsection{Geometry-Informed Neural Operator}

Geometry-Informed Neural Operators (GINO) incorporate geometric embeddings for complex domains \cite{li2023geometry}:
\begin{equation}
\mathcal{G}_\theta[u, \Omega](y) = \sum_{k=1}^p b_k(u, \phi(\Omega)) \cdot t_k(y, \psi(y, \partial\Omega))
\label{eq:supp_gino}
\end{equation}
where $\phi(\Omega)$ encodes domain geometry and $\psi(y, \partial\Omega)$ provides boundary-aware positional encoding. This enables generalization across domain shapes without retraining.

\subsubsection{Latent Neural Operator}

Latent Neural Operators operate in compressed latent space to reduce memory requirements \cite{wang2024exploring}:
\begin{equation}
\mathcal{G}_\theta = \mathcal{D} \circ \mathcal{G}_{\text{latent}} \circ \mathcal{E}
\label{eq:supp_latent}
\end{equation}
where $\mathcal{E}$ encodes to latent space, $\mathcal{G}_{\text{latent}}$ operates in reduced dimensions, and $\mathcal{D}$ decodes to physical space. This enables handling high-resolution 3D problems that exceed GPU memory with standard architectures.

\subsubsection{Foundation Models}

Foundation models pretrained on diverse PDE families represent an emerging direction \cite{herde2024poseidon,sun2025towards}. These large-scale neural operators are trained on datasets spanning multiple equation types:
\begin{itemize}
    \item Elliptic (Poisson, Helmholtz)
    \item Parabolic (heat, diffusion)
    \item Hyperbolic (wave, advection)
    \item Mixed (Navier-Stokes, reaction-diffusion)
\end{itemize}

Preliminary results suggest zero-shot generalization to novel PDEs and few-shot adaptation to new equation classes. However, comprehensive validation remains incomplete, and performance on industrial-scale problems with complex geometries and boundary conditions requires further investigation.


\subsection{Graph Neural Network Architectures}
\label{sec:supp_gnns}

\subsubsection{Graph Network-based Simulators}

Graph Network-based Simulators (GNS) introduced the encode-process-decode framework for learned physical simulation \cite{sanchez2020learning}:

\textbf{Encoder:} Maps particle/node states to latent features
\begin{equation}
z_i^{(0)} = \text{MLP}_{\text{enc}}(s_i)
\label{eq:supp_gns_encoder}
\end{equation}

\textbf{Processor:} Applies $M$ message-passing steps
\begin{equation}
z_i^{(m+1)} = z_i^{(m)} + \text{MLP}_{\text{node}}\left(z_i^{(m)}, \sum_{j \in \mathcal{N}(i)} \text{MLP}_{\text{edge}}(z_i^{(m)}, z_j^{(m)}, r_{ij})\right)
\label{eq:supp_gns_processor}
\end{equation}
where $r_{ij}$ encodes relative positions.

\textbf{Decoder:} Predicts accelerations or state updates
\begin{equation}
\ddot{x}_i = \text{MLP}_{\text{dec}}(z_i^{(M)})
\label{eq:supp_gns_decoder}
\end{equation}

This architecture demonstrated generalization across particle counts and domain configurations for fluid and granular systems.

\subsubsection{MeshGraphNets Extended}

MeshGraphNets handle mesh-based simulations with two edge types \cite{pfaff2020learning}:
\begin{itemize}
    \item \textbf{Mesh edges:} Connect adjacent mesh nodes (encode local connectivity)
    \item \textbf{World edges:} Connect spatially proximate nodes (encode collision/interaction)
\end{itemize}

The processor applies message-passing on both edge types:
\begin{align}
e_{ij}^{(m+1)} &= e_{ij}^{(m)} + \text{MLP}_e(e_{ij}^{(m)}, z_i^{(m)}, z_j^{(m)}) \label{eq:supp_mgn_edge}\\
z_i^{(m+1)} &= z_i^{(m)} + \text{MLP}_v\left(z_i^{(m)}, \sum_{j} e_{ij}^{(m+1)}\right) \label{eq:supp_mgn_node}
\end{align}

This dual-edge structure enables handling of contact mechanics and fluid-structure interaction.

\subsubsection{MP-PDE and Temporal Bundling}

MP-PDE introduces temporal bundling to reduce autoregressive error accumulation \cite{brandstetter2022message}:
\begin{equation}
(u_{t+1}, u_{t+2}, \ldots, u_{t+K}) = f_\theta(u_{t-H+1}, \ldots, u_t)
\label{eq:supp_mppde}
\end{equation}

Predicting multiple future timesteps simultaneously:
\begin{itemize}
    \item Reduces number of autoregressive steps by factor $K$
    \item Improves long-rollout stability
    \item Enables parallel computation of future states
\end{itemize}

The pushforward trick during training uses model predictions rather than ground truth for input, exposing the model to its own error distribution.

\subsubsection{Equivariant Architectures}

Geometric Deep Learning provides frameworks for symmetry-preserving GNNs \cite{bronstein2021geometric}. E(n)-equivariant GNNs maintain invariance to rotations, translations, and reflections \cite{satorras2021n}:
\begin{align}
m_{ij} &= \phi_e(h_i, h_j, \|x_i - x_j\|^2, a_{ij}) \label{eq:supp_egnn_msg}\\
x_i' &= x_i + C \sum_{j \neq i} (x_i - x_j) \phi_x(m_{ij}) \label{eq:supp_egnn_coord}\\
h_i' &= \phi_h(h_i, \sum_{j \neq i} m_{ij}) \label{eq:supp_egnn_feat}
\end{align}

Key properties:
\begin{itemize}
    \item Coordinate updates (\ref{eq:supp_egnn_coord}) use only relative positions
    \item Message function depends on distance, not direction
    \item Features transform as scalars under rotations
\end{itemize}

This ensures predictions transform correctly under coordinate changes---essential for physical systems where laws are coordinate-independent.

\subsubsection{Multiscale Graph Networks}

MultiScale MeshGraphNets construct graph hierarchies via pooling \cite{fortunato2022multiscale}:
\begin{equation}
G^{(l+1)} = \text{Pool}(G^{(l)}), \quad G^{(l)} = \text{Unpool}(G^{(l+1)})
\label{eq:supp_multiscale_gnn}
\end{equation}

The hierarchical structure enables:
\begin{itemize}
    \item Efficient information propagation across scales
    \item $O(\log N)$ effective receptive field growth
    \item Natural handling of multiscale physics
\end{itemize}

This mirrors classical multigrid principles within a learned framework, with coarse levels capturing global behavior and fine levels resolving local details.


\subsection{Transformer Architectures for PDEs}
\label{sec:supp_transformers}

\subsubsection{Galerkin Transformer}

The Galerkin Transformer interprets attention as learning adaptive test and trial spaces within a Petrov-Galerkin framework \cite{cao2021choose}. Replacing softmax with linear attention:
\begin{equation}
\text{Attention}(Q, K, V) = Q(K^T V) / N
\label{eq:supp_galerkin}
\end{equation}
reduces complexity to $O(N)$ while preserving global expressivity. The linear formulation can be interpreted as:
\begin{equation}
u(x) \approx \sum_{i} \phi_i(x) c_i, \quad c_i = \langle \psi_i, f \rangle
\label{eq:supp_galerkin_basis}
\end{equation}
where $\phi_i$ (from $V$) are trial functions and $\psi_i$ (from $K$) are test functions, both learned.

\subsubsection{Fourier Transformer}

Fourier Transformers perform attention in spectral space:
\begin{equation}
\text{FourierAttention}(X) = \mathcal{F}^{-1}\left(\text{Attention}(\mathcal{F}Q, \mathcal{F}K, \mathcal{F}V)\right)
\label{eq:supp_fourier_attn}
\end{equation}

Truncating to $k_{\max}$ modes reduces effective sequence length while:
\begin{itemize}
    \item Preserving global spectral information
    \item Enabling adaptive weighting in frequency domain
    \item Naturally handling periodic boundaries
\end{itemize}

\subsubsection{Factorized and Axial Transformers}

High-dimensional attention is made tractable by factorizing interactions across spatial axes \cite{ho2019axial}:
\begin{equation}
A \approx A_x \otimes A_y \otimes A_z
\label{eq:supp_factorized}
\end{equation}

For 3D problems with $n$ points per dimension (total $N = n^3$ points):
\begin{itemize}
    \item Full attention: $O(N^2) = O(n^6)$ complexity
    \item Factorized attention: $O(3n^4)$ complexity (i.e., $O(N^{4/3})$)
    \item Further reduction with sparse patterns: $O(N \log N)$
\end{itemize}

Axial attention applies this sequentially:
\begin{equation}
X' = \text{Attn}_z(\text{Attn}_y(\text{Attn}_x(X)))
\label{eq:supp_axial}
\end{equation}

\subsubsection{Operator Transformer}

Operator Transformers extend encoder-decoder attention to function-to-function mappings \cite{li2022uniformer}:
\begin{equation}
\mathcal{G}[u](y) = \text{Decoder}(y, \text{Encoder}(u(x_1), \ldots, u(x_m)))
\label{eq:supp_operator_transformer}
\end{equation}

Cross-attention in the decoder queries outputs at arbitrary coordinates:
\begin{equation}
\hat{u}(y) = \sum_{i} \alpha(y, x_i) \cdot v(u(x_i))
\label{eq:supp_cross_attention}
\end{equation}
enabling mesh-free evaluation with learned positional encodings.

\subsubsection{Physics-Informed Transformers}

Physical structure is enforced through architectural constraints \cite{lorsung2024picl}:
\begin{itemize}
    \item \textbf{Causal masking:} For time-dependent problems, attention is masked to prevent future information leakage:
    \begin{equation}
    A_{ij} = 0 \quad \text{if } t_j > t_i
    \label{eq:supp_causal_mask}
    \end{equation}
    
    \item \textbf{Symmetry-preserving attention:} Attention weights respect physical symmetries through equivariant query/key projections.
    
    \item \textbf{Conservation-aware projections:} Output projections enforce divergence-free (incompressibility) or curl-free (irrotationality) constraints.
\end{itemize}

\subsubsection{Efficient Attention Variants}

\paragraph{Linear Attention.}
Replaces softmax with kernelized inner products \cite{katharopoulos2020transformers}:
\begin{equation}
\text{Attention}(Q, K, V) = \phi(Q)(\phi(K)^T V)
\label{eq:supp_linear_attn}
\end{equation}
achieving $O(N)$ complexity. The kernel $\phi$ determines the implicit attention pattern; common choices include $\phi(x) = \text{elu}(x) + 1$.

\paragraph{Sparse Attention.}
Restricts interactions to local, strided, or learned subsets \cite{child2019generating}:
\begin{equation}
A_{ij} \neq 0 \quad \text{only if } j \in \mathcal{S}(i)
\label{eq:supp_sparse_attn}
\end{equation}
where $\mathcal{S}(i)$ is the sparsity pattern. Combining local windows with global tokens balances locality and long-range context.

\paragraph{Low-Rank Attention.}
Approximates attention matrices with rank-$r$ factorizations \cite{wang2020linformer}:
\begin{equation}
\text{Attention}(Q, K, V) \approx \text{softmax}(Q(E_K K)^T / \sqrt{d}) E_V V
\label{eq:supp_lowrank_attn}
\end{equation}
where $E_K, E_V \in \mathbb{R}^{r \times N}$ project keys and values to lower dimension. Effective for smooth operators where attention matrices have low effective rank.

\paragraph{Flash Attention.}
Computes exact attention with optimized memory access \cite{dao2022flashattention}:
\begin{itemize}
    \item Tiling attention computation to fit in SRAM
    \item Recomputation during backward pass to avoid storing attention matrices
    \item 2-4$\times$ speedup with no approximation error
\end{itemize}

\subsubsection{Positional Encoding Strategies}

Unlike NLP, PDEs require continuous positional representations \cite{tancik2020fourier,shaw2018self}:

\paragraph{Fourier Features.}
\begin{equation}
\gamma(x) = [\sin(2\pi B x), \cos(2\pi B x)]
\label{eq:supp_fourier_features}
\end{equation}
where $B$ contains learnable or random frequencies. Higher frequencies enable learning high-frequency functions but may cause optimization difficulties.

\paragraph{Learned Encodings.}
Coordinate MLPs:
\begin{equation}
\text{PE}(x) = \text{MLP}(x)
\label{eq:supp_learned_pe}
\end{equation}
provide maximum flexibility but require more data.

\paragraph{Relative Position Encodings.}
\begin{equation}
A_{ij} = \text{softmax}(q_i^T k_j + q_i^T r_{ij})
\label{eq:supp_relative_pe}
\end{equation}
where $r_{ij}$ depends on $x_i - x_j$, ensuring translation invariance.


\subsection{Generative Model Details}
\label{sec:supp_generative}

\subsubsection{Bayesian Perspective on PDE Solutions}

From a Bayesian perspective, PDE solving becomes inference \cite{stuart2010inverse}. For the PDE $\mathcal{N}[u; \lambda] = 0$ with uncertain parameters $\lambda$, boundary conditions $g$, and observations $\mathcal{D} = \{(x_i, u_i^{\text{obs}})\}$:
\begin{equation}
p(u \mid \mathcal{D}, \lambda, g) \propto p(\mathcal{D} \mid u) \cdot p(u \mid \lambda, g)
\label{eq:supp_bayesian_pde}
\end{equation}
where $p(\mathcal{D} \mid u)$ encodes observation noise and $p(u \mid \lambda, g)$ represents prior structure (often including PDE constraints).

This framing separates:
\begin{itemize}
    \item \textbf{Aleatoric uncertainty:} Irreducible randomness (stochastic forcing, measurement noise, turbulence variability)
    \item \textbf{Epistemic uncertainty:} Reducible uncertainty from limited data, unknown parameters, or model mismatch
\end{itemize}

\subsubsection{Score-Based and Diffusion Models}

Diffusion models generate solutions by reversing a noise corruption process \cite{song2020score}. The forward SDE perturbs a solution toward Gaussian noise:
\begin{equation}
du = f(u, t) dt + g(t) dW
\label{eq:supp_forward_sde}
\end{equation}

The reverse process requires the score function:
\begin{equation}
du = [f(u, t) - g(t)^2 \nabla_u \log p_t(u)] dt + g(t) d\bar{W}
\label{eq:supp_reverse_sde}
\end{equation}

Training via denoising score matching:
\begin{equation}
\mathcal{L} = \mathbb{E}_{t, u_0, \epsilon}\left[\|s_\theta(u_t, t) - \nabla_{u_t} \log p(u_t \mid u_0)\|^2\right]
\label{eq:supp_score_matching}
\end{equation}

\paragraph{Conditional Diffusion.}
Learns $p(u \mid a)$ for parameters/BCs/ICs $a$ through conditioning mechanisms:
\begin{itemize}
    \item \textbf{Concatenation:} $s_\theta(u_t, t, a)$ with $a$ as additional input
    \item \textbf{Cross-attention:} Condition via attention to encoded $a$
    \item \textbf{FiLM modulation} \cite{perez2018film}: $\gamma(a) \odot h + \beta(a)$
\end{itemize}

\paragraph{Classifier-Free Guidance.}
Trades diversity for sharper conditioning \cite{ho2022classifier}:
\begin{equation}
\tilde{s}_\theta = (1 + w) s_\theta(u_t, t, a) - w \cdot s_\theta(u_t, t, \emptyset)
\label{eq:supp_cfg}
\end{equation}
where $w > 0$ amplifies the conditional signal.

\paragraph{Physics-Informed Diffusion.}
Reduces data requirements by steering sampling toward PDE-feasible fields \cite{bastek2024physics}:
\begin{itemize}
    \item Residual-guided corrections during denoising
    \item Projection steps enforcing boundary conditions
    \item Conservation constraint enforcement
\end{itemize}

\subsubsection{Variational Autoencoders}

VAEs learn low-dimensional probabilistic representations via encoder $q_\phi(z \mid u)$ and decoder $p_\theta(u \mid z)$:
\begin{equation}
\mathcal{L}_{\text{VAE}} = \mathbb{E}_{q_\phi(z \mid u)}[\log p_\theta(u \mid z)] - D_{\text{KL}}(q_\phi(z \mid u) \| p(z))
\label{eq:supp_vae_elbo}
\end{equation}

For PDEs, VAEs offer:
\begin{itemize}
    \item Fast sampling (single decode pass)
    \item Interpretable latent structure often aligning with physical modes
    \item Natural dimensionality reduction for high-dimensional fields
\end{itemize}

\paragraph{Latent Dynamics.}
For time-dependent systems, learn evolution in latent space:
\begin{equation}
z_{t+\Delta t} = f_\psi(z_t, \Delta t)
\label{eq:supp_latent_dynamics}
\end{equation}
enabling efficient long-time integration without decoding at each step.

\paragraph{Conditional VAEs.}
Model $p(u \mid a)$ by conditioning encoder and decoder on parameters \cite{sohn2015learning}:
\begin{equation}
\mathcal{L}_{\text{CVAE}} = \mathbb{E}_{q_\phi(z \mid u, a)}[\log p_\theta(u \mid z, a)] - D_{\text{KL}}(q_\phi(z \mid u, a) \| p(z \mid a))
\label{eq:supp_cvae}
\end{equation}

\subsubsection{Normalizing Flows}

Flows provide exact likelihoods through invertible maps $u = f(z)$ \cite{rezende2015variational}:
\begin{equation}
p_U(u) = p_Z(f^{-1}(u)) \left|\det \frac{\partial f^{-1}}{\partial u}\right|
\label{eq:supp_flow_density}
\end{equation}

\paragraph{Coupling Flows.}
Achieve tractable Jacobians through partitioned transformations \cite{huang2017learnable}:
\begin{align}
u_{1:d} &= z_{1:d} \label{eq:supp_coupling1}\\
u_{d+1:D} &= z_{d+1:D} \odot \exp(s(z_{1:d})) + t(z_{1:d}) \label{eq:supp_coupling2}
\end{align}
The Jacobian is triangular with determinant $\sum_i s_i(z_{1:d})$.

\paragraph{Continuous Normalizing Flows.}
Neural ODEs define flows through continuous dynamics \cite{chen2018neural}:
\begin{equation}
\frac{du}{dt} = f_\theta(u, t), \quad \frac{d \log p}{dt} = -\text{tr}\left(\frac{\partial f}{\partial u}\right)
\label{eq:supp_cnf}
\end{equation}
providing maximum flexibility at higher computational cost.

For PDE fields, flows are typically applied in latent space (autoencoder + flow) to avoid the extreme dimensionality of physical space.

\subsubsection{Bayesian Neural Networks}

BNNs quantify epistemic uncertainty through posterior over weights \cite{neal1996priors}:
\begin{equation}
p(u^* \mid x^*, \mathcal{D}) = \int p(u^* \mid x^*, \theta) p(\theta \mid \mathcal{D}) d\theta
\label{eq:supp_bnn_predictive}
\end{equation}

\paragraph{Approximate Inference Methods.}
\begin{itemize}
    \item \textbf{Variational inference} \cite{blundell2015weight}: Approximate $p(\theta \mid \mathcal{D}) \approx q_\phi(\theta)$
    \item \textbf{MC Dropout} \cite{gal2016dropout}: Dropout at test time as approximate variational inference
    \item \textbf{SWAG} \cite{maddox2019simple}: Fit Gaussian to SGD trajectory
    \item \textbf{HMC:} Gold standard but computationally expensive
\end{itemize}

\paragraph{Bayesian PINNs.}
The PDE residual acts as additional likelihood term \cite{yang2022multi}:
\begin{equation}
p(\theta \mid \mathcal{D}) \propto p(\mathcal{D} \mid \theta) \cdot p(\text{PDE satisfied} \mid \theta) \cdot p(\theta)
\label{eq:supp_bpinn}
\end{equation}

Total predictive uncertainty decomposes:
\begin{equation}
\text{Var}[u^*] = \underbrace{\mathbb{E}_\theta[\text{Var}(u^* \mid \theta)]}_{\text{aleatoric}} + \underbrace{\text{Var}_\theta(\mathbb{E}[u^* \mid \theta])}_{\text{epistemic}}
\label{eq:supp_uncertainty_decomp}
\end{equation}

\subsubsection{Generative Adversarial Networks}

GANs learn solution distributions implicitly via adversarial training \cite{goodfellow2014generative}:
\begin{equation}
\min_G \max_D \mathbb{E}_{u \sim p_{\text{data}}}[\log D(u)] + \mathbb{E}_{z \sim p_z}[\log(1 - D(G(z)))]
\label{eq:supp_gan}
\end{equation}

For PDEs, physics can be incorporated through \cite{fuhg2023deep}:
\begin{itemize}
    \item Residual-aware discriminators
    \item Physics constraint penalties in generator loss
    \item Conservation law enforcement
\end{itemize}

Challenges include training instability (mode collapse, oscillation) and lack of likelihood-based evaluation; physics metrics (residuals, conservation errors) are used instead.


\section{Physics-Informed Methods: Extended}
\label{sec:supp_pinns}

\subsection{PINN Variant Formulations}
\label{sec:supp_pinn_variants}

\subsubsection{Variational Physics-Informed Neural Networks}

VPINNs incorporate weak formulations for problems with variational structure \cite{kharazmi2019variational,rojas2024robust}. For elliptic problems with energy functional $E[u]$:
\begin{equation}
\mathcal{L}_{\text{VPINN}} = E[u_\theta] + \lambda_{\text{BC}} \mathcal{L}_{\text{BC}}
\label{eq:supp_vpinn_full}
\end{equation}

For the Poisson equation $-\Delta u = f$:
\begin{equation}
E[u] = \frac{1}{2}\int_\Omega |\nabla u|^2 dx - \int_\Omega fu \, dx
\label{eq:supp_poisson_energy}
\end{equation}

Advantages over strong-form PINNs:
\begin{itemize}
    \item Natural coercivity improves optimization conditioning
    \item Lower derivative requirements (first vs. second order)
    \item Achieves $10^{-4}$--$10^{-2}$ vs. $10^{-3}$--$10^{-1}$ relative errors
\end{itemize}

\subsubsection{Extended PINNs (XPINNs)}

XPINNs employ domain decomposition with separate networks per subdomain \cite{jagtap2020extended,hu2021extended}:
\begin{equation}
\mathcal{L}_{\text{XPINN}} = \sum_{k=1}^K \mathcal{L}_k + \sum_{(i,j)} \lambda_{ij} \mathcal{L}_{\Gamma_{ij}}
\label{eq:supp_xpinn_full}
\end{equation}

Interface conditions on $\Gamma_{ij}$ enforce:
\begin{align}
\mathcal{L}_{\Gamma_{ij}}^{\text{cont}} &= \|u_\theta^{(i)} - u_\theta^{(j)}\|^2_{\Gamma_{ij}} \label{eq:supp_xpinn_cont}\\
\mathcal{L}_{\Gamma_{ij}}^{\text{flux}} &= \left\|\frac{\partial u_\theta^{(i)}}{\partial n} - \frac{\partial u_\theta^{(j)}}{\partial n}\right\|^2_{\Gamma_{ij}} \label{eq:supp_xpinn_flux}
\end{align}

Benefits:
\begin{itemize}
    \item Parallel training across subdomains
    \item Localized network capacity for varying solution complexity
    \item Better handling of heterogeneous materials
\end{itemize}

\subsubsection{Conservative PINNs}

Conservative PINNs ensure strict conservation law satisfaction \cite{jagtap2020conservative}. For conservation law $\partial_t u + \nabla \cdot F(u) = 0$:

\paragraph{Flux-Based Formulation.}
Parameterize potential $\Psi$ such that:
\begin{equation}
u = \nabla \times \Psi \quad \text{(ensures } \nabla \cdot u = 0 \text{ in 3D)}
\label{eq:supp_conservative}
\end{equation}
In 2D, the analogous construction uses a scalar streamfunction $\psi$ with $u = \partial_y \psi$, $v = -\partial_x \psi$, automatically satisfying the incompressibility constraint.

\paragraph{Integral Conservation.}
Enforce conservation over control volumes:
\begin{equation}
\mathcal{L}_{\text{cons}} = \sum_V \left|\frac{d}{dt}\int_V u \, dx + \oint_{\partial V} F \cdot n \, dS\right|^2
\label{eq:supp_integral_cons}
\end{equation}

Critical for long-time integration where spurious sources/sinks corrupt evolution.

\subsubsection{Deep Ritz Method}

The Deep Ritz Method applies variational principles for elliptic PDEs \cite{ji2024deep,xu2024refined}:
\begin{equation}
u^* = \arg\min_{u \in H^1_0(\Omega)} E[u]
\label{eq:supp_deep_ritz}
\end{equation}

For $-\nabla \cdot (a \nabla u) + cu = f$ with Dirichlet BCs:
\begin{equation}
E[u] = \frac{1}{2}\int_\Omega \left(a|\nabla u|^2 + cu^2\right) dx - \int_\Omega fu \, dx
\label{eq:supp_ritz_energy}
\end{equation}

Natural extension to high-dimensional problems where mesh-based methods face curse of dimensionality.

\subsubsection{Weak Adversarial Networks}

WANs employ adversarial training to avoid explicit high-order derivatives \cite{zang2020weak,oliva2022towards}:
\begin{equation}
\min_u \max_v \int_\Omega \mathcal{N}[u] \cdot v \, dx - \lambda \|v\|^2
\label{eq:supp_wan}
\end{equation}

Generator network $u_\theta$ approximates solution; discriminator network $v_\phi$ identifies PDE violations. The weak form transfers derivatives to test functions via integration by parts.

\subsubsection{Multi-Fidelity PINNs}

Multi-fidelity PINNs leverage data of varying accuracy \cite{meng2020composite,taghizadeh2024multi}:
\begin{equation}
u_{\text{HF}}(x) = \rho(x) \cdot u_{\text{LF}}(x) + \delta(x)
\label{eq:supp_multifidelity}
\end{equation}

Components:
\begin{itemize}
    \item $u_{\text{LF}}$: Low-fidelity solution (coarse simulations, simplified physics)
    \item $\rho(x)$: Learned correlation function
    \item $\delta(x)$: Learned discrepancy function
\end{itemize}

Training proceeds hierarchically:
\begin{enumerate}
    \item Train $u_{\text{LF}}$ on abundant low-fidelity data
    \item Fix $u_{\text{LF}}$, train $\rho$ and $\delta$ on sparse high-fidelity data
\end{enumerate}

\subsubsection{Adaptive Sampling Strategies}

Adaptive PINNs dynamically refine collocation point distributions \cite{wu2023comprehensive,torres2025adaptive}:

\paragraph{Residual-Based Sampling.}
\begin{equation}
p(x) \propto |\mathcal{N}[u_\theta](x)|^\alpha
\label{eq:supp_residual_sampling}
\end{equation}
where $\alpha > 0$ controls aggressiveness. Points concentrate where residuals are large.

\paragraph{Gradient-Based Sampling.}
\begin{equation}
p(x) \propto \|\nabla_\theta \mathcal{L}(x)\|
\label{eq:supp_gradient_sampling}
\end{equation}
Samples points with high gradient contribution.

\paragraph{Evolutionary Strategies.}
Maintain population of collocation sets; select and mutate based on loss reduction.

\subsection{Training Dynamics Analysis}
\label{sec:supp_training}

\subsubsection{Gradient Pathologies}

PINN training suffers from competing gradient signals between loss components. For composite loss $\mathcal{L} = \sum_i \lambda_i \mathcal{L}_i$:
\begin{equation}
\nabla_\theta \mathcal{L} = \sum_i \lambda_i \nabla_\theta \mathcal{L}_i
\label{eq:supp_gradient_sum}
\end{equation}

Pathological behaviors include:
\begin{itemize}
    \item \textbf{Gradient dominance:} One term dominates, others neglected
    \item \textbf{Conflicting gradients:} $\nabla \mathcal{L}_i \cdot \nabla \mathcal{L}_j < 0$
    \item \textbf{Scale mismatch:} $\|\nabla \mathcal{L}_i\| \gg \|\nabla \mathcal{L}_j\|$
\end{itemize}

\subsubsection{Loss Balancing Schemes}

\paragraph{Learning Rate Annealing.}
Adjust $\lambda_i$ based on loss magnitude:
\begin{equation}
\lambda_i^{(t+1)} = \lambda_i^{(t)} \cdot \left(\frac{\mathcal{L}_i^{(t)}}{\sum_j \mathcal{L}_j^{(t)}}\right)^{-\beta}
\label{eq:supp_lr_annealing}
\end{equation}

\paragraph{Gradient Normalization.}
Normalize gradient contributions:
\begin{equation}
\tilde{\nabla}_i = \frac{\nabla_\theta \mathcal{L}_i}{\|\nabla_\theta \mathcal{L}_i\| + \epsilon}
\label{eq:supp_grad_norm}
\end{equation}

\paragraph{Neural Tangent Kernel Analysis.}
NTK theory provides insight into PINN training dynamics, revealing how different loss components evolve at different rates determined by the kernel eigenspectrum.

\subsubsection{Causal Training}

For time-dependent problems, causal training respects temporal ordering \cite{toscano2025pinns}:
\begin{equation}
\mathcal{L} = \sum_{n=1}^{N_t} w_n \mathcal{L}(t_n), \quad w_n = \exp\left(-\epsilon \sum_{k=1}^{n-1} \mathcal{L}(t_k)\right)
\label{eq:supp_causal}
\end{equation}

Early times are prioritized; later times receive weight only after earlier times are resolved. This prevents information leakage from future to past during optimization.


\section{Meta-Learning for PDEs}
\label{sec:supp_metalearning}

Meta-learning addresses rapid adaptation to new PDE problems with minimal problem-specific data. The core principle is learning shared structure across a distribution of related PDEs, then exploiting this structure for efficient specialization.

\subsection{Problem Formulation}

Meta-learning defines a task distribution $p(\mathcal{T})$ where each task $\mathcal{T}_i$ corresponds to a PDE instance with parameters $\lambda_i$, boundary conditions $g_i$, or domain $\Omega_i$. The meta-learning objective:
\begin{equation}
\min_\phi \mathbb{E}_{\mathcal{T} \sim p(\mathcal{T})}\left[\mathcal{L}_\mathcal{T}(\mathcal{A}_\phi(\mathcal{D}_\mathcal{T}^{\text{train}}); \mathcal{D}_\mathcal{T}^{\text{test}})\right]
\label{eq:supp_meta_objective}
\end{equation}
where $\mathcal{A}_\phi$ is a learning algorithm parameterized by $\phi$.

\subsection{Optimization-Based Methods}

\subsubsection{Model-Agnostic Meta-Learning (MAML)}

MAML learns initialization enabling rapid gradient-based adaptation \cite{finn2017model}:
\begin{equation}
\theta^* = \arg\min_\theta \mathbb{E}_{\mathcal{T}}\left[\mathcal{L}_\mathcal{T}(\theta - \alpha \nabla_\theta \mathcal{L}_\mathcal{T}(\theta))\right]
\label{eq:supp_maml}
\end{equation}

For $K$ inner steps:
\begin{equation}
\theta_K = \theta - \alpha \sum_{k=0}^{K-1} \nabla_\theta \mathcal{L}_\mathcal{T}(\theta_k)
\label{eq:supp_maml_inner}
\end{equation}

Computing meta-gradients requires second derivatives (Hessian-vector products). First-order approximations (FOMAML) often perform comparably:
\begin{equation}
\nabla_\theta \mathcal{L}_\mathcal{T}(\theta_K) \approx \nabla_{\theta_K} \mathcal{L}_\mathcal{T}(\theta_K)
\label{eq:supp_fomaml}
\end{equation}

\subsubsection{Reptile}

Reptile simplifies MAML by avoiding second derivatives \cite{nichol2018first}:
\begin{equation}
\theta \leftarrow \theta + \epsilon(\tilde{\theta}_\mathcal{T} - \theta)
\label{eq:supp_reptile}
\end{equation}
where $\tilde{\theta}_\mathcal{T}$ results from $K$ SGD steps on task $\mathcal{T}$.

\subsubsection{Meta-Learned PINNs}

Applying MAML to physics-informed networks \cite{li2022meta}:
\begin{equation}
\mathcal{L}_\mathcal{T}(\theta) = \lambda_{\text{PDE}} \|\mathcal{N}_{\lambda_\mathcal{T}}[u_\theta]\|^2 + \lambda_{\text{BC}} \|u_\theta - g_\mathcal{T}\|^2_{\text{BC}}
\label{eq:supp_meta_pinn}
\end{equation}

Reported 1-2 orders of magnitude reduction in adaptation iterations versus random initialization.

\subsection{Metric-Based Methods}

\subsubsection{Prototypical Networks}

Compute class prototypes as mean embeddings \cite{snell2017prototypical}:
\begin{equation}
c_k = \frac{1}{|S_k|} \sum_{(x_i, u_i) \in S_k} f_\phi(x_i, u_i)
\label{eq:supp_prototype}
\end{equation}

Classification via distances:
\begin{equation}
p(y = k \mid x) = \frac{\exp(-d(f_\phi(x), c_k))}{\sum_{k'} \exp(-d(f_\phi(x), c_{k'}))}
\label{eq:supp_proto_classify}
\end{equation}

\subsubsection{Matching Networks}

Attention-weighted predictions \cite{vinyals2016matching}:
\begin{equation}
\hat{u}(x^*) = \sum_{i=1}^{N_C} a(x^*, x_i) u_i
\label{eq:supp_matching}
\end{equation}
where attention $a$ is computed via learned embeddings.

\subsection{Model-Based Methods}

\subsubsection{Neural Processes}

Probabilistic framework for function prediction from context \cite{garnelo2018neural}:
\begin{equation}
p(u_{\text{target}} \mid x_{\text{target}}, C) = \int p(u_{\text{target}} \mid x_{\text{target}}, z) p(z \mid C) dz
\label{eq:supp_np}
\end{equation}

Encoder aggregates context permutation-invariantly:
\begin{equation}
z \sim q(z \mid C) = \mathcal{N}(\mu_\phi(r), \sigma_\phi(r)), \quad r = \frac{1}{N_C} \sum_{i=1}^{N_C} h_\phi(x_i, u_i)
\label{eq:supp_np_encoder}
\end{equation}

\subsubsection{Attentive Neural Processes}

Replace mean aggregation with attention \cite{kim2019attentive}:
\begin{equation}
r(x^*) = \sum_{i=1}^{N_C} \alpha(x^*, x_i) h_\phi(x_i, u_i)
\label{eq:supp_anp}
\end{equation}

Location-dependent aggregation improves interpolation quality.

\subsection{Hypernetworks}

Generate task-specific parameters via forward pass \cite{ha2016hypernetworks}:
\begin{equation}
\theta_\mathcal{T} = h_\phi(P_\mathcal{T}), \quad u(x) = f_{\theta_\mathcal{T}}(x)
\label{eq:supp_hypernetwork}
\end{equation}

For large target networks, use:
\begin{itemize}
    \item \textbf{Low-rank generation:} $\theta = \theta_{\text{base}} + U h_\phi(P) V^T$
    \item \textbf{Layer-wise generation:} Separate hypernetworks per layer
    \item \textbf{Chunked generation:} Parameters in chunks
\end{itemize}

\subsection{Theoretical Foundations}

PAC-Bayes bounds establish \cite{amit2018meta}:
\begin{equation}
\mathcal{L}_{\text{test}} \leq \mathcal{L}_{\text{train}} + O\left(\sqrt{\frac{\text{complexity}(\mathcal{A})}{N_{\text{tasks}}}} + \sqrt{\frac{\text{complexity}(f)}{N_{\text{examples}}}}\right)
\label{eq:supp_pac_bayes}
\end{equation}

Effective meta-learning requires:
\begin{itemize}
    \item Sufficient task diversity (coverage of test distribution)
    \item Parameter ranges encompassing test configurations
    \item Geometric variations including test domain types
\end{itemize}

\begin{table}[h!]
\centering
\caption{Meta-learning method comparison for PDE applications.}
\label{tab:supp_meta_comparison}
\begin{tabular}{p{3cm}p{2.5cm}p{3cm}p{2.5cm}p{2.5cm}}
\toprule
\textbf{Method} & \textbf{Adaptation Cost} & \textbf{Expressiveness} & \textbf{Uncertainty} & \textbf{Best For} \\
\midrule
MAML & Medium & High & No & Maximum quality \\
Reptile & Medium & High & No & Simple implementation \\
Neural Processes & Low & Medium & Yes & Sparse observations \\
Hypernetworks & Low & Medium & No & Real-time adaptation \\
Transfer Learning & Medium & High & No & Hierarchical tasks \\
\bottomrule
\end{tabular}
\end{table}


\section{Hybrid Method Formulations}
\label{sec:supp_hybrid}

\subsection{Random Feature Methods}

Random Feature Methods parameterize solutions as \cite{chen2022bridging}:
\begin{equation}
u_M(x) = \sum_{j=1}^M c_j \sigma(\omega_j \cdot x + b_j)
\label{eq:supp_rf}
\end{equation}
with random $\{\omega_j, b_j\}$ fixed at initialization.

Coefficients solve linear system:
\begin{equation}
\Phi c = f, \quad \Phi_{ij} = \mathcal{N}[\sigma(\omega_j \cdot x_i + b_j)]
\label{eq:supp_rf_system}
\end{equation}

\paragraph{Kernel Connection.}
Random features approximate kernels \cite{rahimi2007random}:
\begin{equation}
k(x, x') = \mathbb{E}_{\omega, b}[\sigma(\omega \cdot x + b) \sigma(\omega \cdot x' + b)]
\label{eq:supp_rf_kernel}
\end{equation}

For Gaussian $\omega$, this yields Gaussian RBF kernels. Approximation theory establishes $O(M^{-1/2})$ convergence in expectation.

\paragraph{Advantages.}
\begin{itemize}
    \item Convex optimization (linear least-squares)
    \item Theoretical error bounds from kernel theory
    \item Deterministic solutions given fixed seeds
    \item No training instabilities
\end{itemize}

\subsection{Adjoint Method Derivation}

For optimization through time-dependent simulations:
\begin{equation}
\min_\theta J = \int_0^T L(u(t), \theta) dt \quad \text{s.t.} \quad \dot{u} = f(u, \theta)
\label{eq:supp_adjoint_problem}
\end{equation}

The adjoint variable $\lambda$ satisfies:
\begin{equation}
\dot{\lambda} = -\left(\frac{\partial f}{\partial u}\right)^T \lambda - \frac{\partial L}{\partial u}, \quad \lambda(T) = 0
\label{eq:supp_adjoint_ode}
\end{equation}

Gradient computation:
\begin{equation}
\frac{dJ}{d\theta} = \int_0^T \left(\frac{\partial L}{\partial \theta} + \lambda^T \frac{\partial f}{\partial \theta}\right) dt
\label{eq:supp_adjoint_gradient}
\end{equation}

Memory reduction: $O(N)$ instead of $O(NT)$ for $T$ timesteps.


\section{Theoretical Foundations}
\label{sec:supp_theory}

\subsection{Universal Approximation for Operators}

The Chen-Chen theorem \cite{chen1995universal} establishes: for compact $K \subset \mathbb{R}^d$ and continuous operator $\mathcal{G}: C(K) \to C(K')$, there exist continuous functions $g_i$, $f_i$, and points $\{x_j\}$ such that:
\begin{equation}
\left|\mathcal{G}[u](y) - \sum_{i=1}^n g_i(y) \sum_{j=1}^m f_i(u(x_j))\right| < \epsilon
\label{eq:supp_universal}
\end{equation}
for all $u \in C(K)$ and $y \in K'$.

This provides the theoretical foundation for DeepONet's branch-trunk architecture.

\subsection{FNO Approximation Rates}

For operators mapping between Sobolev spaces $H^s \to H^{s'}$, FNO achieves \cite{kovachki2023neural}:
\begin{equation}
\|\mathcal{G} - \mathcal{G}_\theta\|_{H^s \to H^{s'}} \leq C \cdot k_{\max}^{-\min(s, s')/d}
\label{eq:supp_fno_rate}
\end{equation}
where $k_{\max}$ is the Fourier mode cutoff.

For smooth solutions ($s$ large), this approaches spectral convergence. However, solutions with discontinuities ($s < 1$) converge slowly.

\subsection{What's Missing}

Despite progress, significant theoretical gaps remain:

\begin{itemize}
    \item \textbf{A priori error bounds:} No rigorous bounds accounting for network architecture, training algorithm, and PDE properties simultaneously
    
    \item \textbf{Stability analysis:} CFL-type conditions for neural PDE solvers do not exist; long-time behavior is empirically characterized
    
    \item \textbf{Convergence guarantees:} Training dynamics analysis (via NTK) provides insight but not guarantees for finite-width networks
    
    \item \textbf{Generalization theory:} Bounds exist but are often vacuous; practical generalization behavior remains empirical
\end{itemize}


\section{Comprehensive Comparison Tables}
\label{sec:supp_tables}

\begin{landscape}
\small
\begin{longtable}{>{\raggedright\arraybackslash}p{2.5cm} >{\raggedright\arraybackslash}p{2.5cm} >{\raggedright\arraybackslash}p{2cm} >{\raggedright\arraybackslash}p{1.5cm} >{\raggedright\arraybackslash}p{1.5cm} >{\raggedright\arraybackslash}p{1.5cm} >{\raggedright\arraybackslash}p{1.5cm} >{\raggedright\arraybackslash}p{1.5cm} >{\raggedright\arraybackslash}p{4cm}}
\caption{Comprehensive ML-PDE method comparison. Accuracy values are indicative relative $L^2$ errors from PDEBench \cite{takamoto2022pdebench} and benchmark studies (2020--2025); results vary substantially with PDE family, coefficient regime (in-distribution vs. out-of-distribution), rollout horizon, resolution, and training protocol.}
\label{tab:supp_comprehensive} \\
\toprule
\textbf{Method} & \textbf{Architecture} & \textbf{Complexity} & \textbf{UQ} & \textbf{Multiscale} & \textbf{Nonlin.} & \textbf{Inference} & \textbf{Accuracy} & \textbf{Notes} \\
\midrule
\endfirsthead
\toprule
\textbf{Method} & \textbf{Architecture} & \textbf{Complexity} & \textbf{UQ} & \textbf{Multiscale} & \textbf{Nonlin.} & \textbf{Inference} & \textbf{Accuracy} & \textbf{Notes} \\
\midrule
\endhead
\midrule
\multicolumn{9}{r}{\textit{Continued on next page}} \\
\endfoot
\bottomrule
\endlastfoot

\multicolumn{9}{l}{\textit{\textbf{Physics-Informed Neural Networks}}} \\
\midrule
Standard PINN & Residual MLP & $O(N_{\text{pts}} \cdot P)$ & Limited & Moderate & Sensitive & Fast & $10^{-3}$--$10^{-1}$ & AD-based; spectral bias \\
VPINN & Energy functional & $O(N_{\text{quad}} \cdot P)$ & Limited & Good & Better & Fast & $10^{-4}$--$10^{-1}$ & Superior conditioning \\
XPINN & Domain decomp. & $O(K \cdot N \cdot P)$ & No & Good & Moderate & Medium & $10^{-3}$--$10^{-1}$ & Parallel subdomains \\
Conservative & Conservation arch. & $O(N_{\text{pts}} \cdot P)$ & No & Moderate & Physical & Fast & $10^{-3}$--$10^{-1}$ & Long-time stable \\
Deep Ritz & Variational & $O(N_{\text{quad}} \cdot P)$ & No & Limited & Smooth & Fast & $10^{-4}$--$10^{-1}$ & High-dimensional \\
WAN & Min-max & $O(N^2 \cdot P)$ & No & Moderate & Good & Medium & $10^{-3}$--$10^{-1}$ & Avoids high derivatives \\
Multi-fidelity & Hierarchical & $O(N_{\text{fid}} \cdot N \cdot P)$ & Yes & Good & Moderate & Fast & $10^{-4}$--$10^{-1}$ & Mixed data quality \\
Adaptive & Residual-adaptive & $O(N_{\text{adapt}} \cdot N \cdot P)$ & Limited & Good & Moderate & Medium & $10^{-4}$--$10^{-1}$ & Dynamic refinement \\

\midrule
\multicolumn{9}{l}{\textit{\textbf{Neural Operators}}} \\
\midrule
FNO & Fourier layers & $O(N \log N)$ & Approx. (ensemble) & Limited & Periodic & Very Fast & $10^{-2}$--$10^{-1}$ & Resolution invariant \\
DeepONet & Branch-trunk & $O(N_s \cdot N_q)$ & Approx. (dropout) & Moderate & Smooth & Fast & $10^{-2}$--$10^{-1}$ & Flexible input/output \\
GNO & Graph kernel & $O(|E|)$ & Limited & Good & Moderate & Fast & $10^{-2}$--$10^{-1}$ & Irregular domains \\
Wavelet NO & Wavelet basis & $O(N \log N)$ & No & Very Good & Good & Fast & $10^{-2}$--$10^{-1}$ & Multi-resolution \\
U-FNO & U-Net + Fourier & $O(N \log N \cdot L)$ & Limited & Very Good & Good & Fast & $10^{-2}$--$10^{-1}$ & Scale separation \\
GINO & Geometry-aware & $O(N \log N)$ & Limited & Good & Good & Fast & $10^{-2}$--$10^{-1}$ & Complex domains \\
Latent NO & Compressed & $O(d_z \log d_z)$ & Limited & Moderate & Good & Fast & $10^{-2}$--$10^{-1}$ & Memory efficient \\
Foundation & Large pretrained & $O(N \log N)$ & Transfer & Good & Transfer & Fast & $10^{-2}$--$10^{-1}$ & Zero/few-shot \\

\midrule
\multicolumn{9}{l}{\textit{\textbf{Graph Neural Networks}}} \\
\midrule
GNS & Encode-process-decode & $O(|E| \cdot M)$ & Limited & Moderate & Good & Fast & $10^{-2}$--$10^{-1}$ & Particles/meshes \\
MeshGraphNets & Dual edges & $O(|E| \cdot M)$ & Limited & Good & Good & Fast & $10^{-2}$--$10^{-1}$ & Mesh dynamics \\
MP-PDE & Temporal bundle & $O(|E| \cdot M \cdot K)$ & Limited & Good & Good & Fast & $10^{-2}$--$10^{-1}$ & Reduced rollout error \\
E(n)-GNN & Equivariant & $O(|E| \cdot M)$ & Limited & Moderate & Good & Fast & $10^{-2}$--$10^{-1}$ & Symmetry preserving \\
Multiscale GNN & Hierarchical & $O(|E| \log |V|)$ & Limited & Very Good & Good & Fast & $10^{-2}$--$10^{-1}$ & Multigrid-like \\

\midrule
\multicolumn{9}{l}{\textit{\textbf{Transformers}}} \\
\midrule
Galerkin & Linear attention & $O(N \cdot d^2)$ & Not native & Good & Good & Medium & $10^{-2}$--$10^{-1}$ & Learned basis \\
Fourier Trans. & Spectral attention & $O(k \cdot d^2)$ & Not native & Good & Good & Medium & $10^{-2}$--$10^{-1}$ & Mode truncation \\
Factorized & Axial attention & $O(N^{4/3})$ & Not native & Good & Good & Medium & $10^{-2}$--$10^{-1}$ & 3D tractable \\
Operator Trans. & Cross-attention & $O(N_s \cdot N_q)$ & Not native & Good & Good & Medium & $10^{-2}$--$10^{-1}$ & Mesh-free output \\

\midrule
\multicolumn{9}{l}{\textit{\textbf{Generative Models}}} \\
\midrule
Diffusion & Score-based & $O(T \cdot N)$ & Native & Limited & Stochastic & Slow & $10^{-2}$--$10^{-1}$ & Full distributions \\
VAE & Latent encoding & $O(d_z^2)$ & Native & Good & Moderate & Fast & $10^{-2}$--$10^{-1}$ & Fast sampling \\
Flow & Invertible & $O(L \cdot N)$ & Native & Limited & Good & Medium & $10^{-2}$--$10^{-1}$ & Exact likelihood \\
BNN & Posterior & $O(S \cdot N \cdot P)$ & Native & Moderate & Good & Slow & $10^{-2}$--$10^{-1}$ & Epistemic UQ \\

\midrule
\multicolumn{9}{l}{\textit{\textbf{Hybrid Methods}}} \\
\midrule
Neural-FEM & FEM + NN & Varies & Limited & Very Good & Good & Medium & $10^{-4}$--$10^{-2}$ & Conservation exact \\
PEDS & LF solver + NN & $O(N_{\text{LF}} + P)$ & Limited & Good & Moderate & Fast & $10^{-3}$--$10^{-1}$ & 10-100$\times$ data efficient \\
DeePoly & NN + polynomial & $O(P + p^d)$ & No & Good & Moderate & Fast & $10^{-10}$--$10^{-4}$$^\ddagger$ & Smooth problems only \\
Random Features & Fixed basis & $O(M \cdot N)$ & Limited & Moderate & Linear & Fast & $10^{-3}$--$10^{-2}$ & Convex, stable \\
Neural Precon. & Learned inverse & $O(N \cdot P)$ & No & Good & Linear & Fast & N/A & Accelerates solvers \\

\midrule
\multicolumn{9}{l}{\textit{\textbf{Meta-Learning}}} \\
\midrule
MAML & Bi-level opt. & $O(K \cdot N \cdot P)$ & Limited & Moderate & Transfer & Fast & $10^{-2}$--$10^{-1}$ & Few-shot adaptation \\
Neural Process & Stochastic & $O(N_C \cdot N_T \cdot d)$ & Native & Good & Contextual & Fast & $10^{-2}$--$10^{-1}$ & Uncertainty from context \\
Hypernetwork & Weight generation & $O(P_{\text{hyper}})$ & Limited & Good & Adaptive & Very Fast & $10^{-2}$--$10^{-1}$ & Instant adaptation \\

\end{longtable}

\vspace{1em}
\textbf{Legend:} UQ = Uncertainty Quantification; Nonlin. = Nonlinearity handling; $P$ = parameters; $N$ = grid/points; $M$ = message-passing steps; $K$ = inner steps/temporal bundle; $T$ = diffusion steps; $S$ = samples; $d_z$ = latent dimension; $k$ = Fourier modes. Accuracy values are indicative relative $L^2$ errors; performance varies significantly with PDE family, coefficient regime (ID/OoD), rollout horizon, resolution, and training protocol. $^\ddagger$For sufficiently smooth solutions admitting high-order polynomial refinement.
\end{landscape}

\end{document}